\documentclass[10pt,twocolumn,letterpaper]{article}

\usepackage{iccv}
\usepackage{times}
\usepackage{epsfig}
\usepackage{graphicx}
\usepackage{amsmath}
\usepackage{amssymb}
\usepackage{capt-of}
\usepackage{booktabs}
\usepackage{color, colortbl}

\usepackage[breaklinks=true,bookmarks=false]{hyperref}
\usepackage{multirow}

\iccvfinalcopy 

\def\iccvPaperID{****} 

\ificcvfinal\fi

\begin{document}

\title{Controlled and Conditional Text to Image Generation with Diffusion Prior}

\author{Pranav Aggarwal \qquad Hareesh Ravi \qquad Naveen Marri \qquad Sachin Kelkar \qquad Fengbin Chen \qquad \\ Vinh Khuc \qquad Midhun Harikumar \qquad Ritiz Tambi \qquad Sudharshan Reddy Kakumanu \qquad Purvak Lapsiya \\ \qquad Alvin Ghouas \qquad Sarah Saber \qquad Malavika Ramprasad \qquad Baldo Faieta \qquad Ajinkya Kale \\  \\ 
Adobe Applied Research}
\twocolumn[{%
\renewcommand\twocolumn[1][]{#1}%
\maketitle
\vspace*{-1cm}
\begin{center}
    \centering
    \includegraphics[height=3in, width=\linewidth]{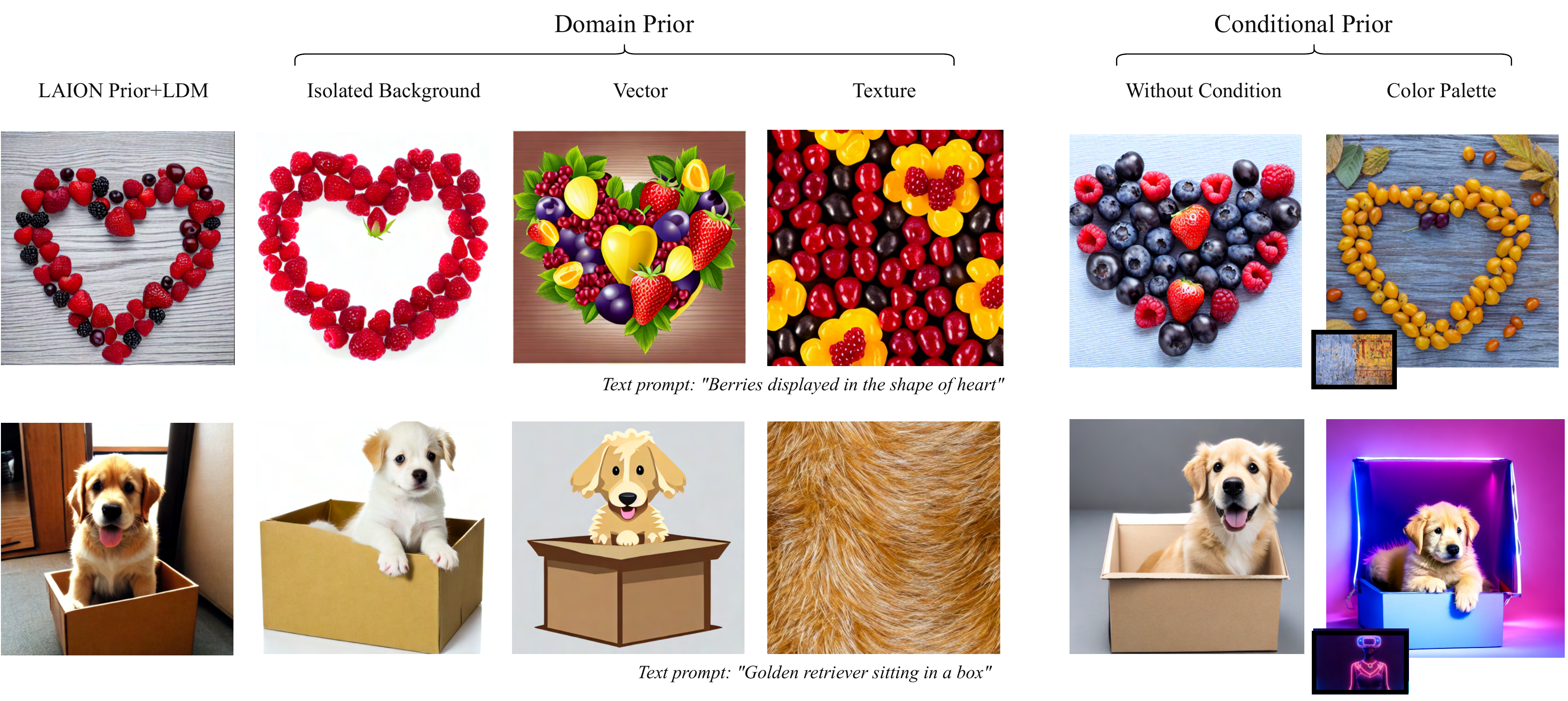}
    \captionof{figure}{\emph{Diffusion Prior} can be trained on image--text pairs of a specific domain to constrain generations without altering the larger \emph{Diffusion Decoder} model. We can see that different domain specific priors generate images within the corresponding domain for the same prompt. Conditioning the prior with color information (last two columns, color image provided at the bottom left edge) drives the generation towards a semantically meaningful color conditioned image. For reference, we also show result from the baseline LAION prior in the first column. Note that all these images were generated by the same Diffusion Decoder LDM showing the versatility of the Diffusion Prior for controllable and conditional text to image generation.}
    \label{fig:image_intro}
\end{center}}]
\ificcvfinal\thispagestyle{empty}\fi

\begin{abstract}
\vspace{-0.01in}
Denoising Diffusion models have shown remarkable performance in generating diverse, high quality images from text. Numerous techniques have been proposed on top of or in alignment with models like \emph{Stable Diffusion} and \emph{Imagen} that generate images directly from text. A lesser explored approach is \emph{DALLE-2}'s two step process comprising a \emph{Diffusion Prior} that generates a CLIP image embedding from text and a \emph{Diffusion Decoder} that generates an image from a CLIP image embedding. We explore the capabilities of the Diffusion Prior and the advantages of an intermediate CLIP representation. We observe that Diffusion Prior can be used in a memory and compute efficient way to constrain the generation to a specific domain without altering the larger Diffusion Decoder. Moreover, we show that the Diffusion Prior can be trained with additional conditional information such as color histogram to further control the generation. We show quantitatively and qualitatively that the proposed approaches perform better than prompt engineering for domain specific generation and existing baselines for color conditioned generation. We believe that our observations and results will instigate further research into the diffusion prior and uncover more of its capabilities. \\

\end{abstract}
\vspace{-0.2in}
\section{Introduction}
Diffusion Models \cite{ddpm} have shown remarkable performance in generating high quality and diverse images from text \cite{classifier_free_guidance, glide}. There have been numerous efforts to improve quality of generation \cite{dalle2, imagen}, improve training and sampling speed \cite{ldm, distillation_diffusion, ddim, dpm_solver_pp} and apply such models for editing \cite{sdedit, prompt2prompt, imagic, instructpix2pix} or finetuning for conditional or domains specific generations \cite{controlnet, multidiffusion}. However, most of these works focus on architectures similar to \cite{ldm, imagen} that condition the diffusion decoder directly on text embedding and encodings. In \cite{dalle2}, authors propose a text--to--image generation model comprising of two steps. Firstly, a \emph{Diffusion Prior} model generates CLIP \cite{clip} image embedding conditioned on CLIP text embedding of the input text. Following that, a \emph{Diffusion Decoder} model generates the final image conditioned on the generated CLIP image embedding. There is limited research based on this approach largely because of the lack of publicly available models, unlike the text conditioned Latent Diffusion Models (LDM) \cite{ldm}.

In \cite{dalle2}, the authors showed that having a common intermediate CLIP representation allows for improved diversity, native support for image variations, interpolations and latent manipulation. In this paper, we look further into the capabilities and advantages of having a common intermediate representation as well as the \emph{Diffusion Prior} model. Since there aren't publicly available weights or code for the DALLE-2 model, we leverage the \emph{Diffusion Prior} model\footnote{https://github.com/LAION-AI/conditioned-prior\label{laionprior}} trained by LAION\footnote{https://laion.ai} as the baseline for our experiments. For the \emph{Diffusion Decoder} as described in Appendix Sec. \ref{appendix:proposed-ldm}, we train a LDM using the publicly available code from \cite{ldm} but modify the cross--attention layers to condition on CLIP-L/14 image embedding instead of text embedding and encodings. We call our setup (similar to \cite{dalle2}) for text-to-image generation with \emph{Diffusion Prior} and \emph{Diffusion Decoder} together as a Hybrid Diffusion Model (HDM) different from DALLE-2 Hierarchical Diffusion Model that comprises of a sequence of diffusion models that generate and then upsample the generated images. 

We explore the possibilities with \emph{Diffusion Prior} model to answer the following question. "Can we control the image generations from HDM to be within a specific constrained \emph{desirable} sub-space in CLIP without finetuning the larger \emph{Diffusion Decoder} model?" To answer this question, we explore the following applications: (i) Text to Texture (ii) Text to Rasterized Vectors (iii) Text to Isolated Objects (iv) Color Conditioned Text to Image. We term the first three applications as \emph{domain specific generation} to emphasize that the generations are pixel images corresponding to specific domains that can further be used to synthesize texture \cite{texture1} or vectors in Scalable Vector Graphics (SVG) \cite{vectorfusion} format. The last problem is the well known \emph{conditional generation} where the conditional input is text and desired color histogram. 
Though we evaluate on these specific domains, we believe the approach is simple and robustly applicable across other domains and conditional inputs.

An example of what is possible with the proposed setup is shown in Fig.\ref{fig:image_intro}. 
These domain specific or conditional generations are difficult to generate with naive prompt engineering in existing models such as Stable Diffusion as shown in Sec.\ref{sec:results}. Generation of these domain specific images usually require finetuning the large model on domain specific data \cite{vectorfusion} which is computationally expensive. We train a new \emph{Diffusion Prior} model for each of these applications while keeping the larger \emph{Diffusion Decoder} model intact from the original pretrained HDM. We show that, by training smaller \emph{Diffusion Prior} models on domain specific dataset or conditioned on additional input, we can effectively control the generated images to a desired domain or correspond to a specific desirable color palette. We further show that the proposed setup is more efficient in terms of memory and compute since \emph{Diffusion Prior} model is significantly smaller (order of 5$\times$ \cite{dalle2}) compared to larger models \cite{vectorfusion, texture-finetune} and performs quantitatively and qualitatively better than existing baselines. 


The summary of our contributions are: \\
-- We explore the versatile capabilities of \emph{Diffusion Prior} and having a common intermediate CLIP embedding space by training small domain specific and color conditional prior models for controllable and conditional text-to-image generation. \\
-- To convert the outputs from our prior models to images, we train an LDM conditioned on normalized CLIP L/14 image embeddings to support our experiments. \\
-- We perform a comprehensive quantitative and qualitative performance evaluation on our proposed setup on three domains \emph{texture}, \emph{vector} and \emph{isolated objects} for domain specific generations and on \emph{color} for conditional generation.\\

To the best of our knowledge, there is no existing work that shows effective semantic aware color conditional generation and domain specific generations using the HDM architecture by modifying only the \emph{Diffusion Prior}. We hope that our observation and results lead to further research into the HDM architecture and the Diffusion Prior for various applications.

\section{Related Work}
\label{sec:related}
Diffusion models \cite{ddpm} (DMs) are likelihood-based models and have become more popular in image synthesis than Generative Adversarial Networks (GANs)\cite{diffusion_beat_gans, gan}. More details about preliminary relevant works in DMs are provided in Appendix Sec.\ref{sec:appendix-related}.

\subsection{Diffusion Prior}
\label{subsec:related-prior}
The \emph{Diffusion Prior} was introduced in OpenAI's DALLE-2 \cite{dalle2} which is a hierarchical text-conditional DM. The \emph{Diffusion Prior} is capable of mapping an input text embedding vector to an image embedding vector in a CLIP latent space. A decoder (unCLIP) then translates the CLIP image embedding into synthetic images. The \emph{Diffusion Prior} is a classifier-free guidance DM that uses a Transformer backbone instead of U-Net. In DALLE-2, the \emph{Diffusion Prior} is shown to outperform the autoregression prior in model size and training time. Recent works that use \emph{Diffusion Prior} models include Make-A-Video \cite{make_a_video}, Dream3D \cite{dream3d} and Shifted Diffusion \cite{shifted_diffusion}. While Shifted Diffusion \cite{shifted_diffusion} proposes a more optimal prior model, Make-A-Video and Dream3D utilize existing prior formulation to support text-to-image generation in their pipeline. Compared to text conditioned diffusion model like \cite{ldm, imagen}, there is limited work that builds on top of the \emph{Diffusion Prior} proposed in DALLE-2\cite{dalle2}. We explore its capabilities and its applications to domain specific and conditional generation. 

\subsection{Color Conditioned Generation}
\label{subsec:related-color}
Color is an important attribute of an image that provides contextual information as well as sets the mood of viewer's perception. Although there has been lot of research around generating images with specific styles as condition \cite{InvCreative, MagicMix, ediff_i, DiffStyler}, using only color palette for generating images has not been explored much by the research community. This could be a useful tool for artists and content creators to generate images with varied color palette portfolios without changing their authentic styles. There are some works on image colorization \cite{palette, Text2Colors, PalGan, deepExemplar} of grayscale images and image color transfer between colored images \cite{CT2001, ImageRecoloring, WCT2, DAST, Lee2020CTHA}, but these methods work well when an image is available as input. In \cite{afifi2021histogan}, the authors propose color conditional generation by controlling the color palette when generating images by injecting color histograms in log-chroma space into the StyleGan \cite{stylegan} architecture but do not show text and color conditioned generation. 

\noindent\textbf{Limitations:} Leveraging the aforementioned techniques for text and color conditioned image generation would be a two step process where we first generate an image given a text prompt and then perform color transfer on it. Though this is a possible solution, the lack of semantic awareness in the color transfer step and its independence to the generation step might lead to unsatisfactory results with color saturation artifacts. 
In our method, since the final generated image is from a valid CLIP embedding, the images look natural and are semantic aware compared to generic color transfer. 

\subsection{Domain Specific Generation}
\label{subsec:related-domain}
In recent years, domain adaptation has become popular in image synthesis where 
large pretrained models are fine-tuned on a smaller dataset from a specific domain like \cite{my_style_gan, dreambooth, textual_inversion}. In contrast to domain adaptation, \emph{domain specific generation} aims to constrain generations from a pretrained model to a specific sub space or domain within the seen larger distribution. \\ 
\noindent\textbf{Texture Domain:} Texture synthesis has been studied for a long time in Computer Vision (CV) literature \cite{early-texture1, early-texture2, early-texture3}. Recently, deep learning techniques have been used to generate textures \cite{texture1, texture2, texture3}. These approaches are a primer to 3D texture synthesis and transfer \cite{texture-3d}. Pretrained text-to-image diffusion models have also been explored for 3D texture and shape generation \cite{dreamfusion, texture-finetune}. \\ 
\noindent\textbf{Vectors and Isolated Objects Domain:} 
Scalable Vector Graphics (SVG) is a suitable format for expressing visual concepts as primitives and for exporting designs at arbitrarily high quality \cite{vectorfusion}. These are most commonly applicable when generating posters, templates, card and other graphics. Previous work such as \cite{vectorascent, vectordraw} optimize CLIP image similarity to generate vectors from text prompts. \cite{styleclipdraw} extends \cite{vectordraw} with style loss to generate vectors from images whereas \cite{fernando2021generative} uses a hierarchical neural Lindenmeyer system to optimize SVG paths and \cite{clipclop} uses evolutionary approach to generate collages. Recently, \cite{vectorfusion} proposed a multi-step optimization approach on top of Stable Diffusion \cite{ldm} to use the large conceptual pretrained knowledge of diffusion models for text to vector generation. \\
\noindent\textbf{Limitations:} Most of these techniques have as a first step to generate images that can easily be converted to vectors either by optimization or by using existing algorithms such as \cite{live}. Since SVGs comprise of simple abstract shapes and curves to compose concepts, photorealistic images generated by existing text-to-image diffusion models are not suitable for conversion to SVGs. However, these diffusion models have seen images that show abstract concepts \cite{vectorfusion} that can be generated using some prompt engineering. Though this works to a certain extent, this is not robust to all prompts
\cite{vectorfusion}. We overcome these limitations by proposing to train the \emph{Diffusion Prior} in HDM on images within a specific domain (textures, vectors or isolated objects). The generated images can then be used with existing vectorization \cite{live} and material generation \cite{texture-3d} techniques or as it is for further applications. \\ \\ 
\noindent\textbf{Concurrent Works:} More recently, concurrent works like ControlNet \cite{controlnet}, Composer \cite{composer} and T2I-Adapter \cite{t2iadapt} that support multiple conditional inputs and composable generations including color have been proposed. We emphasize that our paper is focused on revealing the capabilities of CLIP latent space based diffusion prior model which is lightweight and support conditional and domain specific generation without additional objectives or modifications in the base HDM architecture. This method is robust to domains and conditioning inputs and retains the advantages of the CLIP space such as latent interpolation and directional attribute manipulation while being efficient in terms of compute and memory.

\begin{figure*}[t]
    \centering
    \includegraphics[width=0.48\linewidth]{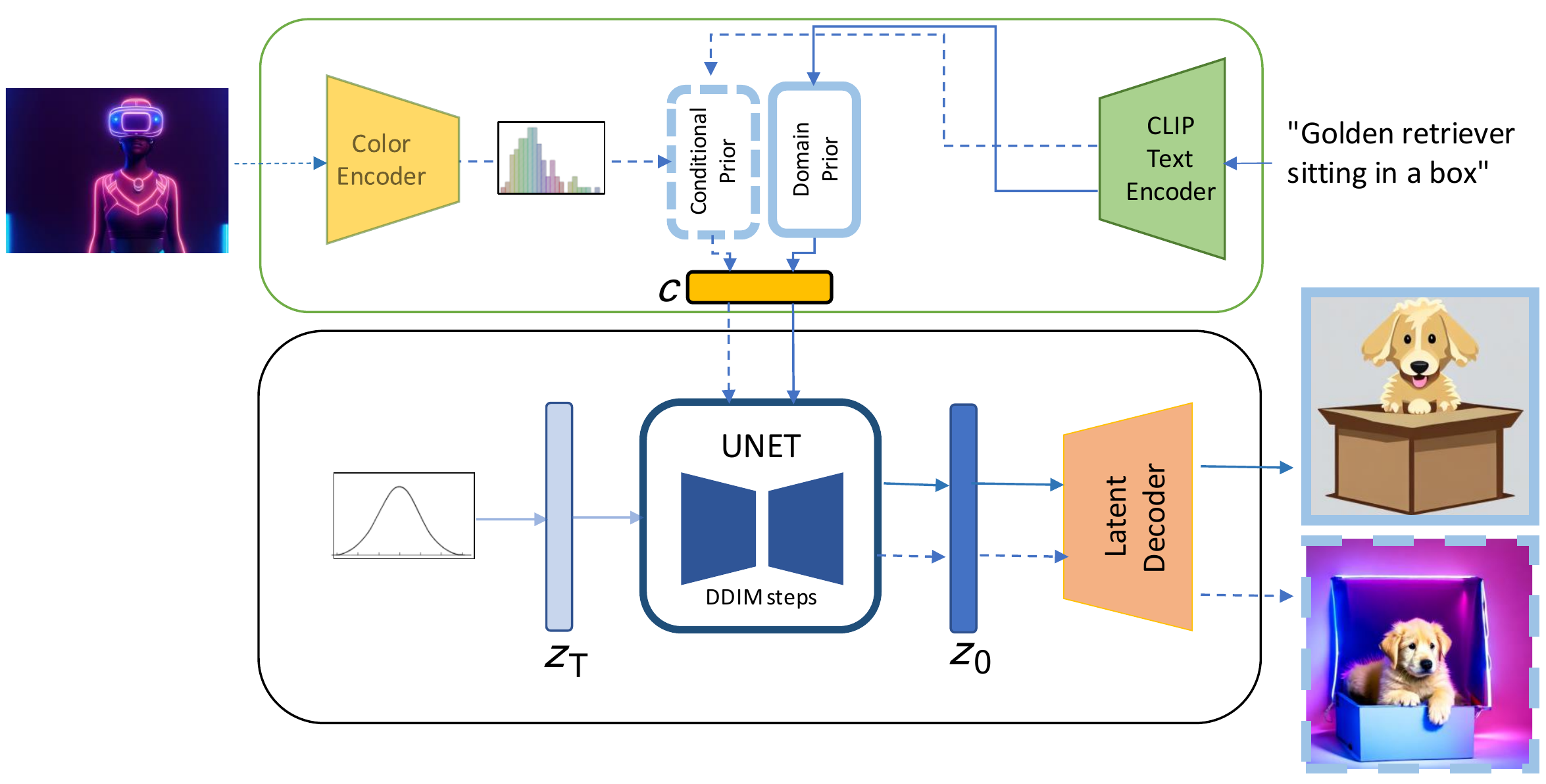}
    \hspace{0.03\linewidth}
    \includegraphics[width=0.48\linewidth]{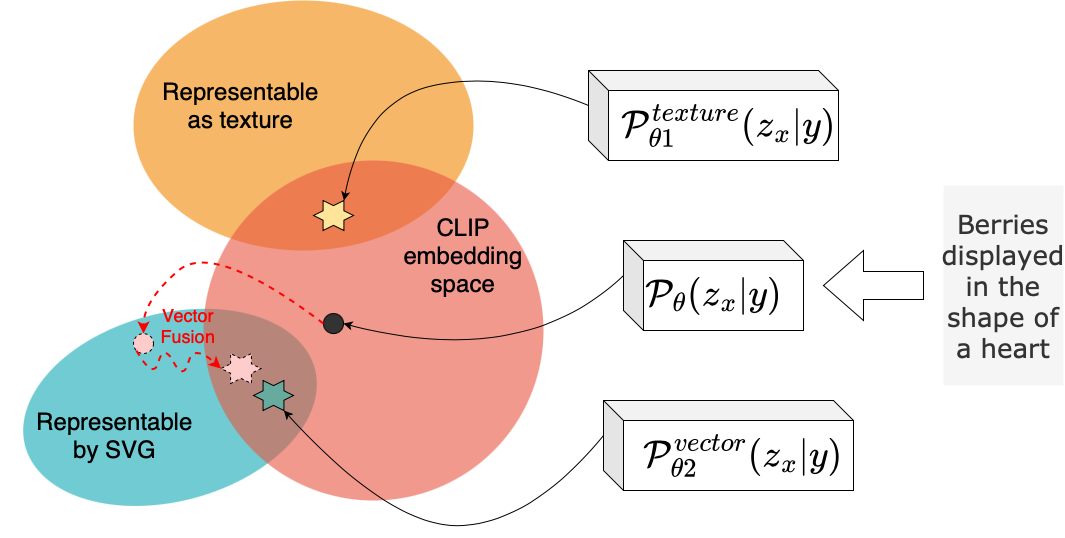}
    \caption{(Left) An overview of the inference pipeline for text to image generation with HDM consisting of \emph{Diffusion Prior} and \emph{Diffusion Decoder} models. The domain prior is shown for vector prior but is the same for other priors. The color prior takes as input text prompt and an optional color image to generate the final image. (Right) An illustration of the domain priors directly predicting plausible domain specific embeddings in the CLIP embedding space. Using prompt engineering is lossy while the domain specific priors do not generate outliers. Example shown for vector and texture priors but the concept applies to isolated object prior as well. Our approach does not require optimization \cite{vectorfusion} or finetuning.}
    \label{fig:block_diagram}
    \vspace{-0.2in}
\end{figure*}
\section{Proposed Method}
\label{sec:proposed}
Our base text-to-image generation model, HDM follows DALLE-2 architecture and has a \emph{Diffusion Prior} and \emph{Diffusion Decoder} model. Let $y$ be the text prompt provided as input to the HDM and $x$ the final generated image. An example of inference with our setup is shown in Fig.\ref{fig:block_diagram}. We follow the same two step hybrid architecture for text-to-image generation as \cite{dalle2}. For the prior, we use the pre-trained and publicly available LAION model as baseline for comparison and the corresponding architecture and code for training the domain and color priors. For the decoder, we train a custom LDM conditioned on normalized CLIP L/14 image embeddings using \cite{ldm} as the base model architecture. Some example generations using the publicly available LAION prior and our trained LDM is shown in the first column of Fig.\ref{fig:image_intro} and Appendix Sec.\ref{sec:appendix-results}. Further details about the HDM architecture are provided in Appendix Sec.\ref{sec:appendix-hdm}. We describe the modifications for domain specific and color conditional prior in this section.

\subsection{Domain Specific Prior}
\label{subsec:proposed-subspace}
For each of the domain specific priors, we retain the HDM pipeline as is but train a separate \emph{Diffusion Prior} model for each domain. We first obtain a curated internal dataset of images with \emph{texture}, suitable for \emph{vectors} and are of \emph{isolated} objects or concepts as described in Sec.\ref{sec:dataset}. Then, we train a domain specific \emph{Diffusion Prior} model as described in Appendix Sec.\ref{sec:appendix-hdm} on the curated dataset following the same setup as LAION prior. In contrast to the base prior model $\mathcal{P}_{\theta}$, the domain specific models $\mathcal{P}_{\theta1}^{texture}$, $\mathcal{P}_{\theta2}^{vector}$ and $\mathcal{P}_{\theta3}^{isolated}$ generate image embeddings within a specific sub-space in the CLIP embedding space that corresponds to their respective domain. Embeddings from each of the priors can be visualized by the LDM to generate domain specific images. Note that the domain specific priors only differ by the data they were trained on. There is no additional domain specific information, conditional input or finetuning. 

To better understand the advantages of the domain specific prior, we provide a conceptual illustration in Fig.\ref{fig:block_diagram} (Right). In VectorFusion \cite{vectorfusion}, a sample generated by a diffusion model or a random SVG is further optimized to be within the desirable domain specific space (red dotted arrows). Our domain specific priors directly generate desirable domain specific images by generating CLIP embeddings within the desirable subspace highlighted by the intersecting areas (generation path shown by black arrows from the domain priors). This is because, the domain specific prior models are trained only on CLIP embeddings of images from the specific domain. This also ensures the priors do not generate embeddings outside this area even for complex prompts.
\subsection{Conditional Prior}
\label{subsec:proposed-color}
There are many studies on style, shape and semantic map conditioned generations (ref. Sec.\ref{subsec:related-color}) but little to no work that focuses on color conditioned text-to-image generation. Being able to condition image generations with a color palette directly or obtained from another exemplar image is helpful for many creative workflows. We modify the \emph{Diffusion Prior} $\mathcal{P_{\theta}}$ described in Appendix Sec.\ref{appendix:proposed-prior} to take as input another token that represents the desired color information. Formally, our conditional Diffusion Prior is $\mathcal{P}_{\theta4}(z_x | y, c)$ where $c$ corresponds to the color information. If $z_c$ corresponds to the color representation, we simply add it as an additional token to $z_y$ as $[z_c, z_t, w_{1}, w_{2}, ..., w_n]$ utilized by the prior by cross attention. \\
\noindent\textbf{Color Histogram}:
Studies of color distribution for images are generally seen in the form 3D histograms \cite{ColourMapping, CT2001}, color triads \cite{shugrina2020color} or 2D Histograms such as log-chroma space \cite{colorconstancy, SIIE, afifi2021histogan}. In our experiments, we use 3D color histograms proposed in \cite{motiian2022multi} as $z_c$ obtained from the ground truth image $x$. The reason to use the LAB space over log-chroma or RGB space is because it is perceptually closer with respect to human color vision as its perceptual distances corresponds to Euclidean distances \cite{motiian2022multi}. We choose a 10$\times$8$\times$9 histogram and zero-pad to get the total dimension to 768 corresponding to CLIP L/14 text embedding and encodings. We take the square root of the histogram before passing it in the prior to make it more uniform. We also follow classifier free guidance approach \cite{classifier_free_guidance} during training by dropping color histogram with $p$ 0.5 to better capture textual content and relevance and both color and text with $p$ 0.1 for unconditional training. Color histogram is never provided without text to prevent content dependency on color information for generation. 

\section{Experimental Setup}
\subsection{Dataset}
\label{sec:dataset}
We use an internal 
dataset to train the prior and decoder models. \\
\noindent\textbf{Data for Decoder:} To train the \emph{Diffusion Decoder} LDM as described in Appendix Sec.\ref{appendix:proposed-ldm}, we remove images that contain humans or texts as detected using classifiers to reduce the size and complexity of the dataset to 77M images. The classifiers are a single linear layer on top of frozen CLIP L/14 embeddings. Only the linear layer is trained. We tested other complicated architecture but this worked reasonably well for all cases. The details about the classifiers are provided in Appendix Sec.\ref{sec:appendix-data-decoder}. \\
\noindent\textbf{Data for Priors:} For the domain specific priors, we train separate classifiers to detect the domains from images and use them to gather data. For color prior, we use a subset of text image pairs from the 77M filtered data. More information in Appendix Sec.\ref{sec:appendix-data-prior}.

\subsection{Training and Inference Details}
\label{sec:trainingdetails}
Our \emph{Diffusion Decoder} LDM has the same number of parameters as Stable Diffusion but trained on a relatively smaller dataset for fewer GPU hours as shown in Table.\ref{tab:training_specs}. We can also see that the prior models have significantly lower number of parameters and training time, measured in A100-40GB hours similar to \cite{ldm}. All the prior models are trained from scratch in 8-GPU A100-40GB instances. We train a smaller prior for vectors as we observed reasonable performance with a smaller model. The priors are trained with cosine schedule and all parameters are similar to those used in LAION prior\textsuperscript{\ref{laionprior}} while the LDM uses linear schedule with 1000 DDPM timesteps for training and uses the same architecture as \cite{ldm}. 

The dataset size and training time for all the models are provided in Table.\ref{tab:training_specs}. We emphasize that training the prior model from scratch takes lesser time and compute compared to finetuning the larger decoder or stable diffusion for the same application. For reference, \cite{controlnet} trains only a small part of stable diffusion model and uses an average of 600 A100-80GB GPU-hours for canny-edge conditioned finetuning of Stable Diffusion model on 3M images. In comparison, our isolated object prior model takes 1344 A100-40GB hours to train on a dataset of 20M image-text pairs. If we use a conservative 1.5$\times$ factor speed up from A100-40GB to A100-80GB, this translates to 896 A100-80GB GPU hours for 20M samples. If we also scale the training data with another conservative 6$\times$ factor to get closer to 3M samples, we get 150 A1000-80GB GPU hours which is significantly less despite the differences in conditioning input. Similarly, the color prior with same calculations get to around 112 GPU hours on A100-80GB compared to 150-600 GPU hours for various conditioning inputs based finetuning in \cite{controlnet}. 

For all our experiments we use 100 DDIM steps for sampling the CLIP embedding from \emph{Diffusion Prior} and 50 DDIM steps to generate an image using the \emph{Diffusion Decoder}. We use linear sampling schedule. To improve relevance in the prior, we generate 10 embeddings per prompt and choose the embedding with highest CLIP score to input text prompt as done in \cite{dalle2}. For all qualitative examples, we use the same random seed per prompt across all baselines.

\begin{table}[t]
\caption{Training details: Compute measured in A100 40GB GPU-hours, number of parameters and cardinality of the dataset use to train all the \emph{Diffusion Prior} models and LDM \emph{Diffusion Decoder} compared with those of Stable Diffusion.}
\vspace{0.1in}
          \begin{tabular}{lcccc}
          \toprule
          Models & Compute & Nparams & Data Size \\
           & (A100 hours)  & (Million)  & (Million)\\
          \midrule
          Isolated Prior & 1344 & 249.22 & 20 \\

          Vector Prior & 1680 & 101.76  & 26\\

          Texture Prior & 576  & 249.22 & 10\\

          Color Prior & 3072  & 249.28 & 61\\
        
          Our LDM & 117600 & 859.52 & 77\\

          Stable Diffusion & 150000 & 859.52 & 2000\\
          \bottomrule
          \end{tabular}
          \vspace{-0.2in}
    \label{tab:training_specs}
\end{table}

\subsection{Baselines}
\label{sec:baseline}
We represent our method in the results as \lq{}ours\rq{}. This corresponds to the HDM pipeline consisting of the prior and LDM. Unless otherwise specified, both prior and LDM are the models we trained as described in Sec.\ref{sec:trainingdetails}.
\subsubsection{Domain Specific Generation}
\label{sec:domainbaseline}
For qualitative examples we randomly sample prompts from our test set and use the following methods to generate images for visual comparison. For quantitative experiments, we compare our method with Stable Diffusion baselines. 
\begin{figure*}[h]
 \centering
 \footnotesize
 \begin{tabular}{c@{\hspace{0.1cm}}c@{\hspace{0.1cm}}c@{\hspace{0.1cm}}c@{\hspace{0.1cm}}c@{\hspace{0.1cm}}c@{\hspace{0.1cm}}c@{\hspace{0.1cm}}c@{\hspace{0.1cm}}c@{\hspace{0.1cm}}c@{\hspace{0.1cm}}}
 \centering
 \vspace{0.05in}
Ours& SD &SD-prompt & LAION & LAION-prompt&Ours& SD & SD-prompt & LAION & LAION-prompt \\ 
 \includegraphics[height=0.65in, width=0.65in]{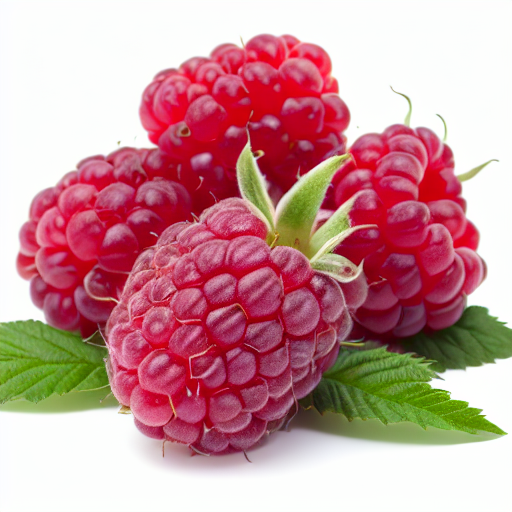}&
 \includegraphics[height=0.65in, width=0.65in]{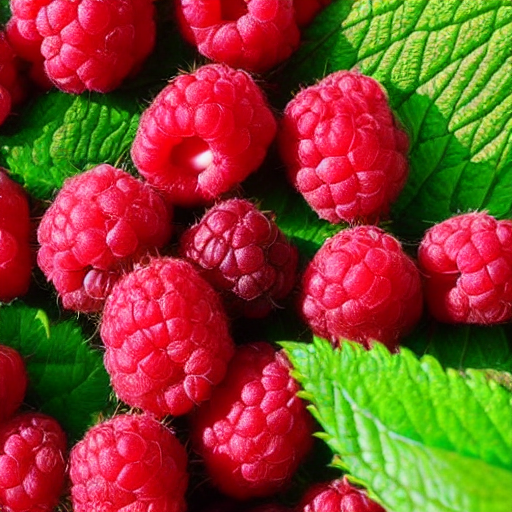}&
 \includegraphics[height=0.65in, width=0.65in]{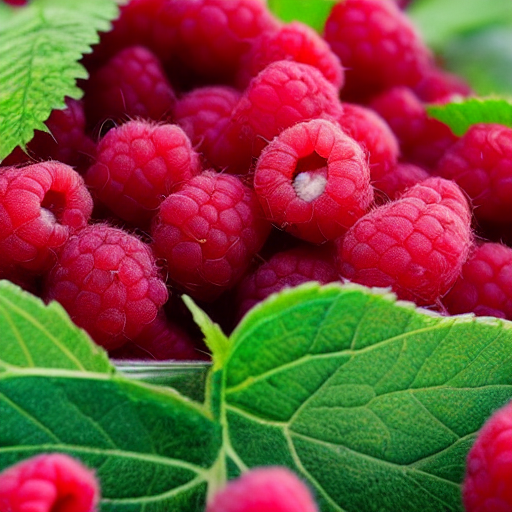}&
 \includegraphics[height=0.65in, width=0.65in]{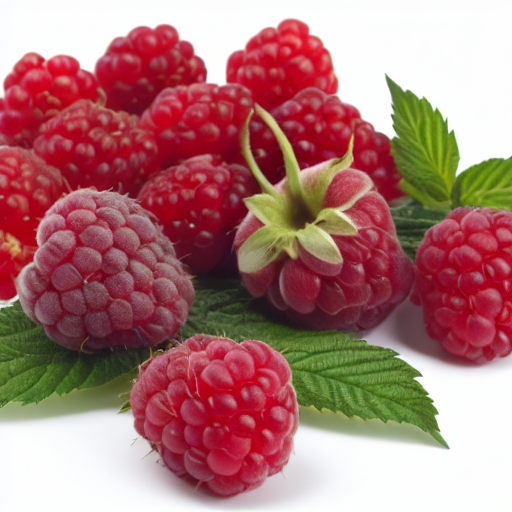}&
 \includegraphics[height=0.65in, width=0.65in]{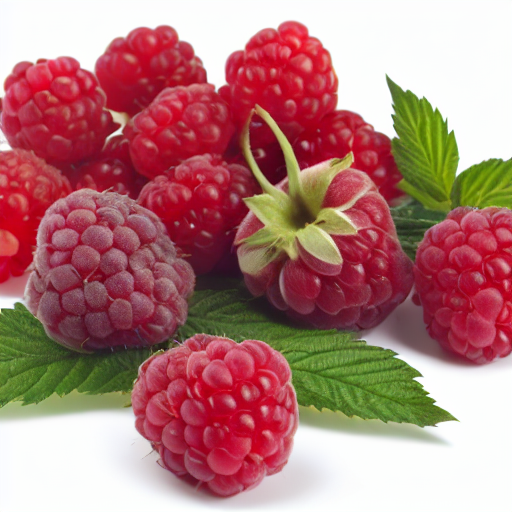} &
 \includegraphics[height=0.65in, width=0.65in]{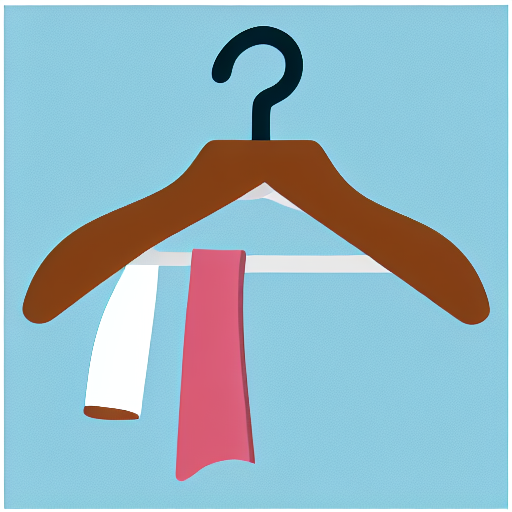}&
 \includegraphics[height=0.65in, width=0.65in]{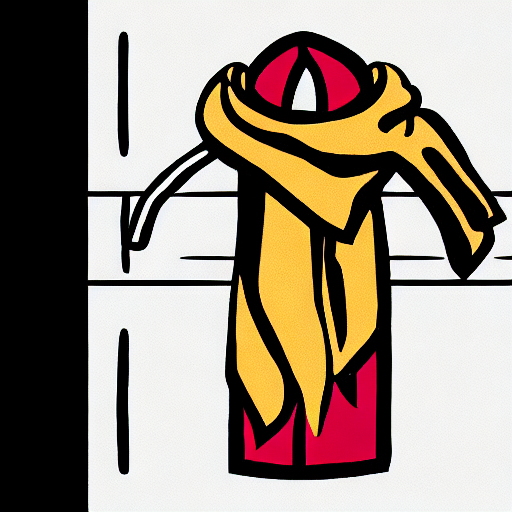}&
 \includegraphics[height=0.65in, width=0.65in]{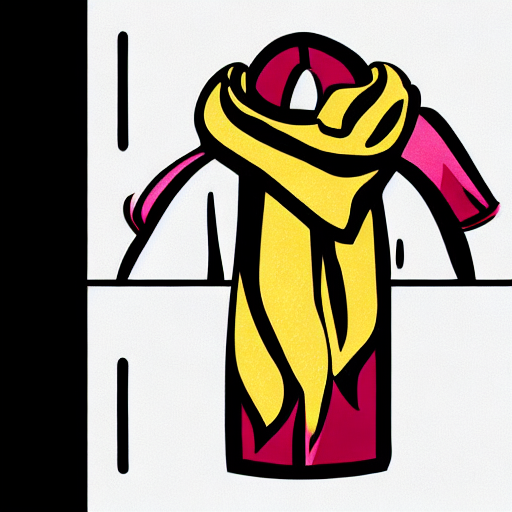}&
 \includegraphics[height=0.65in, width=0.65in]{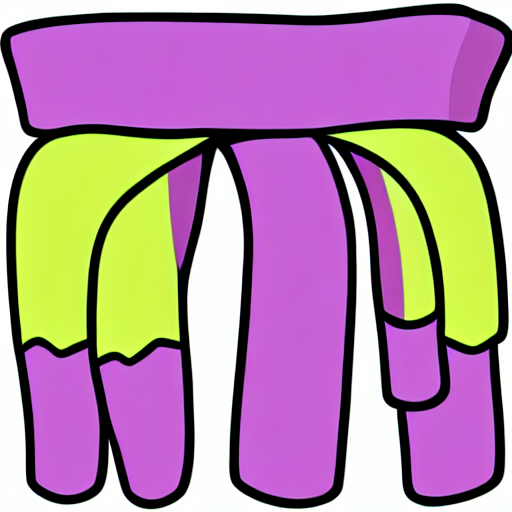}&
 \includegraphics[height=0.65in, width=0.65in]{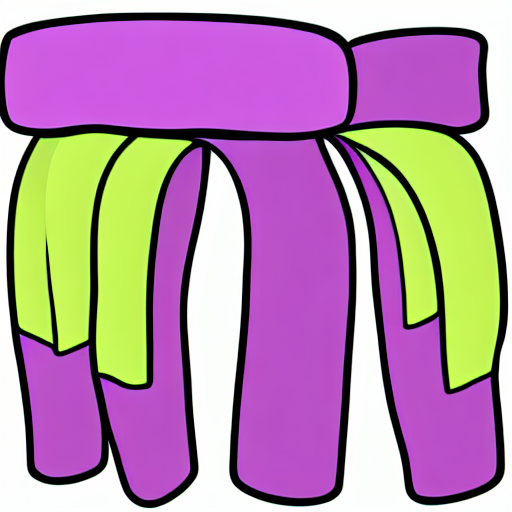} \\
  \multicolumn{5}{c}{\emph{Text prompt: "heap of ripe raspberries with leaves"}}&\multicolumn{5}{c}{\emph{Text prompt: "scarf on coat-hanger cartoon icon"}}\\ \\
 \includegraphics[height=0.65in, width=0.65in]{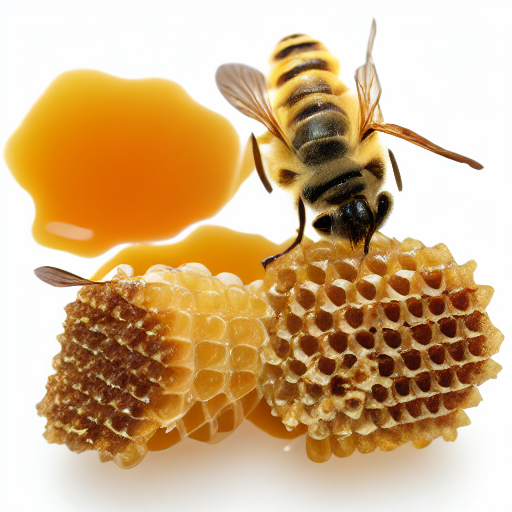}&
 \includegraphics[height=0.65in, width=0.65in]{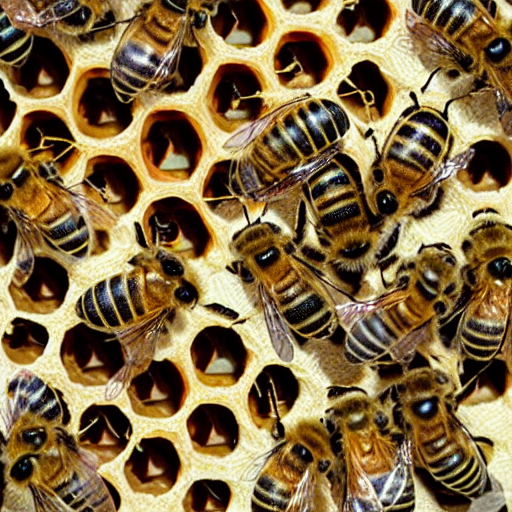}&
 \includegraphics[height=0.65in, width=0.65in]{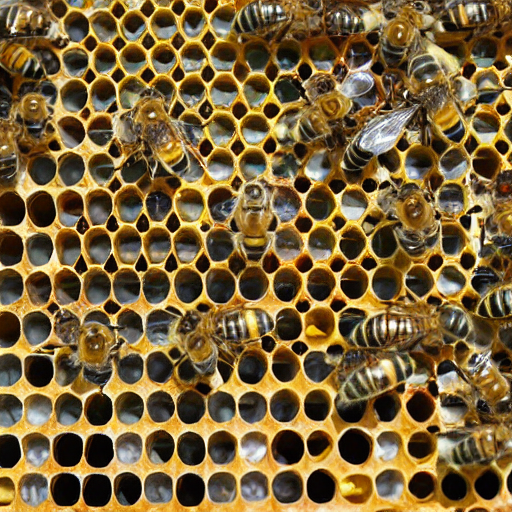}&
 \includegraphics[height=0.65in, width=0.65in]{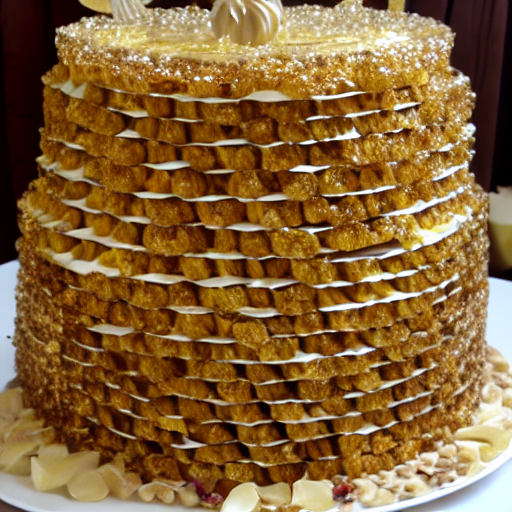}&
 \includegraphics[height=0.65in, width=0.65in]{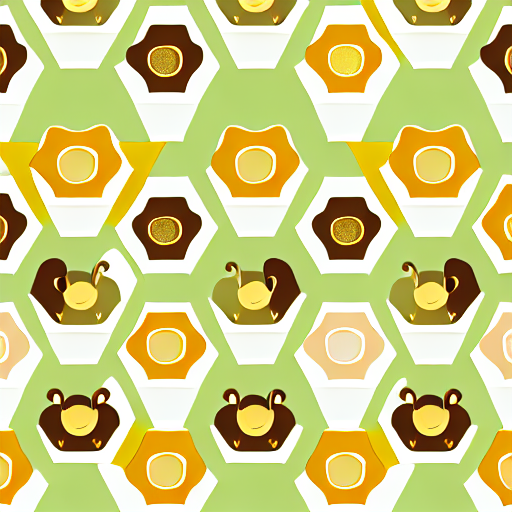} &
\includegraphics[height=0.65in, width=0.65in]{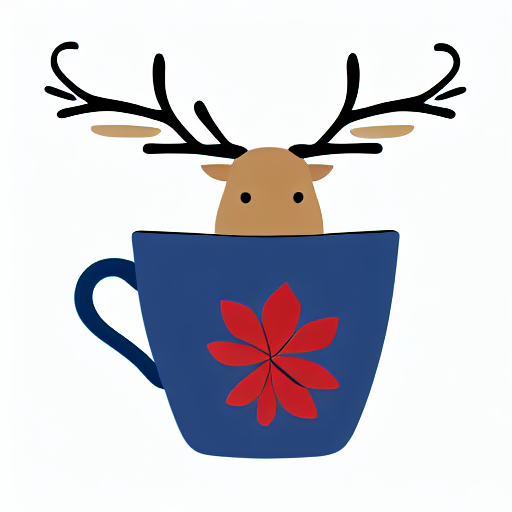}&
 \includegraphics[height=0.65in, width=0.65in]{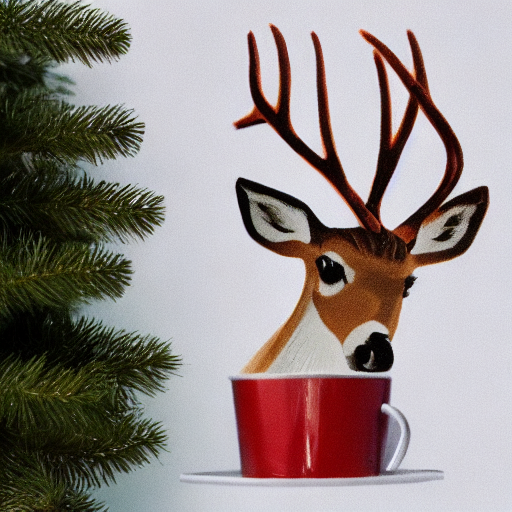}&
 \includegraphics[height=0.65in, width=0.65in]{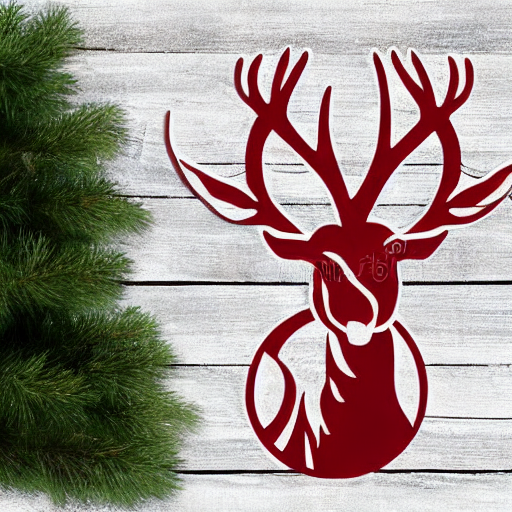}&
 \includegraphics[height=0.65in, width=0.65in]{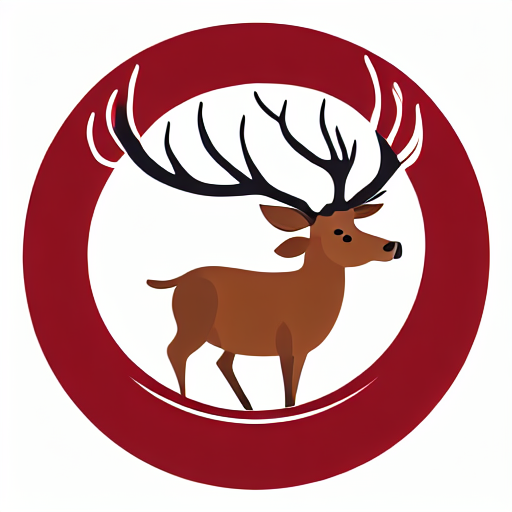}&
 \includegraphics[height=0.65in, width=0.65in]{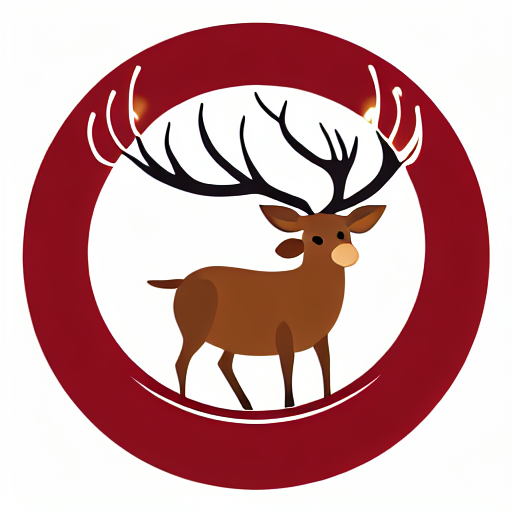}\\ 
 \multicolumn{5}{c}{\emph{Text prompt: "bees on honeycomb"}}&\multicolumn{5}{c}{\emph{Text prompt: "deer with spruce branches in a red cup in a blue circle"}}\\ \\
 \includegraphics[height=0.65in, width=0.65in]{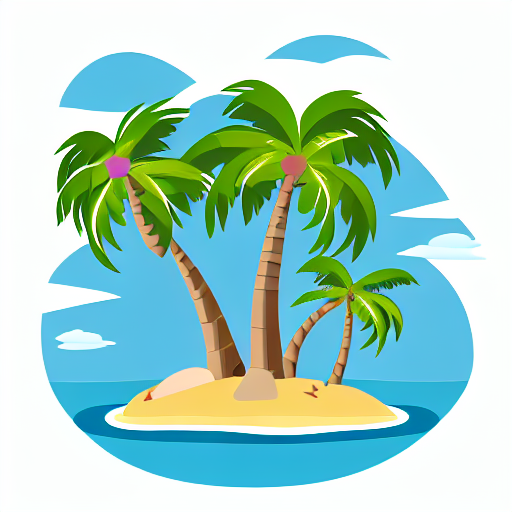}&
 \includegraphics[height=0.65in, width=0.65in]{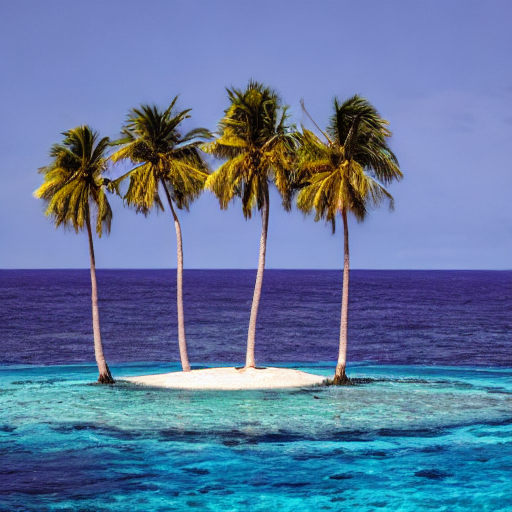}&
 \includegraphics[height=0.65in, width=0.65in]{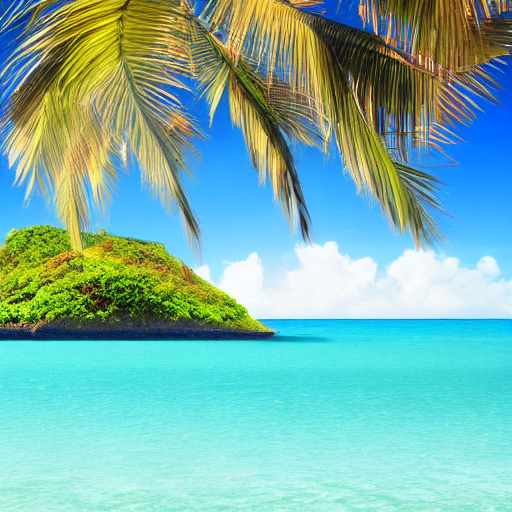}&
 \includegraphics[height=0.65in, width=0.65in]{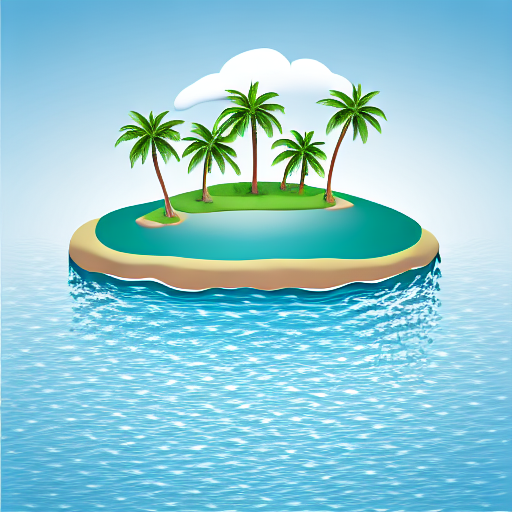}&
 \includegraphics[height=0.65in, width=0.65in]{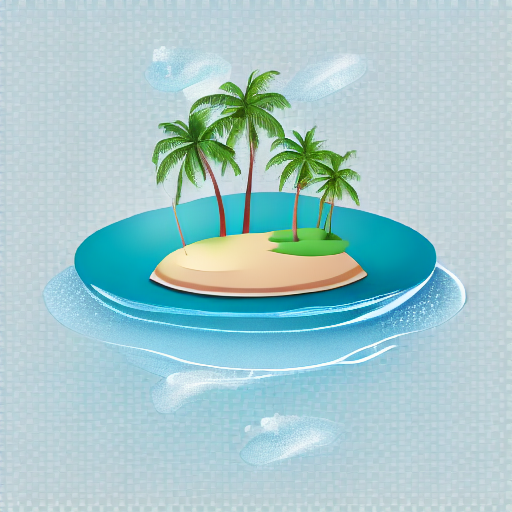} &
 \includegraphics[height=0.65in, width=0.65in]{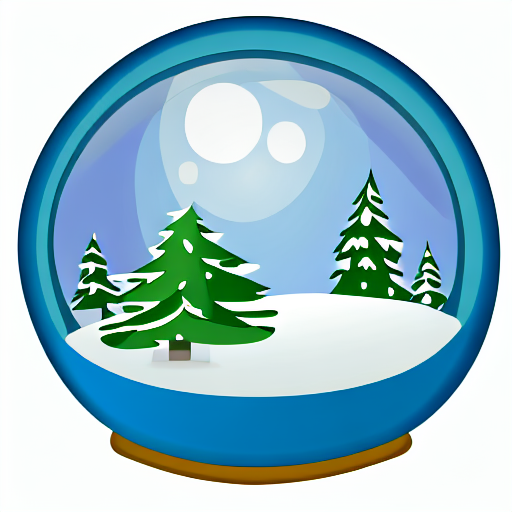}&
 \includegraphics[height=0.65in, width=0.65in]{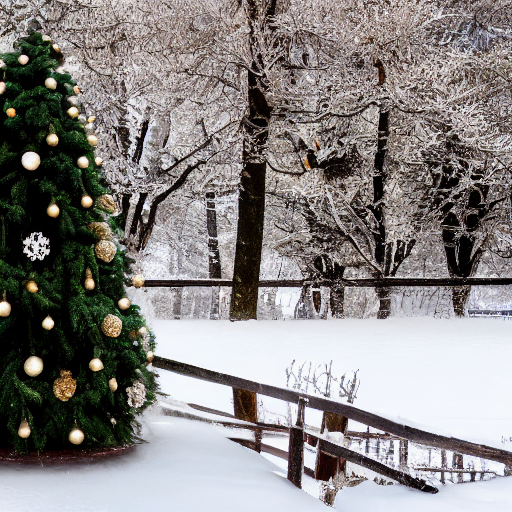}&
 \includegraphics[height=0.65in, width=0.65in]{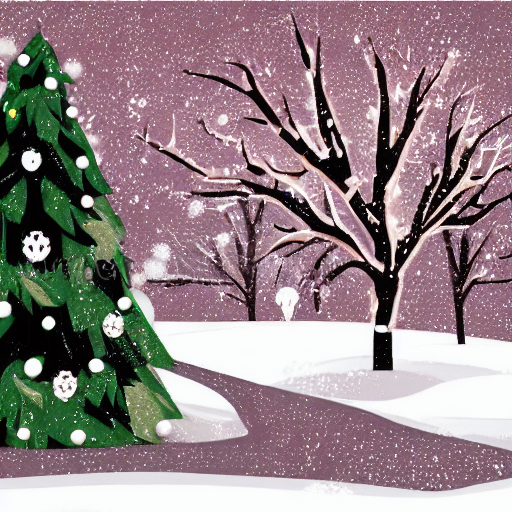}&
 \includegraphics[height=0.65in, width=0.65in]{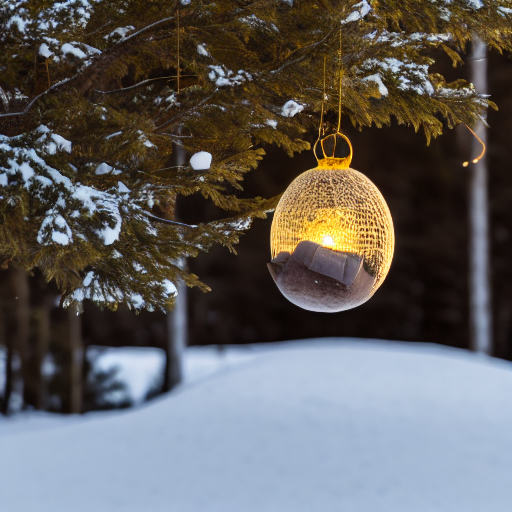}&
 \includegraphics[height=0.65in, width=0.65in]{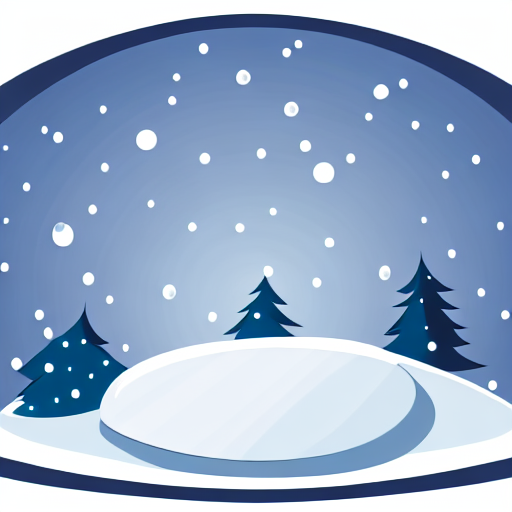} \\
 \multicolumn{5}{c}{\emph{Text prompt: "Island in the middle of the ocean with palm trees"}}&\multicolumn{5}{c}{\emph{Text prompt: "a winter landscape in a christmas ball"}} \\ \\
\end{tabular}

 \caption{Qualitative examples showing the effectiveness of isolated background domain specific prior (left) and the rasterized vector prior (right) compared with baseline models and prompt engineering.}
 \label{fig:domain-isol_quant}
 \vspace{-0.1in}
\end{figure*}

\begin{figure}[h]
 \centering
 \footnotesize
 \begin{tabular}{c@{\hspace{0.1cm}}c@{\hspace{0.1cm}}c@{\hspace{0.1cm}}c@{\hspace{0.1cm}}c@{\hspace{0.1cm}}}
 \centering
 \vspace{0.05in} 
 Ours& SD & SD-prompt &LAION& LAION-prompt\\
 \includegraphics[height=0.6in, width=0.6in]{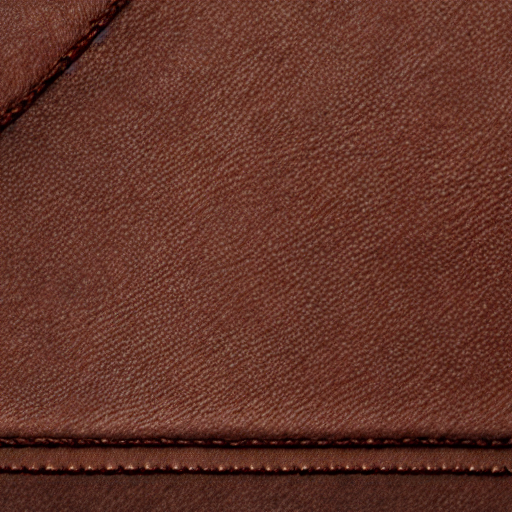}&
 \includegraphics[height=0.6in, width=0.6in]{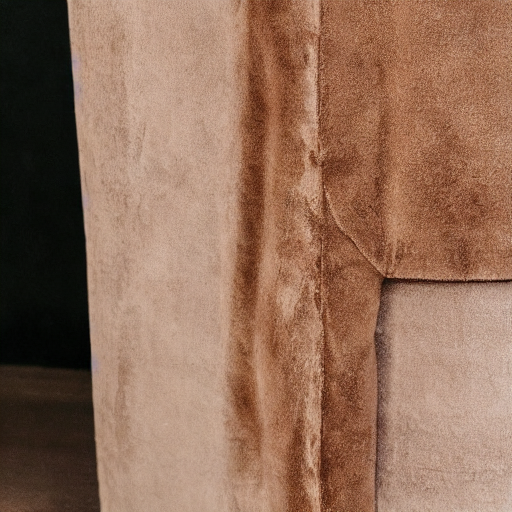}&
 \includegraphics[height=0.6in, width=0.6in]{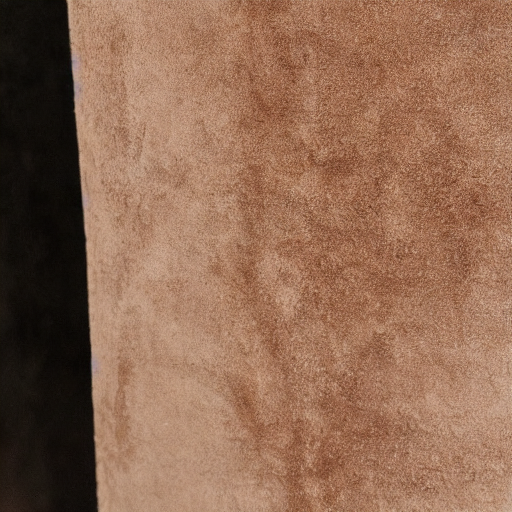}&
 \includegraphics[height=0.6in, width=0.6in]{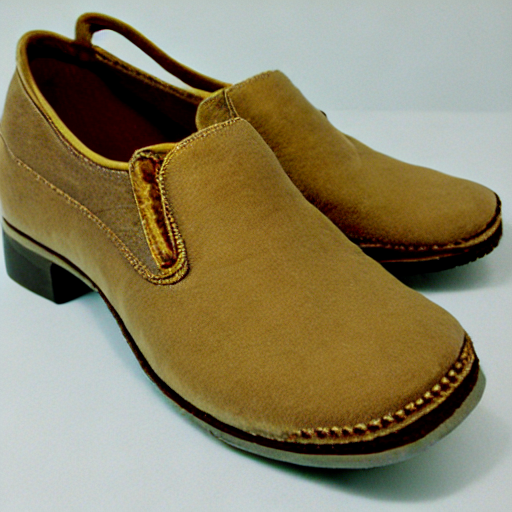}&
 \includegraphics[height=0.6in, width=0.6in]{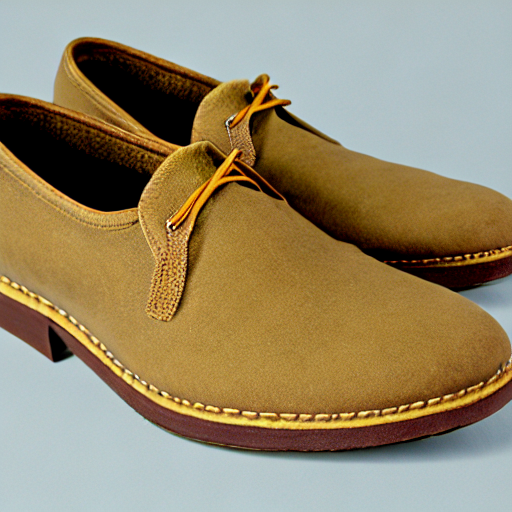} \\
 \multicolumn{5}{c}{\emph{Text prompt: "brown suede detail"}} \\ 
 \includegraphics[height=0.6in, width=0.6in]{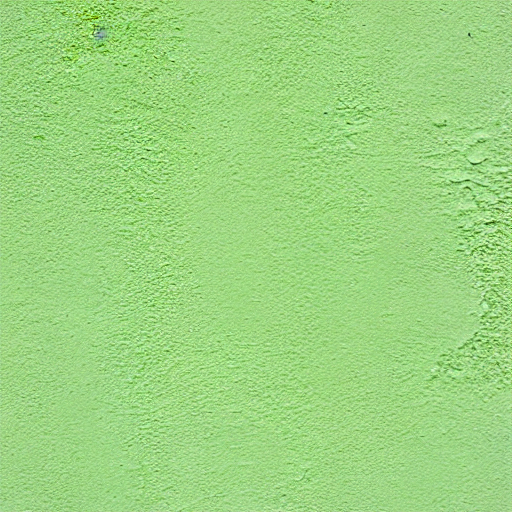}&
 \includegraphics[height=0.6in, width=0.6in]{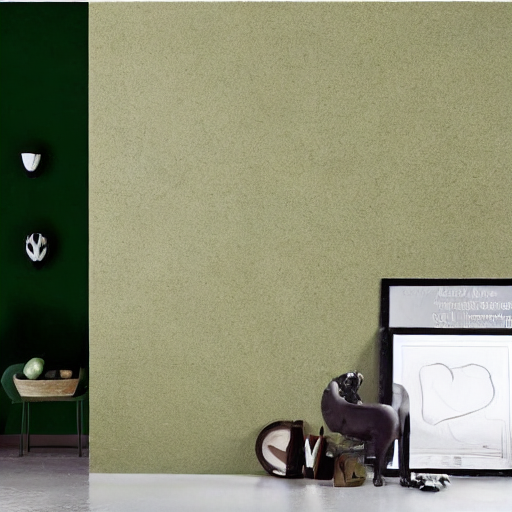}&
 \includegraphics[height=0.6in, width=0.6in]{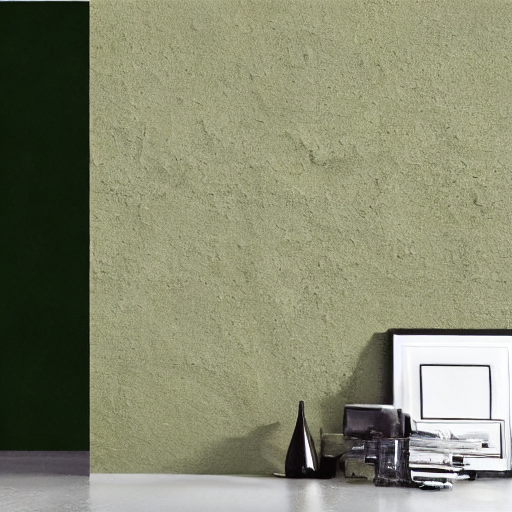}&
 \includegraphics[height=0.6in, width=0.6in]{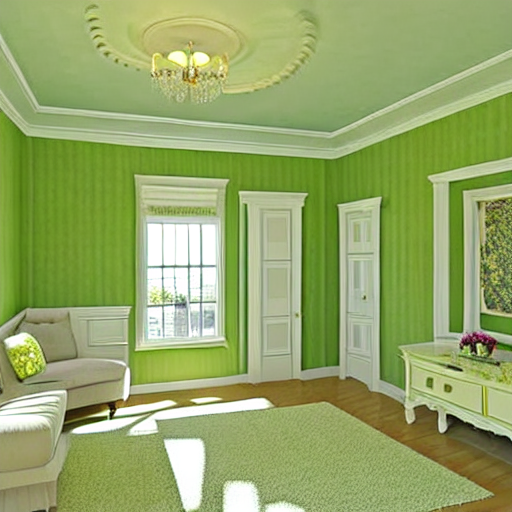}&
 \includegraphics[height=0.6in, width=0.6in]{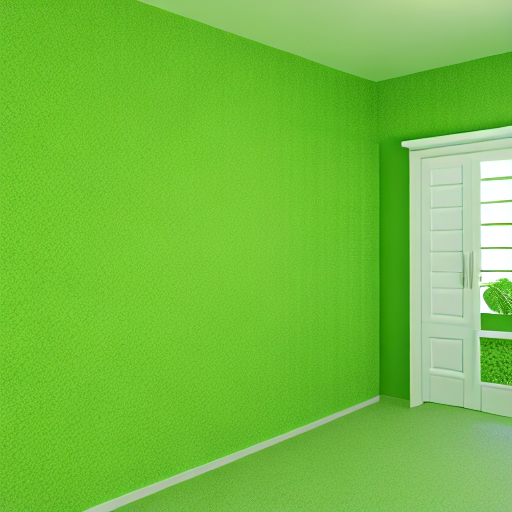} \\ 
 \multicolumn{5}{c}{\emph{Text prompt: "green glossy plaster wall paint"}} \\ 
 \includegraphics[height=0.6in, width=0.6in]{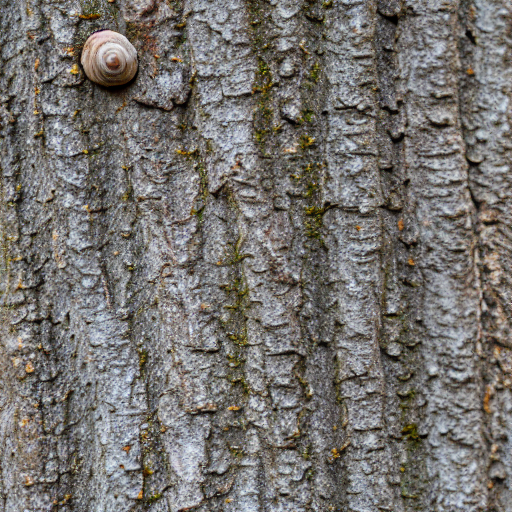}&
 \includegraphics[height=0.6in, width=0.6in]{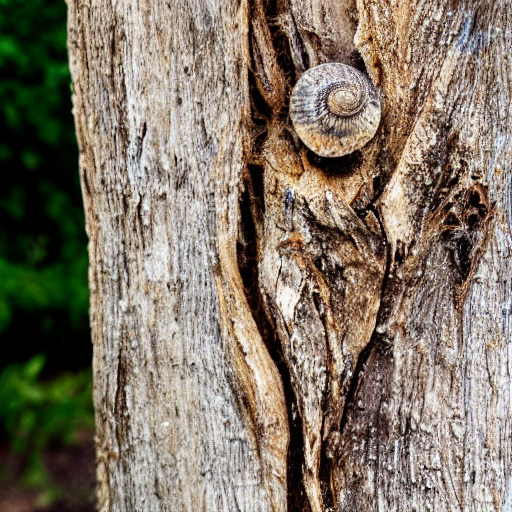}&
 \includegraphics[height=0.6in, width=0.6in]{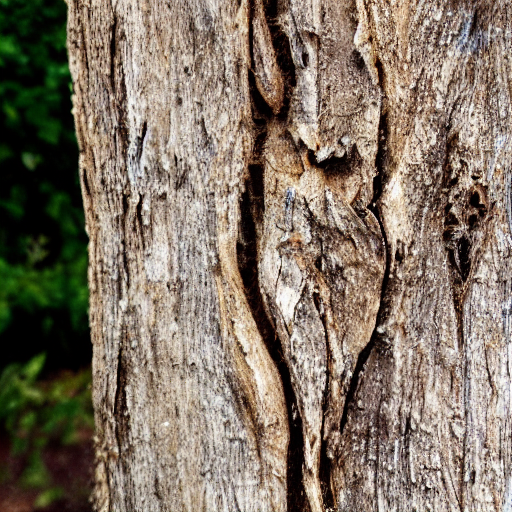}&
 \includegraphics[height=0.6in, width=0.6in]{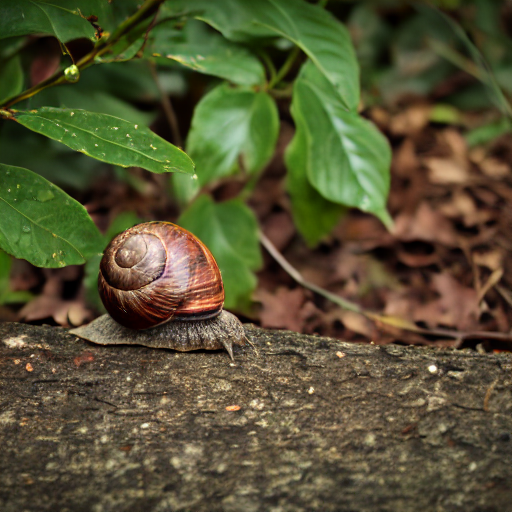}&
 \includegraphics[height=0.6in, width=0.6in]{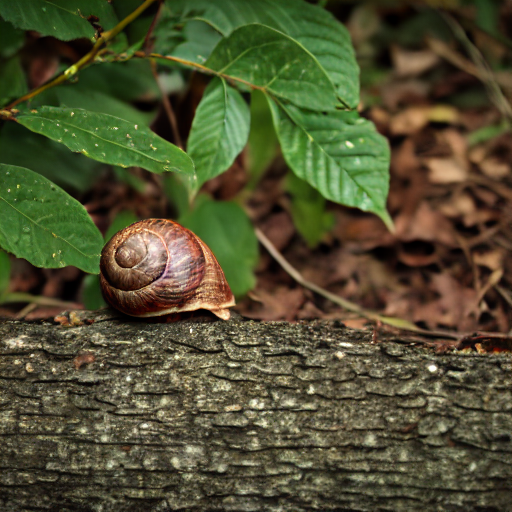} \\
 \multicolumn{5}{c}{\emph{Text prompt: "a texture of snail climbing a tree bark"}} \\ 
\end{tabular}
 \caption{Qualitative examples showing the effectiveness of texture domain specific prior compared to existing models.}
 \label{fig:domain-texture_quant}
 \vspace{-0.1in}
\end{figure}

\noindent\textbf{Stable Diffusion (SD):}
We use Stable Diffusion v1.4 \cite{ldm} for comparison as a baseline. \\
\noindent\textbf{Stable Diffusion with Suffix (SD-prompt):}
Existing methods like \cite{vectorfusion} use prompt engineering with stable diffusion to get images closer to the desirable space. Similarly, we add \lq{}texture background\rq{}, \lq{}vector illustration\rq{} and \lq{}isolated background\rq{} as suffix to prompts to compare with texture, vector and isolated priors. \\
\noindent\textbf{Baseline Prior (LAION):}
As a baseline for the domain \emph{Diffusion Prior} models, we use pretrained and publicly available LAION prior \footnote{https://github.com/LAION-AI/conditioned-prior} with our trained LDM \emph{Diffusion Decoder}.\\
\noindent\textbf{Baseline Prior with Suffix (LAION-prompt):}
We also use the suffixes mentioned above with the LAION prior to support the baseline further.

\subsubsection{Color Conditional Generation}
\label{sec:colorbaseline}
To the best of our knoweldge, there aren't existing techniques that perform color conditioned text to image generation (excluding concurrent works). We hence use popular color transfer techniques on images generated by Stable Diffusion for comparison. \\
\noindent\textbf{WCT-RGB \cite{WCT}:} We combine \cite{WCT} that does style transfer using covariance matrix matching on an intermediate deep space with a traditional mean/std matching way \cite{CT2001} of performing color transfer on RGB color space. This leads to covariance matrix matching on RGB color space. \\
\noindent\textbf{CTHA \cite{Lee2020CTHA}:} This method encodes color histograms of reference color image and generated image using an encoder and repeatedly fuses into features of a U-Net \cite{unet} decoder network. This method also uses semantic segmentations of the images to perform color transfer and therefore, is semantic aware. We use \cite{HRNet} to generate the semantic segmentations. \\
\noindent\textbf{ReHistoGAN \cite{afifi2021histogan}:} This is primarily a noise to image generator model which works by injecting color histogram of reference color image into StyleGAN\cite{stylegan}. The model is trained to transfer color while keeping the image realistic and preserving the input image's structure. We use the Universal ReHistoGAN model for our comparisons. 

\subsection{Metrics}
\label{sec:metrics}
For all quantitative results, for each domain, we use 5000 random prompts from their respective test set to generate images for comparison.\\
\noindent\textbf{Quality:} To measure quality of the generated image and its alignment with the training distribution, we use the Frechet
inception distance (FID) \cite{parmar2021cleanfid} metric. For all priors, we use ground truth images corresponding to 5000 prompts from test set as the real distribution. The test set is different for different domains (ref. Appendix Sec.\ref{sec:appendix-data-prior}). \\
\noindent\textbf{Prompt Relevance:} To measure relevance of the generated image to input text prompt irrespective of the domain or conditional input used, we use the CLIP \cite{clip} score.\\
\noindent\textbf{Domain Relevance:} To further measure relevance to the specific domain of generation, we use our domain classifiers mentioned in \ref{sec:appendix-data-prior} to get average confidence over all generated images from the test set. \\
\noindent\textbf{Color Relevance:} To measure the relevance of color palette in the generated image to the exemplar image, we use Hellinger distance and KL divergence on the color histograms\cite{motiian2022multi} from both the images. 

\begin{figure*}[t]
    \centering
    \includegraphics[width=\linewidth]{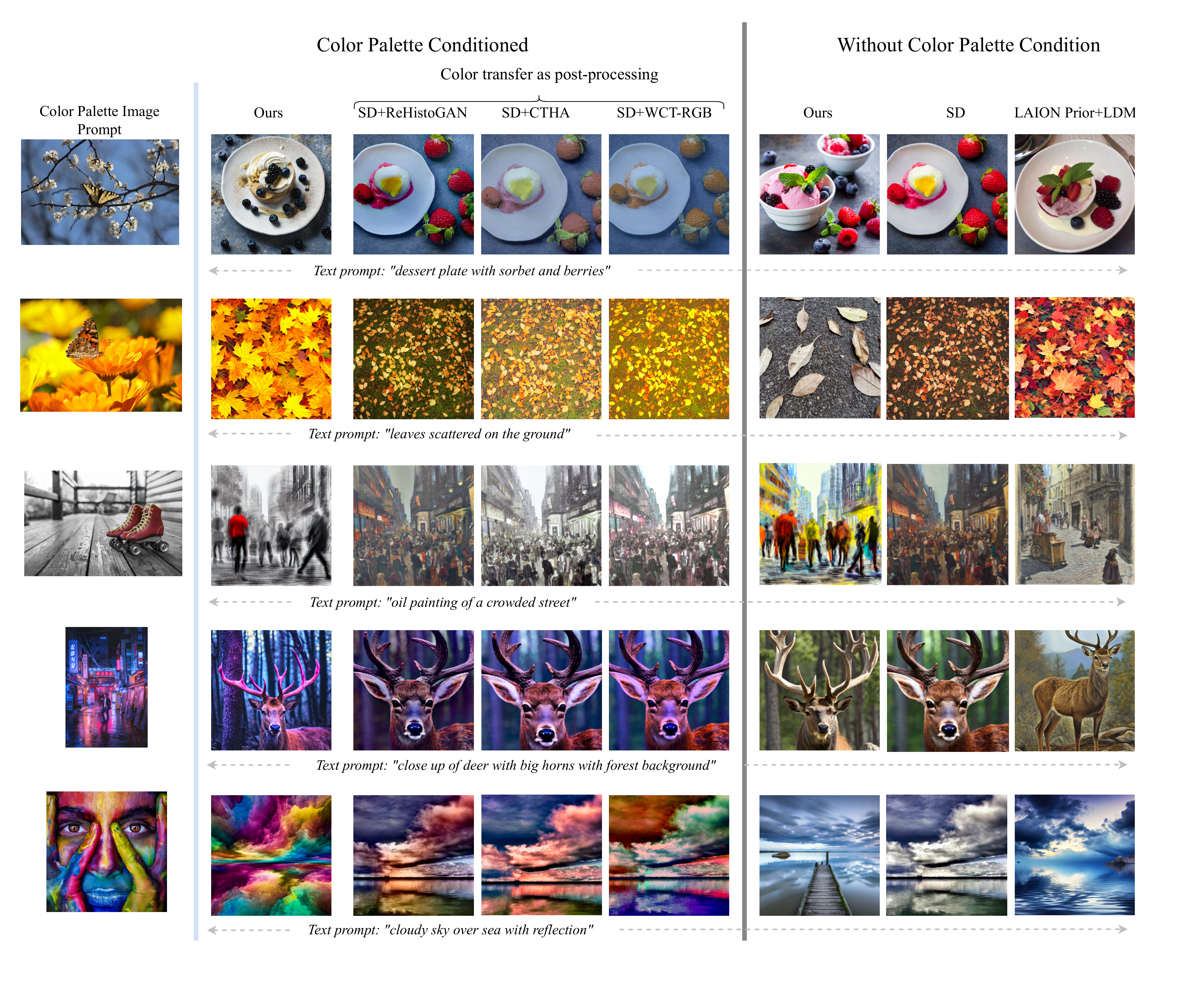}
    \caption{Qualitative analysis of our Conditional prior approach. Images for each prompt have been generated using the same seed value.}
    \label{fig:color_qual}
    \vspace{-0.2in}
\end{figure*}
\section{Results and Analysis}
\label{sec:results}
\subsection{Domain Specific Generation}

We can see from Table.\ref{tab:domain_prior_quant} that our method outperforms existing large pretrained Stable Diffusion across all metrics including quality and relevance. Moreover, adding domain specific modifiers to the prompts though helps a little for vector and textures in improving domain relevance measured by the classifiers, it is significantly lower compared to using the domain specific prior. The FID scores also show that the generated images from the proposed technique are of higher quality and are more relevant to the specific domain's real distribution. 

We also show some qualitative examples in Fig.\ref{fig:domain-isol_quant} and \ref{fig:domain-texture_quant} for all the domain specific priors and generation. We can observe that irrespective of the complexity of the prompt, the generated image using our method does not generate out of domain images. The LDM and the common CLIP space ensure relevance to prompt and generalization across concepts whereas the domain specific priors ensures the images are within the desired domain. For example, \lq{}bees on honeycomb\rq{} generates the background as a honeycomb in Stable Diffusion and LAION prior baseline whereas our model generated relevant image while maintaining the white background. Similarly, prompts such as \lq{}snail climbing on a tree\rq{} or \lq{} a winter landscape in a christmas ball\rq{} bias the baseline models towards content based generation even with additional prompt engineering. However, the domain prior results in relevant images within the desired domain. 

\begin{table}[t]
\caption{Quantitative analysis of the domain specific prior models compared with stable diffusion baselines. Clf.score corresponds to the average confidence for positive class from the domain specific classifiers described in Sec.\ref{sec:appendix-data-prior}. CLIP is the CLIP Score between text prompt and generated image while FID is the Frechet Inception Distance metric as described in Sec.\ref{sec:metrics}}\vspace{0.1in}
  \begin{tabular}{p{0.25\linewidth}p{0.2\linewidth}p{0.15\linewidth}p{0.15\linewidth}}
  \hline
  \toprule
  Method &Clf.Score $\uparrow$&CLIP $\uparrow$ & FID $\downarrow$ \\
  \midrule
  \multicolumn{4}{c}{Isolated Background Domain} \\
  \midrule
  SD & 0.268 & \textbf{0.267} & 35.257  \\

  SD-prompt & 0.250 & 0.264 & 38.569  \\
\rowcolor[gray]{.9}
  Ours & \textbf{0.496} & 0.265 & \textbf{25.795}\\
    \midrule
  \multicolumn{4}{c}{Vector Domain} \\
    \midrule
  SD & 0.550 & 0.247 & 58.200  \\

  SD-prompt & 0.761 & 0.245 & 69.344  \\
\rowcolor[gray]{.9}
  Ours & \textbf{0.950} & \textbf{0.248}  & \textbf{21.600}\\
    \midrule
  \multicolumn{4}{c}{Texture Domain} \\
    \midrule
  SD & 0.663 & 0.258 & 40.559  \\

  SD-prompt & 0.788 & 0.255 &  40.284 \\
\rowcolor[gray]{.9}
  Ours & \textbf{0.860} & \textbf{0.261} & \textbf{35.524}\\
 \bottomrule
  \end{tabular}
    \label{tab:domain_prior_quant}
\end{table}



\subsection{Color Conditional Generation}
\label{sec:resultscolot}
Quantitative results for the color prior model comparing with baselines is provided in Table \ref{tab: color_prior_quant}. We observe a clear trade off between color transfer and quality of the images generated in existing baselines. For example, SD+WCTRGB shows the best performance in color transfer as measured by Hellinger distance and KL divergence metrics but has the least quality in generation. Since WCTRGB is a statistical histogram matching based algorithm, it ensures the color palette of the transferred image matches with the exemplar while sacrificing quality. Similarly, SD+ReHistoGAN shows least performance in color transfer with better FID score relative to the proposed model showing conservative color transfers. Our methods strikes the right balance and ensures the generated images are of high quality while also having color relevant to the exemplar image. Interestingly, all color transfer methods degrade the quality of the baseline Stable Diffusion generated image showing absence of semantic awareness. 
\begin{table}[h]
\caption{Quantitative analysis of the proposed color conditional prior compared with existing color transfer methods on images generated by Stable Diffusion. H dist. corresponds to Hellinger's distance while KL div indicates KL divergence metric between generated image's color histogram and that of the exemplar image.}\vspace{0.1in}
          \begin{tabular}{lccc}
          \toprule
          Method & $\downarrow$ H dist. & $\downarrow$ KL div. &  $\downarrow$ FID \\
          \midrule
          SD & - & - & 20.613 \\

           SD+WCTRGB & 0.468 & 1.779 & 23.616\\

          SD+ReHistoGAN & 0.566 & 4.704 & 21.299\\

          SD+CTHA & 0.496 & 2.330 & 21.950\\

          Ours with zero cond & - & - & 22.144 \\
        \rowcolor[gray]{.9}
          Ours &  0.480 & 3.164 & 21.670\\
          \bottomrule
          \end{tabular}
    \label{tab: color_prior_quant}
\end{table}
We can make similar observations from the qualitative examples provided in Fig.\ref{fig:color_qual}. The proposed technique relates the input text prompt and color histogram generating a realistic possible image visualizing the inputs. Our model generates images with objects and concepts that realistically fit the color distribution of the exemplar image. Berries in the first row become \emph{blueberries} because of the input exemplar color image (compare with \emph{raspberries} seen in the generation without color palette). Though SD+WCTRGB performed better with color transfer metrics, the images generated for all examples shown in Fig.\ref{fig:color_qual} are unrealistic and have a global hue that corresponds to the color palette applied over an existing image. Though this is mainly due to the color transfer being applied on an existing image of \emph{strawberries} from Stable Diffusion, it shows the inability of these models as a plug-in for color conditioned text to image generation.

\section{Conclusion}
In this paper, we show the advantages of having a common CLIP embedding space in the text to image generation pipeline and the effectiveness of the \emph{Diffusion Prior} model trained on the CLIP embedding space. We show that the prior model is smaller in memory and requires significantly less time to be trained on a specific desired domain of text-image pairs. Once trained, this prior model can be combined with an existing large decoder model to generate domain specific images. Since the prior model has not seen embeddings outside the domain specific subspace in the CLIP embedding space, it avoids generating out of domain images and is robust to complex prompts. Prompt Engineering on existing diffusion models show lower performance and finetuning such large models even with just few layers or optimization is expensive in compute and memory. 

We also show that the \emph{Diffusion Prior} model can be trained to accept additional conditioning input (color histogram in our paper) to generate images that align with the input text prompt as well as the color palette without losing on semantics or quality. We compare our domain specific and color conditioned priors qualitatively and quantitatively with existing baselines and show on par or better performance across all metrics and domains.

We hope that this paper opens up new research and possibilities into the HDM architecture and the capabilities of a small latent space diffusion model like the \emph{Diffusion Prior}.



{\small
\bibliographystyle{ieee_fullname}
\bibliography{egbib}

\begin{thebibliography}{10}\itemsep=-1pt

\bibitem{SIIE}
Mahmoud Afifi and Michael~S. Brown.
\newblock Sensor-independent illumination estimation for dnn models.
\newblock 2019.

\bibitem{afifi2021histogan}
Mahmoud Afifi, Marcus~A. Brubaker, and Michael~S. Brown.
\newblock Histogan: Controlling colors of gan-generated and real images via
  color histograms.
\newblock In {\em Proceedings of the IEEE Conference on Computer Vision and
  Pattern Recognition}, 2021.

\bibitem{ImageRecoloring}
Mahmoud Afifi, Brian Price, Scott Cohen, and Michael Brown.
\newblock Image recoloring based on object color distributions.
\newblock 05 2019.

\bibitem{colorconstancy}
Mahmoud Afifi, Brian Price, Scott Cohen, and Michael~S. Brown.
\newblock When color constancy goes wrong: Correcting improperly white-balanced
  images.
\newblock In {\em 2019 IEEE/CVF Conference on Computer Vision and Pattern
  Recognition (CVPR)}, pages 1535--1544, 2019.

\bibitem{ediff_i}
Yogesh Balaji, Seungjun Nah, Xun Huang, Arash Vahdat, Jiaming Song, Karsten
  Kreis, Miika Aittala, Timo Aila, Samuli Laine, Bryan Catanzaro, Tero Karras,
  and Ming-Yu Liu.
\newblock ediff-i: Text-to-image diffusion models with an ensemble of expert
  denoisers.
\newblock {\em \href{https://arxiv.org/abs/2211.01324} {arXiv:2211.01324}},
  2022.

\bibitem{multidiffusion}
Omer Bar-Tal, Lior Yariv, Yaron Lipman, and Tali Dekel.
\newblock Multidiffusion: Fusing diffusion paths for controlled image
  generation, 2023.

\bibitem{texture-3d}
Toby~P Breckon and Robert~B Fisher.
\newblock A hierarchical extension to 3d non-parametric surface relief
  completion.
\newblock {\em Pattern Recognition}, 45(1):172--185, 2012.

\bibitem{instructpix2pix}
Tim Brooks, Aleksander Holynski, and Alexei~A. Efros.
\newblock Instructpix2pix: Learning to follow image editing instructions, 2022.

\bibitem{muse}
Huiwen Chang, Han Zhang, Jarred Barber, AJ Maschinot, Jose Lezama, Lu Jiang,
  Ming-Hsuan Yang, Kevin Murphy, William~T. Freeman, Michael Rubinstein,
  Yuanzhen Li, and Dilip Krishnan.
\newblock Muse: Text-to-image generation via masked generative transformers.
\newblock 2023.

\bibitem{Text2Colors}
Wonwoong Cho, Hyojin Bahng, David~Keetae Park, Seungjoo Yoo, Ziming Wu,
  Xiaojuan Ma, and Jaegul Choo.
\newblock Text2colors: Guiding image colorization through text-driven palette
  generation.
\newblock {\em CoRR}, abs/1804.04128, 2018.

\bibitem{early-texture1}
Jeremy~S De~Bonet.
\newblock Multiresolution sampling procedure for analysis and synthesis of
  texture images.
\newblock In {\em Proceedings of the 24th annual conference on Computer
  graphics and interactive techniques}, pages 361--368, 1997.

\bibitem{diffusion_beat_gans}
Prafulla Dhariwal and Alex Nichol.
\newblock Diffusion models beat gans on image synthesis.
\newblock {\em \href{https://arxiv.org/abs/2105.05233}{arXiv:2105.05233}},
  2021.

\bibitem{early-texture2}
Alexei~A Efros and Thomas~K Leung.
\newblock Texture synthesis by non-parametric sampling.
\newblock In {\em Proceedings of the seventh IEEE international conference on
  computer vision}, volume~2, pages 1033--1038. IEEE, 1999.

\bibitem{vqgan}
Patrick Esser, Robin Rombach, and Björn Ommer.
\newblock Taming transformers for high-resolution image synthesis.
\newblock {\em \href{https://arxiv.org/abs/2012.09841}{arXiv:2012.09841}},
  2020.

\bibitem{ColourMapping}
H.~Sheikh Faridul, T. Pouli, C. Chamaret, J. Stauder, E. Reinhard, D. Kuzovkin,
  and A. Tremeau.
\newblock {Colour Mapping: A Review of Recent Methods, Extensions and
  Applications}.
\newblock {\em Computer Graphics Forum}, 2016.

\bibitem{fernando2021generative}
Chrisantha Fernando, SM Eslami, Jean-Baptiste Alayrac, Piotr Mirowski, Dylan
  Banarse, and Simon Osindero.
\newblock Generative art using neural visual grammars and dual encoders.
\newblock {\em arXiv preprint arXiv:2105.00162}, 2021.

\bibitem{vectordraw}
Kevin Frans, Lisa~B Soros, and Olaf Witkowski.
\newblock Clipdraw: Exploring text-to-drawing synthesis through language-image
  encoders.
\newblock {\em arXiv preprint arXiv:2106.14843}, 2021.

\bibitem{texture2}
Anna Frühstück, Ibraheem Alhashim, and Peter Wonka.
\newblock Tilegan: Synthesis of large-scale non-homogeneous textures.
\newblock {\em {ACM} Transactions on Graphics}, 38(4):1--11, jul 2019.

\bibitem{textual_inversion}
Rinon Gal, Yuval Alaluf, Yuval Atzmon, Or Patashnik, Amit~H. Bermano, Gal
  Chechik, and Daniel Cohen-Or.
\newblock An image is worth one word: Personalizing text-to-image generation
  using textual inversion.
\newblock 2022.

\bibitem{gan}
Ian~J. Goodfellow, Jean Pouget-Abadie, Mehdi Mirza, Bing Xu, David
  Warde-Farley, Sherjil Ozair, Aaron Courville, and Yoshua Bengio.
\newblock Generative adversarial networks.
\newblock {\em \href{https://arxiv.org/abs/1406.2661}{arXiv:1406.2661}}, 2014.

\bibitem{deepExemplar}
Mingming He, Dongdong Chen, Jing Liao, Pedro~V. Sander, and Lu Yuan.
\newblock Deep exemplar-based colorization, 2018.

\bibitem{early-texture3}
David~J Heeger and James~R Bergen.
\newblock Pyramid-based texture analysis/synthesis.
\newblock In {\em Proceedings of the 22nd annual conference on Computer
  graphics and interactive techniques}, pages 229--238, 1995.

\bibitem{prompt2prompt}
Amir Hertz, Ron Mokady, Jay Tenenbaum, Kfir Aberman, Yael Pritch, and Daniel
  Cohen-Or.
\newblock Prompt-to-prompt image editing with cross attention control.
\newblock {\em arXiv preprint arXiv:2208.01626}, 2022.

\bibitem{ddpm}
Jonathan Ho, Ajay Jain, and Pieter Abbeel.
\newblock Denoising diffusion probabilistic models.
\newblock {\em \href{https://arxiv.org/abs/2006.11239}{arXiv:2006.11239}},
  2020.

\bibitem{classifier_free_guidance}
Jonathan Ho and Tim Salimans.
\newblock Classifier-free diffusion guidance.
\newblock {\em \href{https://arxiv.org/abs/2207.12598} {arXiv:2207.12598}},
  2022.

\bibitem{DAST}
Kibeom Hong, Seogkyu Jeon, Huan Yang, Jianlong Fu, and Hyeran Byun.
\newblock Domain-aware universal style transfer, 2021.

\bibitem{composer}
Lianghua Huang, Di Chen, Yu Liu, Yujun Shen, Deli Zhao, and Jingren Zhou.
\newblock Composer: Creative and controllable image synthesis with composable
  conditions, 2023.

\bibitem{DiffStyler}
Nisha Huang, Yuxin Zhang, Fan Tang, Chongyang Ma, Haibin Huang, Yong Zhang,
  Weiming Dong, and Changsheng Xu.
\newblock Diffstyler: Controllable dual diffusion for text-driven image
  stylization, 2022.

\bibitem{vectorascent}
Ajay Jain.
\newblock Vectorascent: Generate vector graphics from a textual description,
  2021.

\bibitem{vectorfusion}
Ajay Jain, Amber Xie, and Pieter Abbeel.
\newblock Vectorfusion: Text-to-svg by abstracting pixel-based diffusion
  models, 2022.

\bibitem{k_diffusion}
Tero Karras, Miika Aittala, Timo Aila, and Samuli Laine.
\newblock Elucidating the design space of diffusion-based generative models,
  2022.

\bibitem{stylegan}
Tero Karras, Samuli Laine, Miika Aittala, Janne Hellsten, Jaakko Lehtinen, and
  Timo Aila.
\newblock Analyzing and improving the image quality of stylegan, 2019.

\bibitem{imagic}
Bahjat Kawar, Shiran Zada, Oran Lang, Omer Tov, Huiwen Chang, Tali Dekel, Inbar
  Mosseri, and Michal Irani.
\newblock Imagic: Text-based real image editing with diffusion models, 2022.

\bibitem{custom_diffusion}
Nupur Kumari, Bingliang Zhang, Richard Zhang, Eli Shechtman, and Jun-Yan Zhu.
\newblock Multi-concept customization of text-to-image diffusion.
\newblock 2022.

\bibitem{Lee2020CTHA}
Junyong Lee, Hyeongseok Son, Gunhee Lee, Jonghyeop Lee, Sunghyun Cho, and
  Seungyong Lee.
\newblock Deep color transfer using histogram analogy.
\newblock {\em The Visual Computer}, 36(10):2129--2143, 2020.

\bibitem{WCT}
Yijun Li, Chen Fang, Jimei Yang, Zhaowen Wang, Xin Lu, and Ming-Hsuan Yang.
\newblock Universal style transfer via feature transforms, 2017.

\bibitem{MagicMix}
Jun~Hao Liew, Hanshu Yan, Daquan Zhou, and Jiashi Feng.
\newblock Magicmix: Semantic mixing with diffusion models, 2022.

\bibitem{pndm}
Luping Liu, Yi Ren, Zhijie Lin, and Zhou Zhao.
\newblock Pseudo numerical methods for diffusion models on manifolds.
\newblock {\em \href{https://arxiv.org/abs/arXiv:2202.09778}
  {arXiv:2202.09778}}, 2022.

\bibitem{dpm_solver_pp}
Cheng Lu, Yuhao Zhou, Fan Bao, Jianfei Chen, Chongxuan Li, and Jun Zhu.
\newblock Dpm-solver++: Fast solver for guided sampling of diffusion
  probabilistic models.
\newblock {\em \href{https://arxiv.org/abs/arXiv:2211.01095}
  {arXiv:2211.01095}}, 2022.

\bibitem{repaint}
Andreas Lugmayr, Martin Danelljan, Andres Romero, Fisher Yu, Radu Timofte, and
  Luc~Van Gool.
\newblock Repaint: Inpainting using denoising diffusion probabilistic models.
\newblock 2022.

\bibitem{live}
Xu Ma, Yuqian Zhou, Xingqian Xu, Bin Sun, Valerii Filev, Nikita Orlov, Yun Fu,
  and Humphrey Shi.
\newblock Towards layer-wise image vectorization.
\newblock In {\em Proceedings of the IEEE/CVF Conference on Computer Vision and
  Pattern Recognition}, pages 16314--16323, 2022.

\bibitem{sdedit}
Chenlin Meng, Yutong He, Yang Song, Jiaming Song, Jiajun Wu, Jun-Yan Zhu, and
  Stefano Ermon.
\newblock Sdedit: Guided image synthesis and editing with stochastic
  differential equations.
\newblock 2021.

\bibitem{distillation_guided_diffusion}
Chenlin Meng, Robin Rombach, Ruiqi Gao, Diederik~P. Kingma, Stefano Ermon,
  Jonathan Ho, and Tim Salimans.
\newblock On distillation of guided diffusion models.
\newblock {\em \href{https://arxiv.org/abs/arXiv:2210.03142}
  {arXiv:2210.03142}}, 2022.

\bibitem{clipclop}
Piotr Mirowski, Dylan Banarse, Mateusz Malinowski, Simon Osindero, and
  Chrisantha Fernando.
\newblock Clip-clop: Clip-guided collage and photomontage.
\newblock {\em arXiv preprint arXiv:2205.03146}, 2022.

\bibitem{motiian2022multi}
Saeid Motiian, Zhe Lin, Samarth Gulati, Pramod Srinivasan, Jose
  Ignacio~Echevarria Vallespi, and Baldo~Antonio Faieta.
\newblock Multi-resolution color-based image search, Jan.~4 2022.
\newblock US Patent 11,216,505.

\bibitem{t2iadapt}
Chong Mou, Xintao Wang, Liangbin Xie, Jian Zhang, Zhongang Qi, Ying Shan, and
  Xiaohu Qie.
\newblock T2i-adapter: Learning adapters to dig out more controllable ability
  for text-to-image diffusion models, 2023.

\bibitem{improved_diffusion}
Alex Nichol and Prafulla Dhariwal.
\newblock Improved denoising diffusion probabilistic models.
\newblock {\em \href{https://arxiv.org/abs/2102.09672}{arXiv:2102.09672}},
  2021.

\bibitem{glide}
Alex Nichol, Prafulla Dhariwal, Aditya Ramesh, Pranav Shyam, Pamela Mishkin,
  Bob McGrew, Ilya Sutskever, and Mark Chen.
\newblock Glide: Towards photorealistic image generation and editing with
  text-guided diffusion models.
\newblock {\em \href{https://arxiv.org/abs/2112.10741}{arXiv:2112.10741}},
  2021.

\bibitem{my_style_gan}
Yotam Nitzan, Kfir Aberman, Qiurui He, Orly Liba, Michal Yarom, Yossi
  Gandelsman, Inbar Mosseri, Yael Pritch, and Daniel Cohen-or.
\newblock Mystyle: A personalized generative prior.
\newblock 2022.

\bibitem{parmar2021cleanfid}
Gaurav Parmar, Richard Zhang, and Jun-Yan Zhu.
\newblock On aliased resizing and surprising subtleties in gan evaluation.
\newblock In {\em CVPR}, 2022.

\bibitem{dreamfusion}
Ben Poole, Ajay Jain, Jonathan~T Barron, and Ben Mildenhall.
\newblock Dreamfusion: Text-to-3d using 2d diffusion.
\newblock {\em arXiv preprint arXiv:2209.14988}, 2022.

\bibitem{clip}
Alec Radford, Jong~Wook Kim, Chris Hallacy, Aditya Ramesh, Gabriel Goh,
  Sandhini Agarwal, Girish Sastry, Amanda Askell, Pamela Mishkin, Jack Clark,
  Gretchen Krueger, and Ilya Sutskever.
\newblock Learning transferable visual models from natural language
  supervision.
\newblock 2021.

\bibitem{t5}
Colin Raffel, Noam Shazeer, Adam Roberts, Katherine Lee, Sharan Narang, Michael
  Matena, Yanqi Zhou, Wei Li, and Peter~J. Liu.
\newblock Exploring the limits of transfer learning with a unified text-to-text
  transformer.
\newblock 2019.

\bibitem{dalle2}
Aditya Ramesh, Prafulla Dhariwal, Alex Nichol, Casey Chu, and Mark Chen.
\newblock Hierarchical text-conditional image generation with clip latents.
\newblock {\em \href{https://arxiv.org/abs/2204.06125} {arXiv:2204.06125}},
  2022.

\bibitem{CT2001}
Erik Reinhard, Michael Ashikhmin, Bruce Gooch, and Peter Shirley.
\newblock Color transfer between images.
\newblock {\em IEEE Computer Graphics and Applications}, 21:34--41, 10 2001.

\bibitem{texture-finetune}
Elad Richardson, Gal Metzer, Yuval Alaluf, Raja Giryes, and Daniel Cohen-Or.
\newblock Texture: Text-guided texturing of 3d shapes, 2023.

\bibitem{ldm}
Robin Rombach, Andreas Blattmann, Dominik Lorenz, Patrick Esser, and Björn
  Ommer.
\newblock High-resolution image synthesis with latent diffusion models.
\newblock {\em \href{https://arxiv.org/abs/2112.10752}{arXiv:2112.10752}},
  2021.

\bibitem{unet}
Olaf Ronneberger, Philipp Fischer, and Thomas Brox.
\newblock U-net: Convolutional networks for biomedical image segmentation,
  2015.

\bibitem{dreambooth}
Nataniel Ruiz, Yuanzhen Li, Varun Jampani, Yael Pritch, Michael Rubinstein, and
  Kfir Aberman.
\newblock Dreambooth: Fine tuning text-to-image diffusion models for
  subject-driven generation.
\newblock 2022.

\bibitem{palette}
Chitwan Saharia, William Chan, Huiwen Chang, Chris~A. Lee, Jonathan Ho, Tim
  Salimans, David~J. Fleet, and Mohammad Norouzi.
\newblock Palette: Image-to-image diffusion models.
\newblock {\em CoRR}, abs/2111.05826, 2021.

\bibitem{imagen}
Chitwan Saharia, William Chan, Saurabh Saxena, Lala Li, Jay Whang, Emily
  Denton, Seyed Kamyar~Seyed Ghasemipour, Burcu~Karagol Ayan, S.~Sara Mahdavi,
  Rapha~Gontijo Lopes, Tim Salimans, Jonathan Ho, David~J Fleet, and Mohammad
  Norouzi.
\newblock Photorealistic text-to-image diffusion models with deep language
  understanding.
\newblock 2022.

\bibitem{distillation_diffusion}
Tim Salimans and Jonathan Ho.
\newblock Progressive distillation for fast sampling of diffusion models.
\newblock {\em \href{https://arxiv.org/abs/arXiv:2202.00512}
  {arXiv:2202.00512}}, 2022.

\bibitem{styleclipdraw}
Peter Schaldenbrand, Zhixuan Liu, and Jean Oh.
\newblock Styleclipdraw: Coupling content and style in text-to-drawing
  translation.
\newblock {\em arXiv preprint arXiv:2202.12362}, 2022.

\bibitem{texture3}
Omry Sendik and Daniel Cohen-Or.
\newblock Deep correlations for texture synthesis.
\newblock {\em ACM Trans. Graph.}, jul 2017.

\bibitem{shugrina2020color}
Maria Shugrina, Amlan Kar, Karan Singh, and Sanja Fidler.
\newblock Nonlinear color triads for approximation, learning and direct
  manipulation of color distributions.
\newblock 2020.

\bibitem{make_a_video}
Uriel Singer, Adam Polyak, Thomas Hayes, Xi Yin, Jie An, Songyang Zhang, Qiyuan
  Hu, Harry Yang, Oron Ashual, Oran Gafni, Devi Parikh, Sonal Gupta, and Yaniv
  Taigman.
\newblock Make-a-video: Text-to-video generation without text-video data.
\newblock 2022.

\bibitem{dickstein_2015}
Jascha Sohl-Dickstein, Eric~A. Weiss, Niru Maheswaranathan, and Surya Ganguli.
\newblock Deep unsupervised learning using nonequilibrium thermodynamics, 2015.

\bibitem{ddim}
Jiaming Song, Chenlin Meng, and Stefano Ermon.
\newblock Denoising diffusion implicit models.
\newblock {\em \href{https://arxiv.org/abs/2010.02502}{arXiv:2010.02502}},
  2020.

\bibitem{HRNet}
Ke Sun, Yang Zhao, Borui Jiang, Tianheng Cheng, Bin Xiao, Dong Liu, Yadong Mu,
  Xinggang Wang, Wenyu Liu, and Jingdong Wang.
\newblock High-resolution representations for labeling pixels and regions,
  2019.

\bibitem{transformer}
Ashish Vaswani, Noam Shazeer, Niki Parmar, Jakob Uszkoreit, Llion Jones,
  Aidan~N. Gomez, Lukasz Kaiser, and Illia Polosukhin.
\newblock Attention is all you need, 2017.

\bibitem{PalGan}
Yi Wang, Menghan Xia, Lu Qi, Jing Shao, and Yu Qiao.
\newblock Palgan: Image colorization with palette generative adversarial
  networks, 2022.

\bibitem{dream3d}
Jiale Xu, Xintao Wang, Weihao Cheng, Yan-Pei Cao, Ying Shan, Xiaohu Qie, and
  Shenghua Gao.
\newblock Dream3d: Zero-shot text-to-3d synthesis using 3d shape prior and
  text-to-image diffusion models.
\newblock 2022.

\bibitem{WCT2}
Jaejun Yoo, Youngjung Uh, Sanghyuk Chun, Byeongkyu Kang, and Jung-Woo Ha.
\newblock Photorealistic style transfer via wavelet transforms, 2019.

\bibitem{parti}
Jiahui Yu, Yuanzhong Xu, Jing~Yu Koh, Thang Luong, Gunjan Baid, Zirui Wang,
  Vijay Vasudevan, Alexander Ku, Yinfei Yang, Burcu~Karagol Ayan, Ben
  Hutchinson, Wei Han, Zarana Parekh, Xin Li, Han Zhang, Jason Baldridge, and
  Yonghui Wu.
\newblock Scaling autoregressive models for content-rich text-to-image
  generation.
\newblock 2022.

\bibitem{controlnet}
Lvmin Zhang and Maneesh Agrawala.
\newblock Adding conditional control to text-to-image diffusion models.
\newblock {\em arXiv preprint arXiv:2302.05543}, 2023.

\bibitem{deis}
Qinsheng Zhang and Yongxin Chen.
\newblock Fast sampling of diffusion models with exponential integrator.
\newblock {\em \href{https://arxiv.org/abs/arXiv:2204.13902}
  {arXiv:2204.13902}}, 2022.

\bibitem{InvCreative}
Yuxin Zhang, Nisha Huang, Fan Tang, Haibin Huang, Chongyang Ma, Weiming Dong,
  and Changsheng Xu.
\newblock Inversion-based creativity transfer with diffusion models, 2022.

\bibitem{shifted_diffusion}
Yufan Zhou, Bingchen Liu, Yizhe Zhu, Xiao Yang, Changyou Chen, and Jinhui Xu.
\newblock Shifted diffusion for text-to-image generation.
\newblock 2022.

\bibitem{texture1}
Yang Zhou, Zhen Zhu, Xiang Bai, Dani Lischinski, Daniel Cohen-Or, and Hui
  Huang.
\newblock Non-stationary texture synthesis by adversarial expansion.
\newblock (4), 2018.

\end{thebibliography}
}
\clearpage

\appendix
\twocolumn[{%
\renewcommand\twocolumn[1][]{#1}%
\maketitle
\vspace*{-1cm}
\begin{center}
    \centering
    \vspace{0.5in}
    \Huge{Appendix}  
    \vspace{0.5in}
\end{center}}]

\normalsize
\section{Introduction}
\label{appendix:intro}
In this paper, we leverage the Diffusion Prior model trained to generate CLIP image embeddings conditioned on CLIP text embeddings for controlled and conditional text to image generation. We show that training the Diffusion Prior is memory and compute efficient since it works on a low dimensional latent space and is architecturally less complex compared to the larger decoder models. Moreover, training the prior on a dataset of just \emph{textures}, \emph{vectors} or \emph{isolated objects} is less expensive than finetuning larger models and results in more robust domain specific generations. The Diffusion Prior can also take additional conditional input like \emph{color} histogram to generate images from text, conditioned on a specific color palette. Since the prior generates valid CLIP embeddings, the generations are semantic aware and do not lose realism as shown qualitatively and quantitatively in the main paper.

In the Appendix, we start by discussing additional related works in Sec.\ref{sec:appendix-related}. We then give detailed information on the Diffusion Prior and Diffusion Decoder models that we trained in Sec.\ref{sec:appendix-hdm}. We describe the classifiers that were used to get domain specific data in Sec.\ref{sec:appendix-data-prior} and those used to simplify the data to train the Diffusion Decoder in Sec.\ref{sec:appendix-data-decoder}. Note that we use the same classifiers to also quantitatively measure if the generations are within a specific domain in Sec.\ref{sec:results}. We discuss limitations in Sec.\ref{sec:appendix-limitations} followed by additional examples of text to image generation and image variations using the trained HDM in Sec.\ref{sec:appendix-results}. We also provide additional results for domain specific generations for all the domains and color conditioned generation in Sec.\ref{sec:appendix-results}. 

\section{Related Works}
\label{sec:appendix-related}
Diffusion Models (DMs) are based on Gaussian denoising process \cite{dickstein_2015} which assumes that the noises added to the original images are drawn from Gaussian distributions. The denoising process involves predicting the added noises using a convolutional neural network called U-Net \cite{unet}. Compared to GANs, DMs are easier to train and scale. DMs have been shown to achieve state-of-the-art of image quality \cite{diffusion_beat_gans, ddpm}.
\subsection{Diffusion Models}
\label{sec:appendix-related-dm}
Interesting approaches to strengthen controllability as well as improve efficiency have been proposed in recent times. The classifier guidance method allows DMs to condition on the predictions of a classifier \cite{diffusion_beat_gans, glide} during sampling, whereas \cite{classifier_free_guidance} proposes classifier-free guidance that does not require pretrained classifiers. DMs can be conditioned on texts, images, or both \cite{ediff_i, muse, diffusion_beat_gans, classifier_free_guidance, improved_diffusion, glide, dalle2, ldm, imagen, parti}. These conditions are usually in the forms of embedding vectors from CLIP \cite{clip} or T5 \cite{t5} which are based on the Transformer architecture \cite{transformer}. DMs can also be applied to other computer vision tasks such as super-resolution \cite{glide, imagen}, and inpainting \cite{repaint, sdedit, dalle2, palette}. 

There have been many techniques introduced recently to improve DM's training and sampling speed. Instead of operating in the pixel space, Latent DMs \cite{ldm} are trained and sampled from a latent space \cite{vqgan} which is much smaller. Fast sampling methods \cite{k_diffusion, pndm, dpm_solver_pp, ddim, deis}, on the other hand, reduce significantly the number of sampling steps. Recent distillation techniques \cite{distillation_guided_diffusion, distillation_diffusion} reduce the model's size and sampling speed even further. These works though improve diffusion models in general, do not directly analyze the \emph{Diffusion Prior} model or its applications for domain specific and conditional generation.

In domain adaptation techniques like MyStyle \cite{my_style_gan}, a pretrained StyleGAN face generator is fine-tuned on a small set of 100 images of a specific entity. However, MyStyle works only in the face domain. In Dreambooth \cite{dreambooth}, a pretrained DM is fine-tuned on a much smaller set of 3-5 images of different types of subjects such as animals, objects, etc. By utilizing a new class-specific prior preservation loss, Dreambooth can synthesize the subject in different scenes and poses that are not available in the reference photos. Instead of fine-tuning all parameters in a pretrained DM model, Textual Inversion \cite{textual_inversion} trains a new word embedding vector for the new subject. In Custom Diffusion \cite{custom_diffusion}, a subset of cross-attention layers is fine-tuned on new concepts. 

\section{Training HDM}
\label{sec:appendix-hdm}
The HDM can be used to generate images from text the same way as \cite{dalle2} up to 512$\times$512 resolution.
\subsection{Diffusion Prior Model}
\label{appendix:proposed-prior}
The Prior model is a denoising diffusion model as proposed in \cite{dalle2} that generates a normalized CLIP L/14 image embedding $\hat{z}_x$ conditioned on an input prompt $y$. The Diffusion Prior $\mathcal{P}_\theta(z_x | y)$ parameterized by $\theta$ is a Causal Transformer \cite{transformer, dalle2} that takes as input a random noise $\epsilon$ sampled from $\mathcal{N}(0, I)$ and a CLIP text embedding $z_y$ = $[z_t, w_{1}, w_{2}, ..., w_n]$ where $z_t$ is the l2 normalized text embedding while $w_{i}$ is the per token encoding, both from a pretrained CLIP L/14 text encoder \cite{clip}. The maximum sequence length for CLIP is $77$ and hence $z_{y}$ has dimensions $78$ x $768$. Additionally, an embedding for the diffusion timestep, the noised CLIP image embedding and a final embedding whose output from the Transformer is used to predict the unnoised CLIP image embedding are added with a causal attention mask and a Mean Squared Error (MSE) objective as done in the LAION prior\textsuperscript{\ref{laionprior}}.

\subsection{Diffusion Decoder Model}
\label{appendix:proposed-ldm}
For the Diffusion Decoder, instead of the multi-stage pixel diffusion model proposed in \cite{dalle2}, we train a custom latent space model inspired by \cite{ldm} for memory and compute efficiency. Our LDM is a denoising diffusion model $\mathcal{D}_\phi(x | z_x)$ parameterized by $\phi$ that takes as input random sample $\epsilon$ from $\mathcal{N}(0, I)$ and the CLIP image embedding $z_x$ to generate the VAE \cite{ldm} latent $\mathbf{z_0}$. It is to be noted that the bold notation for VAE latent $\mathbf{z}$ is different from the notation for CLIP embeddings $z$. The generated latent $\mathbf{z_0}$ is passed through a frozen decoder VAE$_{dec}$ \cite{ldm} of a pretrained VAE to generate the final image $\mathbf{x}$. The LDM's architecture is unchanged from \cite{ldm} except the conditioning input to the U-Net \cite{unet} is modified to be normalized CLIP L/14 embedding $z_x$ instead of $z_y$ as in Stable Diffusion. The pretrained VAEs from \cite{ldm} are used as is. During inference, we can provide $z_x$ from an image $x$ to generate variations of it \cite{dalle2} or provide the generated embedding $\hat{z}_x$ from the Diffusion Prior to generate image from text $\mathbf{y}$ for text-to-image generation. The Diffusion Prior, pre-trained VAE and CLIP L/14 models are frozen while the Diffusion Decoder is trained. 


\section{Data Collection using Classifiers}
\label{sec:appendix-data}
\begin{figure*}[h]
 \centering
 \begin{tabular}{c@{\hspace{0.2cm}}c@{\hspace{0.2cm}}c@{\hspace{0.2cm}}c@{\hspace{0.2cm}}c@{\hspace{0.2cm}}}
 \centering
 \vspace{0.05in} 
 \parbox{.2\linewidth}{Watercolor illustration of Christmas tree toys on a white background} & \parbox{.2\linewidth}{Japanese teapot with flower. Isolated on white background} & \parbox{.2\linewidth}{bengal tiger isolated} & \parbox{.2\linewidth}{Pink Bows set of realistic, isolated on white background}& \parbox{.2\linewidth}{Pattern, glasses with homemade macaroons on a colored background}\\
 \includegraphics[height=1.3in, width=1.3in]{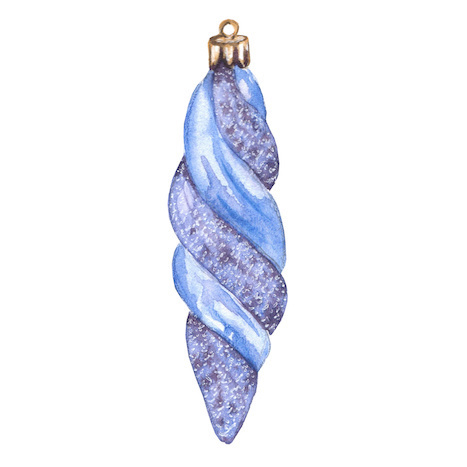}&
 \includegraphics[height=1.3in, width=1.3in]{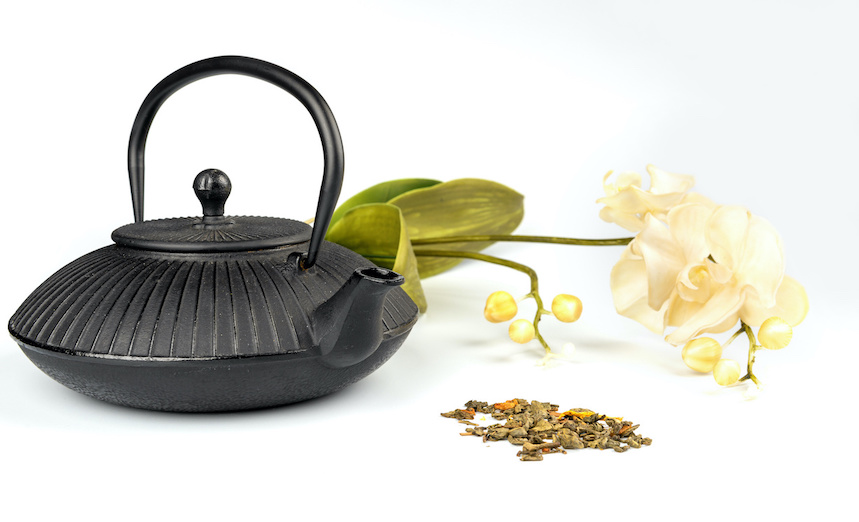}&
 \includegraphics[height=1.3in, width=1.3in]{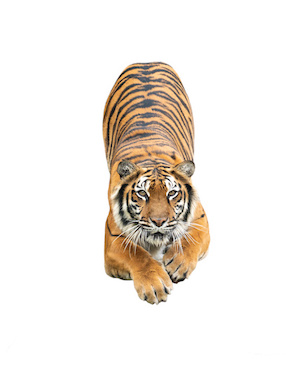}&
 \includegraphics[height=1.3in, width=1.3in]{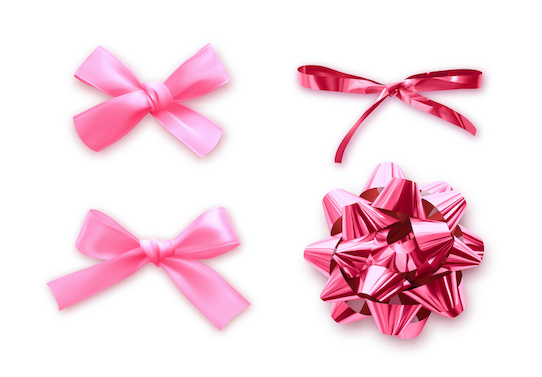}&
 \includegraphics[height=1.3in, width=1.3in]{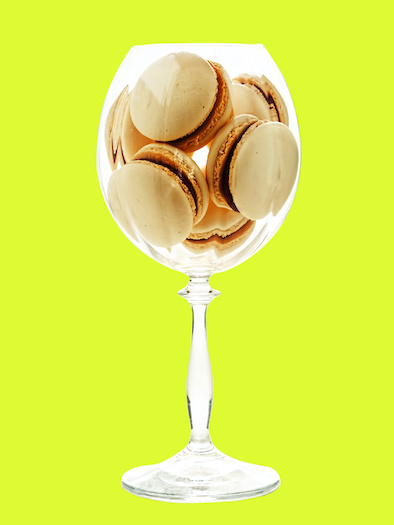} \\
 \end{tabular}
 \caption{Text-image example pairs from the training data for isolated objects prior model}
 \label{fig:appendix-training-isolated}
\end{figure*}

\begin{figure*}[t]
 \centering
 \begin{tabular}{c@{\hspace{0.2cm}}c@{\hspace{0.2cm}}c@{\hspace{0.2cm}}c@{\hspace{0.2cm}}c@{\hspace{0.2cm}}}
 \centering
 \vspace{0.05in} 
 \parbox{.2\linewidth}{Collection of trees. tree set isolated on white background. vector illustration}& \parbox{.2\linewidth}{Abstract horizontal background with colorful waves. Trendy vector illustration in style retro 60s, 70s. Pastel colors} &\parbox{.2\linewidth}{Octopus eating Ramen inside a bowl vector illustration. Food, restaurant, comics, funny design concept}& \parbox{.2\linewidth}{Eggs on plate pattern , illustration, vector on white background} & \parbox{.2\linewidth}{Lion head vector hand drawn}\\
 \includegraphics[height=1.3in, width=1.3in]{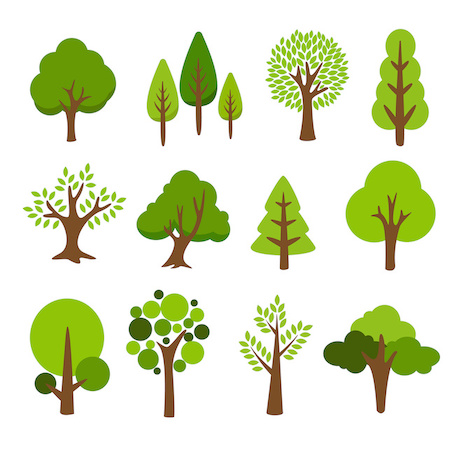}&
 \includegraphics[height=1.3in, width=1.3in]{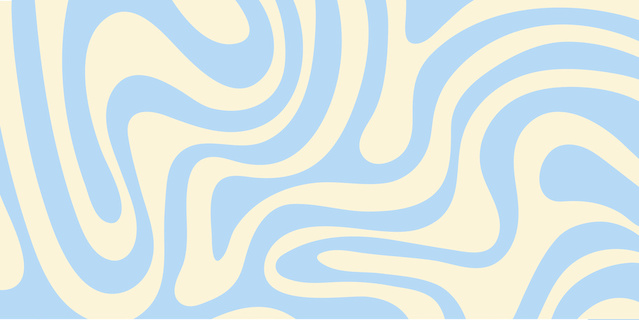}&
 \includegraphics[height=1.3in, width=1.3in]{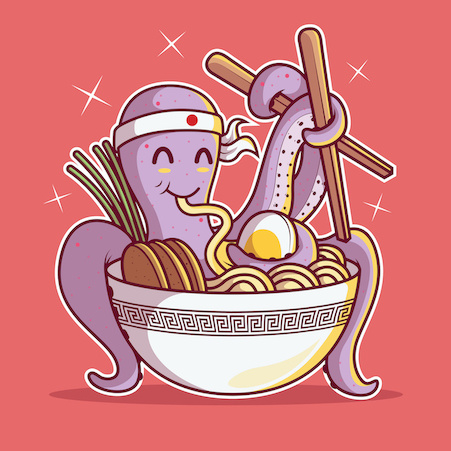}&
 \includegraphics[height=1.3in, width=1.3in]{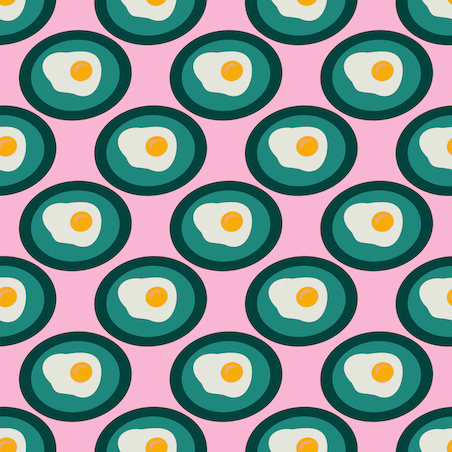}&
 \includegraphics[height=1.3in, width=1.3in]{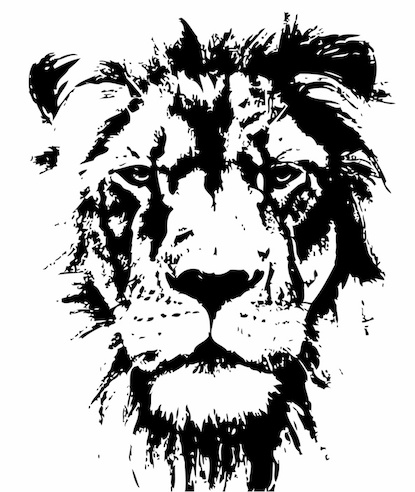} \\
\end{tabular}
 \caption{Text-image example pairs from the training data for the vector prior model}
 \label{fig:appendix-training-vector}
\end{figure*}

\begin{figure*}[t]
 \centering
 \begin{tabular}{c@{\hspace{0.2cm}}c@{\hspace{0.2cm}}c@{\hspace{0.2cm}}c@{\hspace{0.2cm}}c@{\hspace{0.2cm}}}
 \centering
 \vspace{0.05in} 
 \parbox{.2\linewidth}{Plaid and check modern repeat pattern}& \parbox{.2\linewidth}{blank old vintage gold wood table, wall or floor for work and place object on top view horizontal, or wooden board for food preparation in the kitchen and use for background} &\parbox{.2\linewidth}{Marble ink abstract art from exquisite original painting for abstract background. Painting was painted on high quality paper texture to create smooth marble background pattern of ombre alcohol ink }& \parbox{.2\linewidth}{green leaf texture - in detail} & \parbox{.2\linewidth}{Real skin texture of Leopard}\\

 \includegraphics[height=1.3in, width=1.3in]{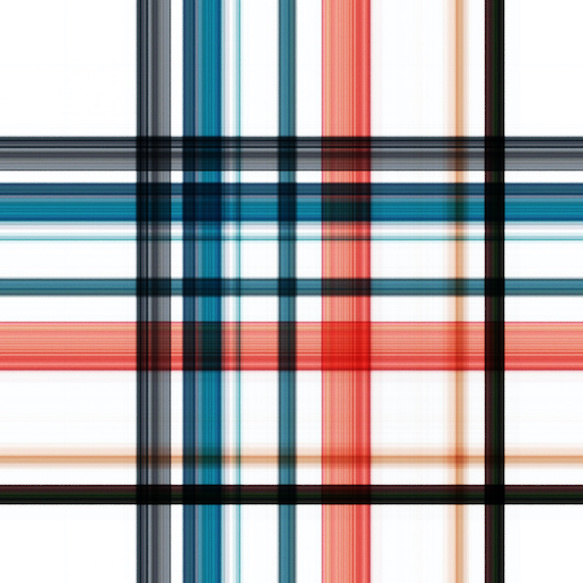}&
 \includegraphics[height=1.3in, width=1.3in]{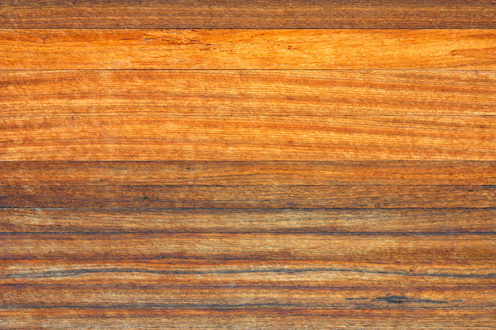}&
 \includegraphics[height=1.3in, width=1.3in]{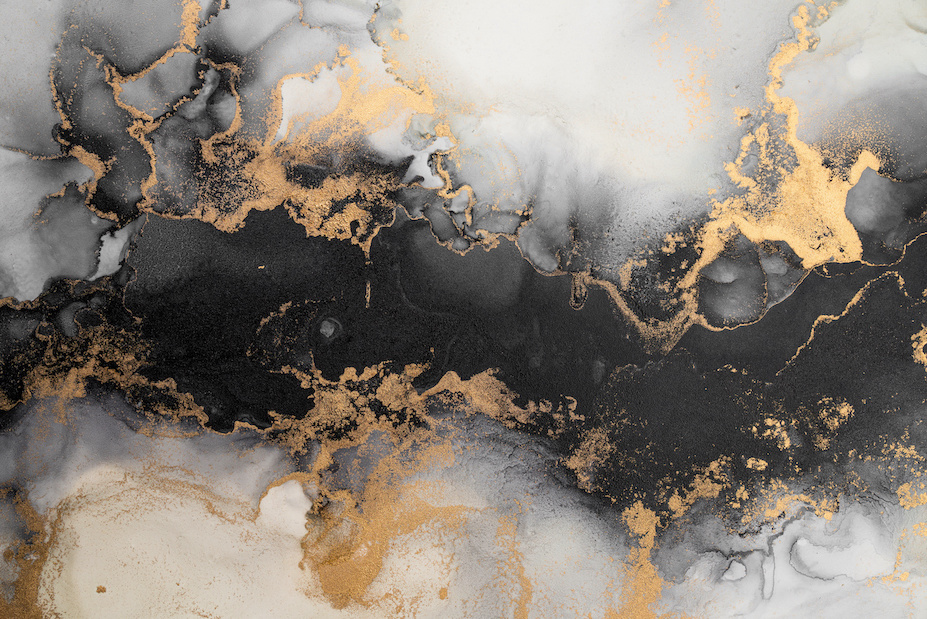}&
 \includegraphics[height=1.3in, width=1.3in]{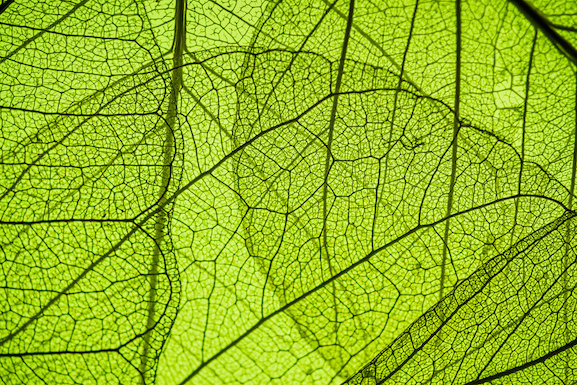}&
 \includegraphics[height=1.3in, width=1.3in]{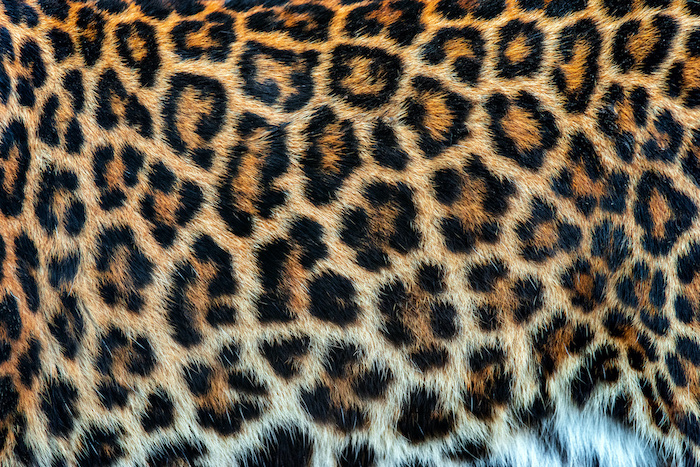} \\
\end{tabular}
 \caption{Text-image example pairs from the training data for the texture prior model.}
 \label{fig:appendix-training-texture}
\end{figure*}

\begin{figure*}[t]
 \centering
 \begin{tabular}{c@{\hspace{0.2cm}}c@{\hspace{0.2cm}}c@{\hspace{0.2cm}}c@{\hspace{0.2cm}}c@{\hspace{0.2cm}}}
 \centering
 \vspace{0.05in} 
 \parbox{.2\linewidth}{closeup of water waves isolated on white} &  \parbox{.2\linewidth}{Two siberian husky play among themselves}& \parbox{.2\linewidth}{Seahorse and starfish seamless pattern. Sea life summer background. Cute sea life background. Design for fabric and decor}& \parbox{.2\linewidth}{Abstract Golden Christmas - christmas background illustration} & \parbox{.2\linewidth}{Vase with still life a bouquet of flowers. Oil painting
}\\
 \includegraphics[height=1.3in, width=1.3in]{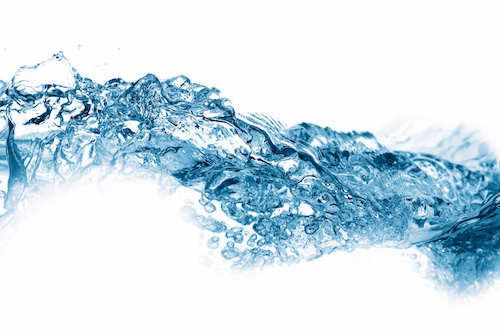}&
 \includegraphics[height=1.3in, width=1.3in]{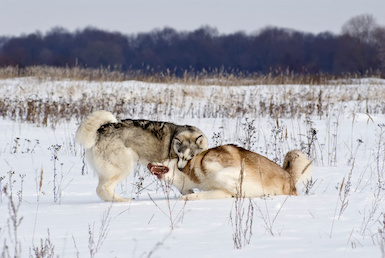}&
 \includegraphics[height=1.3in, width=1.3in]{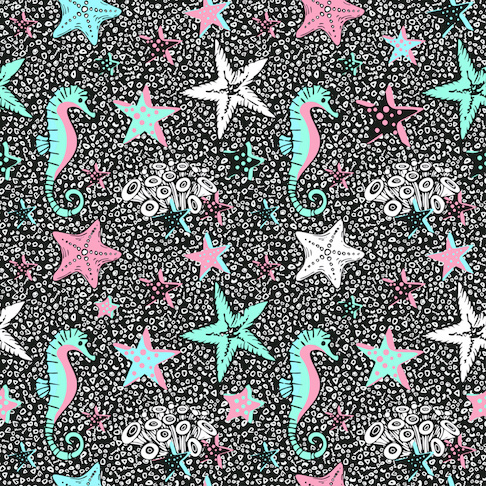}&
 \includegraphics[height=1.3in, width=1.3in]{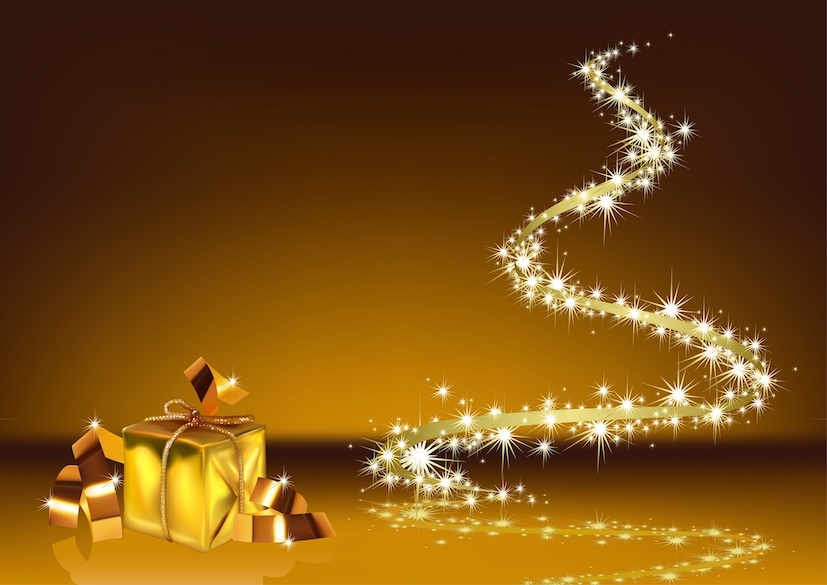}&
 \includegraphics[height=1.3in, width=1.3in]{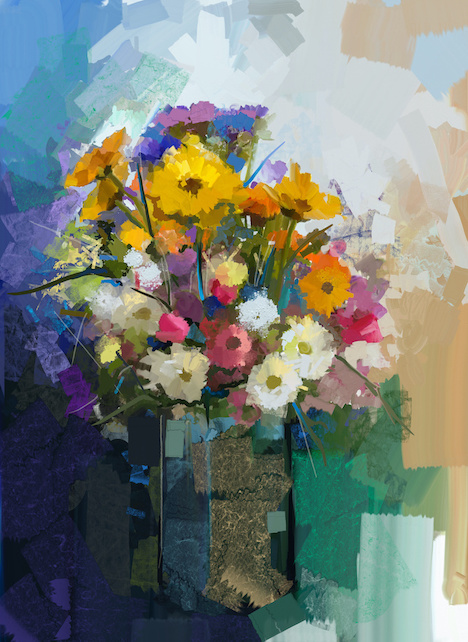} \\  
 \end{tabular}
 \caption{Text-image example pairs from the training data used to train the larger Diffusion Decoder LDM model.}
 \label{fig:appendix-training-ldm}
\end{figure*}

We show few examples of text image pairs from the training data obtained for each domain specific prior as well as the large decoder. Example training samples for the \emph{isolated objects} domain is shown in Fig.\ref{fig:appendix-training-isolated}, \emph{vectors} in Fig.\ref{fig:appendix-training-vector}, for \emph{texture} in Fig.\ref{fig:appendix-training-texture} and for the larger decoder in Fig.\ref{fig:appendix-training-ldm}.
\subsection{Prior}
\label{sec:appendix-data-prior}
\noindent\textbf{Texture:} We manually annotated 30K images from stock to use as positive samples for training. We get an F1-score of 69\% for the positive class of texture images and 89\% for the negative class. We then use this to get 10M images that were used to train the texture prior $\mathcal{P}_{\theta1}^{texture}$. \\
\noindent\textbf{Vectors:} We use stock metadata to gather 1M positive and negative samples for vectors to train a classifier. We then use this to get 26M images that were used to train the vector prior $\mathcal{P}_{\theta2}^{vector}$\\
\noindent\textbf{Isolated Objects:} We manually annotated 28K images from stock to use as positive samples. We get an F1-score of 85\% for the positive class of images isolated on plain backgrounds and 74\% for the negative class. We then use this to get 20M images that were used to train the isolated object prior $\mathcal{P}_{\theta3}^{isolated}$\\
\noindent\textbf{Color:} For color prior $\mathcal{P}_{\theta4}$, we train on a 61M only English subset of image-text pairs from the original 77M filtered stock data used to train the LDM as described in Sec.\ref{subsec:proposed-color} 

\subsection{Decoder}
\label{sec:appendix-data-decoder}
\noindent\textbf{Human presence:} To detect images that contain humans, we manually annotated 180K images from stock and then train a classifier to classify human presence/absence. We get an F1-score of 95\% for presence of humans and 98\% for the absence. \\
\noindent\textbf{Text Presence:} Similarly, we manually annotate 40K images for text presence and train a classifier. We get an F1-score of 94\% in detecting images without text and 78\% for images with text as tested on a random 5K images test set. 

Once we have human and text presence classifiers, we run those on whole of stock data to get 77M images that have no humans or text. This simplifies the distribution but has no consequence over the priors, its capabilities and the results. With the LDM trained on a large dataset of diverse images, we now describe the process of dataset curation for the specific domains. To ensure that the model doesn't generate NSFW images, we also remove NSFW images from the training corpus using a pretrained NSFW classifier. 

\section {Composable Diffusion}
\begin{figure}[t]
 \centering
 \footnotesize
 \begin{tabular}{c@{\hspace{0.1cm}}c@{\hspace{0.1cm}}c@{\hspace{0.1cm}}c@{\hspace{0.1cm}}c@{\hspace{0.1cm}}}
 \centering
 \vspace{0.03in} 
 Color Image & Color Only& Texture Only & Color + Texture \\
 \includegraphics[height=0.76in, width=0.76in]{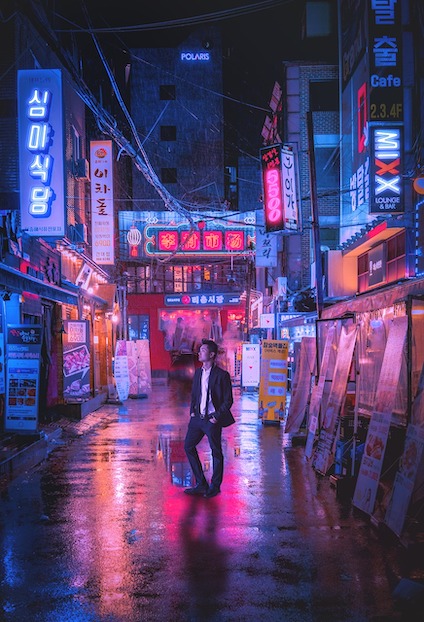}&
 \includegraphics[height=0.76in, width=0.76in]{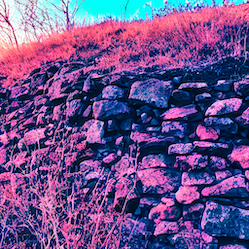}&
 \includegraphics[height=0.76in, width=0.76in]{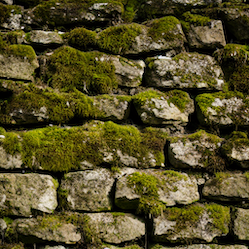}&
 \includegraphics[height=0.76in, width=0.76in]{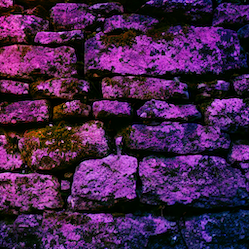}&
 \\
 \multicolumn{5}{c}{\emph{Text prompt: "old stone moss wall at the hillside"}} \\ 
 Color Image & Color Only& Isolated Only  & Color + Isolated \\
 \includegraphics[height=0.76in, width=0.76in]{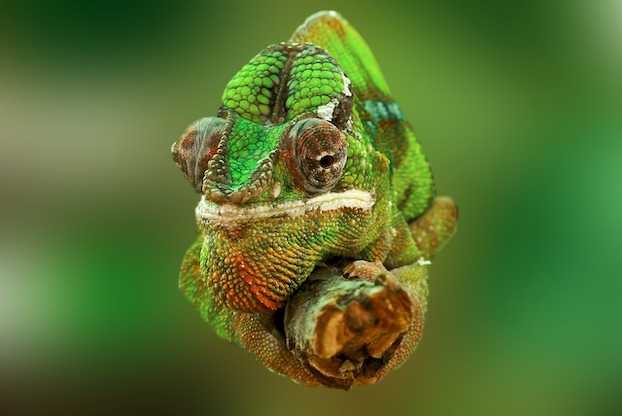}&
 \includegraphics[height=0.76in, width=0.76in]{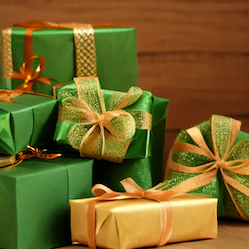}&
 \includegraphics[height=0.76in, width=0.76in]{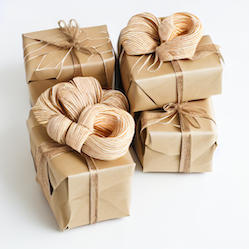}&
 \includegraphics[height=0.76in, width=0.76in]{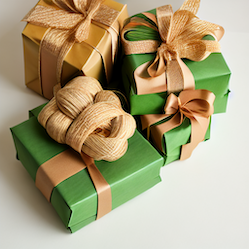}&
 \\
 \multicolumn{5}{c}{\emph{Text prompt: "gift boxes wrapped with ribbons"}} \\ 
\end{tabular}
 \caption{Composable diffusion results. Images for each prompt have been generated using the same seed value.}
 \label{fig:composable}
 \vspace{-0.3in}
\end{figure}

In Fig. \ref{fig:composable} we show that the different priors are compositional and can be combined to generate domain specific images in a specific color. For this experiment, we sample from both prior models and at each timestep, the output from the prior models are composed before being passed on to the models for next step. We can see from Fig.\ref{fig:composable} that when composed, we can get domain specific images based on the domain prior used, while also adhering to the color palette passed through the color prior. This enables the possibility of training multiple smaller specialized priors and ensembling for better controllability.

\section{Limitations}  
\label{sec:appendix-limitations}
We see that with the color prior, when a vector/illustration is used as an exemplar, we get a vector image as output. We believe that feeding the color histogram of the ground truth image to the Diffusion Prior model during training makes the model associate the color distribution to the image to be generated. We tried to discritize the color histograms for all images to match that of vectors to overcome this issue, but this caused loss in color relevance possibly because the discretization is a lossy process. We also experimented with prompts which already have color words in them and observed that the model generally gives more priority to the color histogram to get the color cues over the color words present in the text prompt unless there is an overlap between colors in the text and color histogram. For the domain priors, we believe that the \emph{Diffusion Prior} could be reduced in capacity even further without tradinf off quality or domain relevance. Though we show results for texture, isolated objects and vectors as the domains and color as the additional conditional input, we believe that the general approach would work for most other domains as well as other conditional inputs.

\section{Additional Examples}
\label{sec:appendix-results}
\begin{figure*}[t]
 \centering
 \begin{tabular}{c@{\hspace{0.2cm}}c@{\hspace{0.2cm}}c@{\hspace{0.2cm}}c@{\hspace{0.2cm}}}
 \centering
 Input & \multicolumn{3}{c}{Variations}\\
 \includegraphics[width=.23\linewidth]{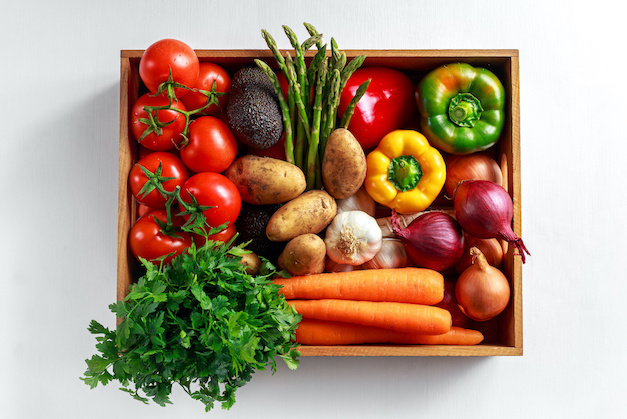}&
 \includegraphics[width=.23\linewidth]{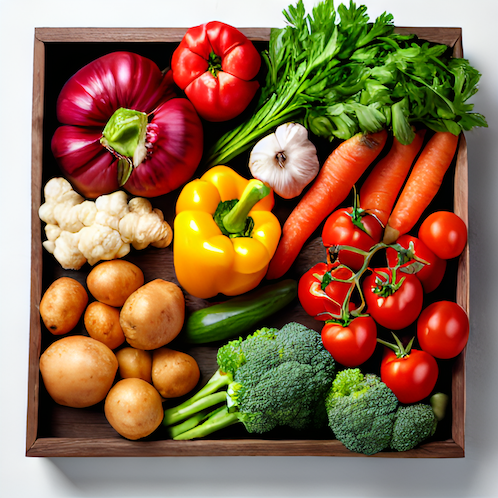}&
 \includegraphics[width=.23\linewidth]{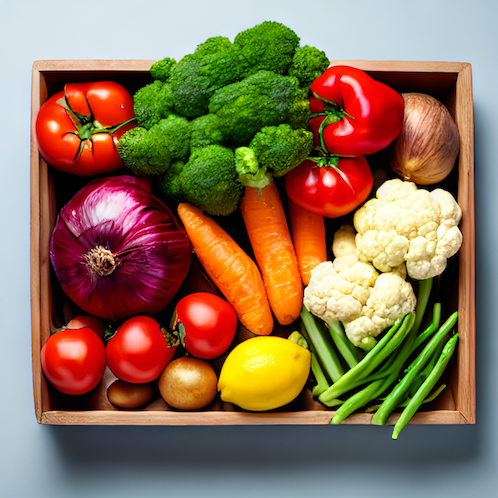}&
 \includegraphics[width=.23\linewidth]{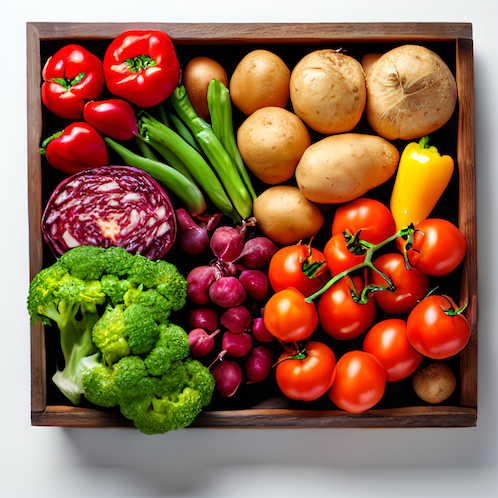} \\
 &
 \includegraphics[width=.23\linewidth]{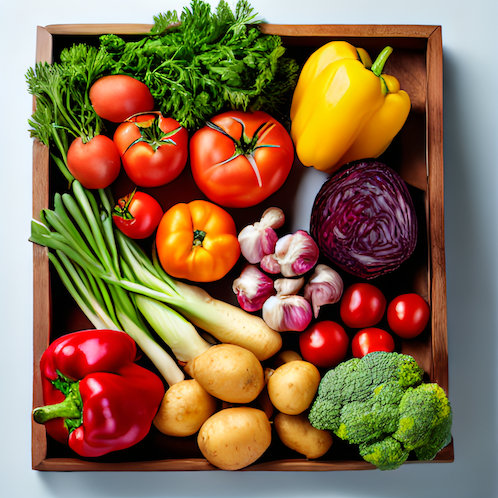}&
 \includegraphics[width=.23\linewidth]{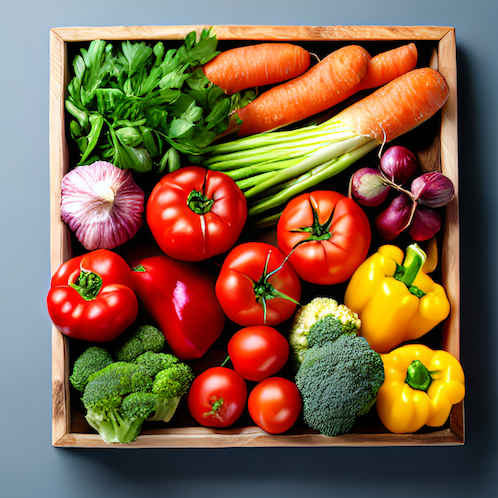}&
 \includegraphics[width=.23\linewidth]{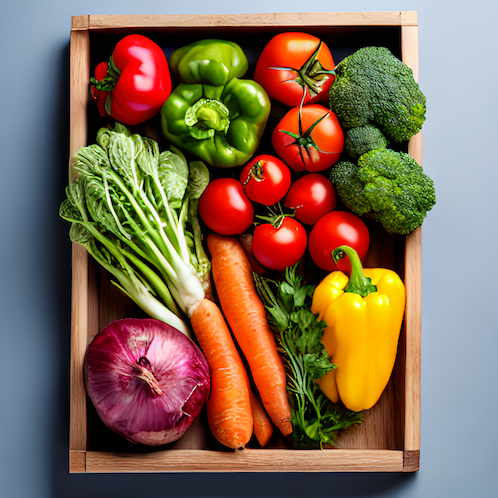}\\ \\
 \includegraphics[width=.23\linewidth]{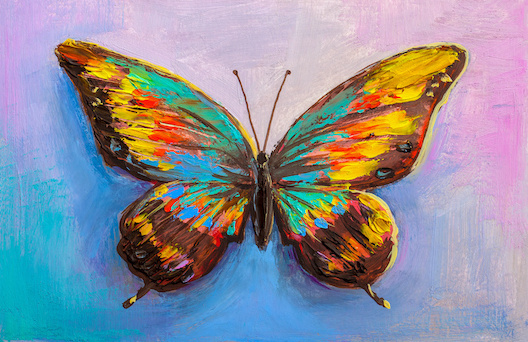}&
 \includegraphics[width=.23\linewidth]{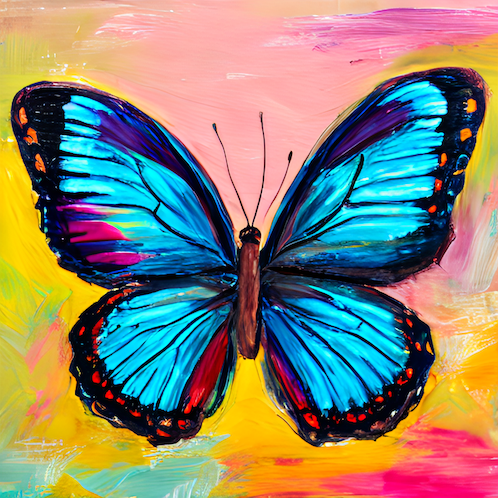}&
 \includegraphics[width=.23\linewidth]{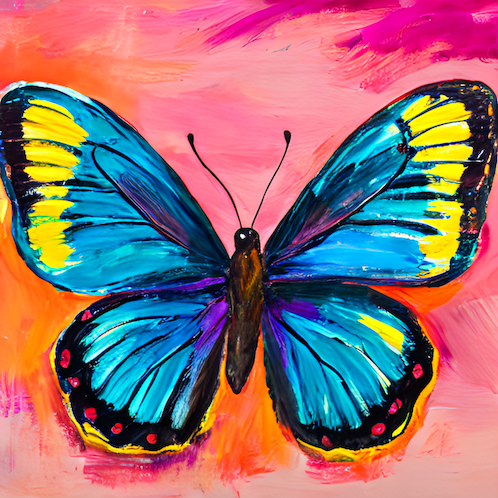}&
 \includegraphics[width=.23\linewidth]{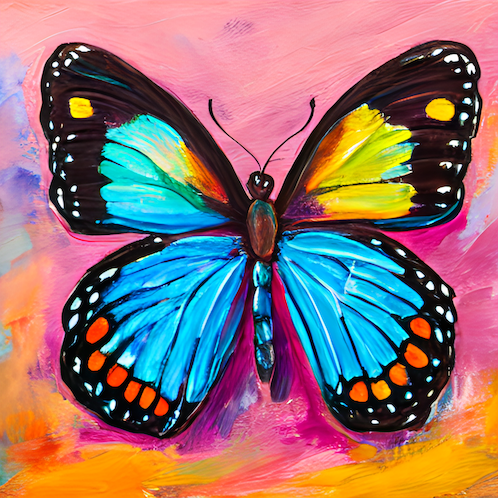} \\
 &
 \includegraphics[width=.23\linewidth]{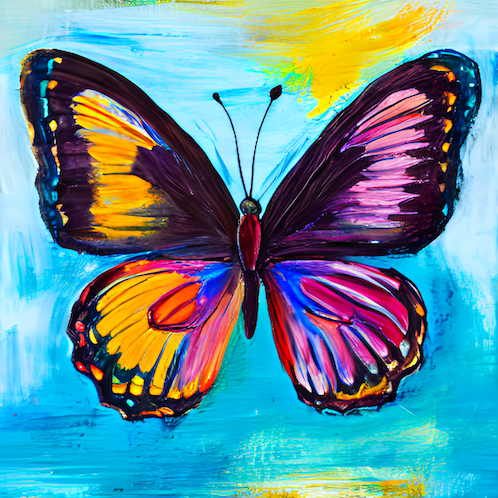}&
 \includegraphics[width=.23\linewidth]{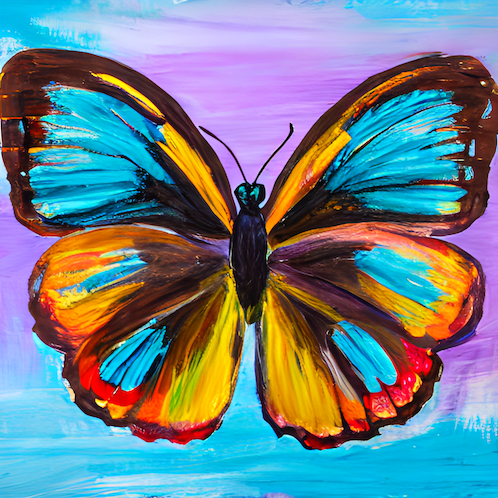}&
 \includegraphics[width=.23\linewidth]{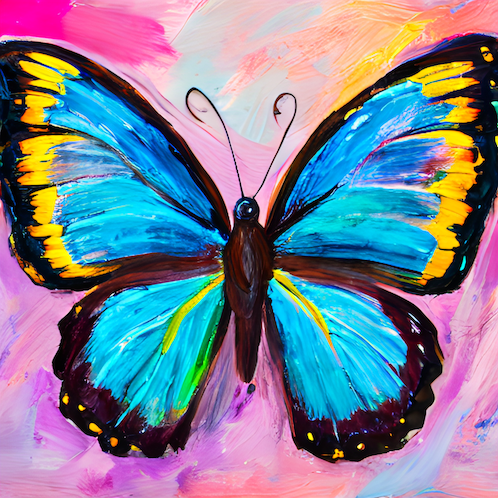}
 \end{tabular}
 \caption{Example image variations generated by conditioning the Diffusion Decoder on the CLIP image embedding of the input image showin the capacity of our trained LDM model.}
 \label{fig:appendix-variations}
\end{figure*}

We provide more qualitative examples for the proposed method. 
We show example image variations from the largest LDM decoder model trained on an internal dataset for quality check in Fig.\ref{fig:appendix-variations}. 

\begin{figure*}[t]
 \centering
 \begin{tabular}{c@{\hspace{0.3cm}}c@{\hspace{0.3cm}}c@{\hspace{0.3cm}}}
 \centering
 \parbox{.3\linewidth}{Acrylic colors and ink in water. Ink blot}&\parbox{.3\linewidth}{bulb breaking with splash of color, creative idea}&\parbox{.3\linewidth}{Cute cat. Portrait of a black tabby cat sleeping}\\ \\
 \includegraphics[width=.3\linewidth]{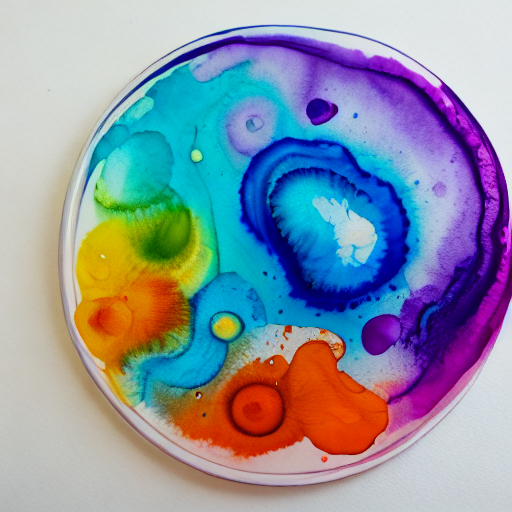}&
 \includegraphics[width=.3\linewidth]{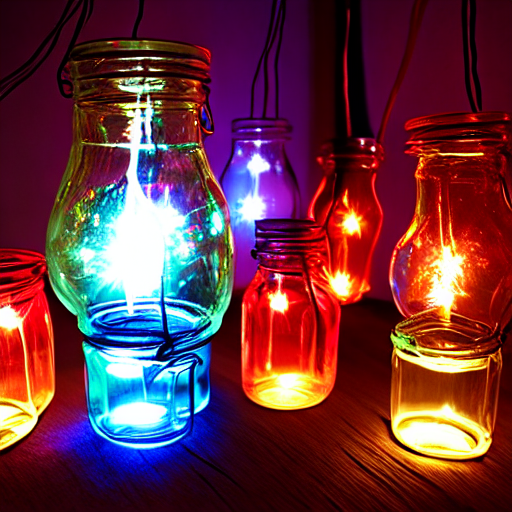}&
 \includegraphics[width=.3\linewidth]{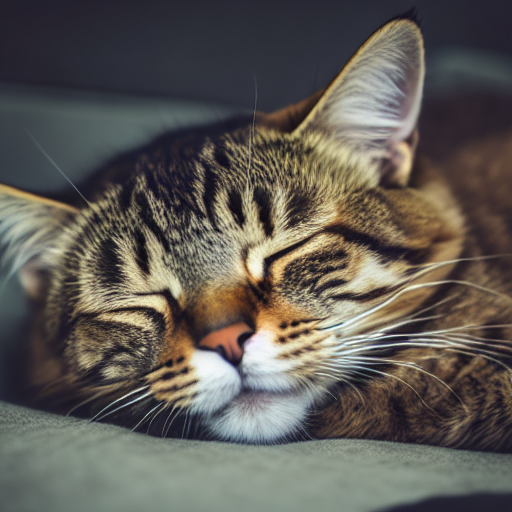} \\ \\
 \parbox{.3\linewidth}{geometric polygonal of earth globe}&\parbox{.3\linewidth}{Fantasy hero in armor. sketch art for artist creativity and inspiration}&\parbox{.3\linewidth}{The interior design of a lavish side outside garden, with a teak hardwood deck and pergola}\\ \\
 \includegraphics[width=.3\linewidth]{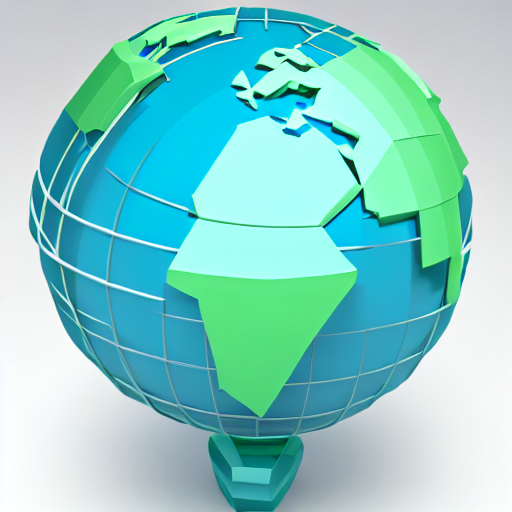}&
 \includegraphics[width=.3\linewidth]{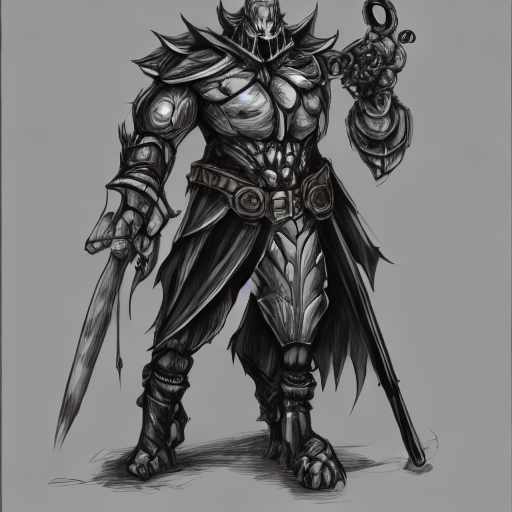}&
 \includegraphics[width=.3\linewidth]{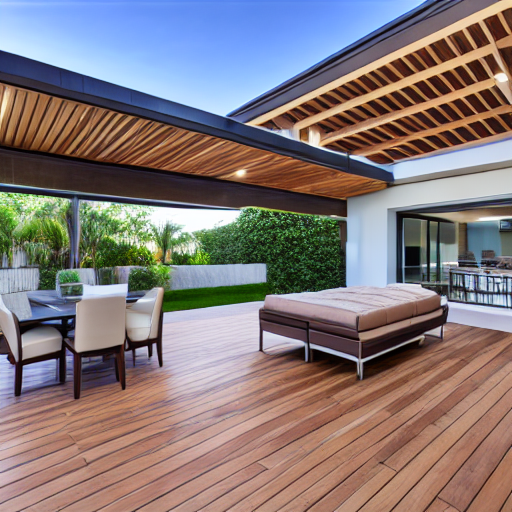} \\ \\
 \parbox{.3\linewidth}{dramatic sea dark sky thunderstorm ghost ship}&\parbox{.3\linewidth}{backpack covered with leaves}&\parbox{.3\linewidth}{new york city on a foggy afternoon, watercolor}\\ \\
 \includegraphics[width=.3\linewidth]{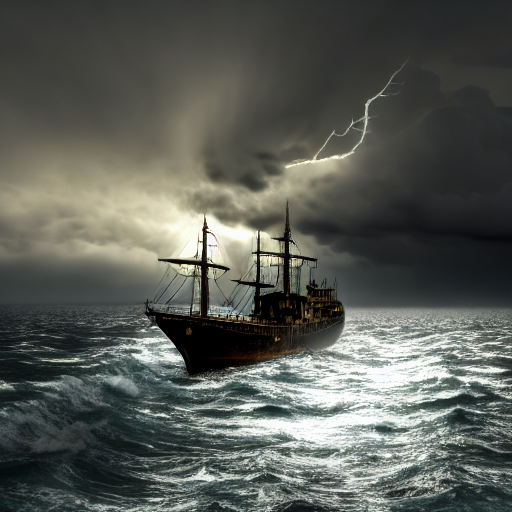}&
 \includegraphics[width=.3\linewidth]{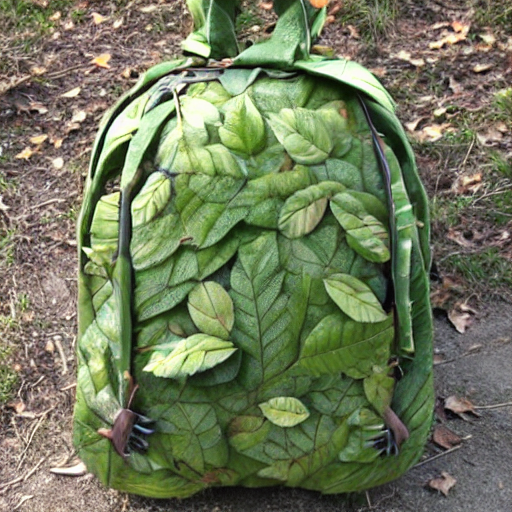}&
 \includegraphics[width=.3\linewidth]{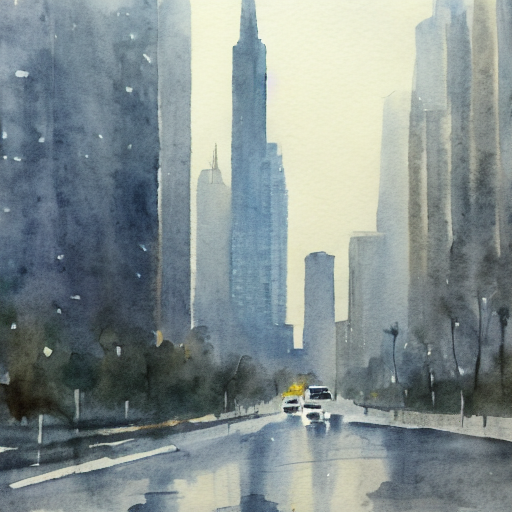}
 \end{tabular}
 \caption{Example of text to image generation using the HDM architecture. The \emph{Diffusion Prior} is the publicly available LAION prior model while the \emph{Diffusion Decoder} is our LDM model. We can observe high quality and relevant generations from the HDM.}
 \label{fig:appendix-hdm}
\end{figure*}

To ensure the overall HDM pipeline works, we show examples of images generated by using the existing publicly available LAION prior model with our trained LDM for text to image generation in Fig.\ref{fig:appendix-hdm}.

\begin{figure*}[t]
 \centering
 \begin{tabular}{c@{\hspace{0.3cm}}c@{\hspace{0.3cm}}c@{\hspace{0.3cm}}}
 \centering
 \parbox{.3\linewidth}{dark chocolate bonbons with milk souffle fillings}&\parbox{.3\linewidth}{ball of colorful strings}&\parbox{.3\linewidth}{geometric polygonal raging bull charging}\\ \\
 \includegraphics[width=.3\linewidth]{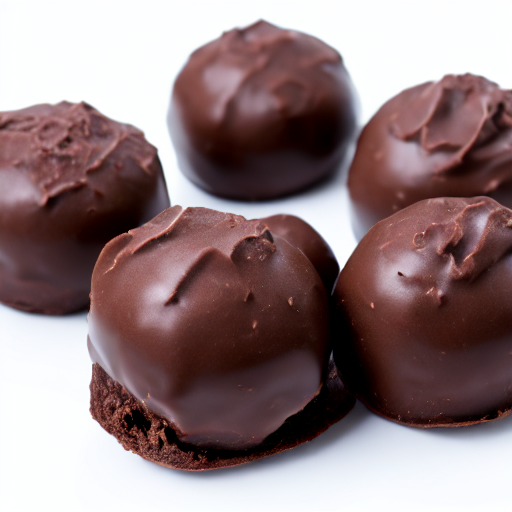}&
 \includegraphics[width=.3\linewidth]{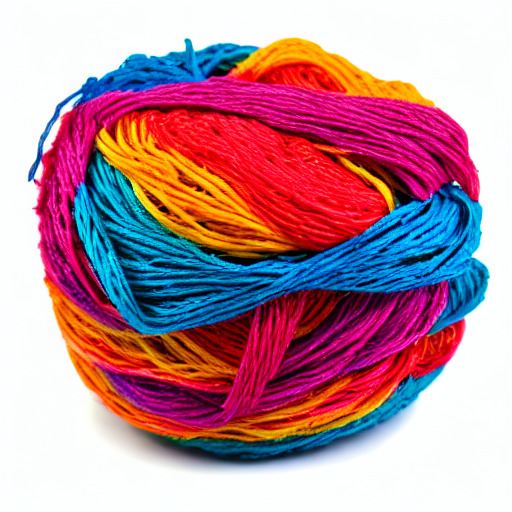}&
 \includegraphics[width=.3\linewidth]{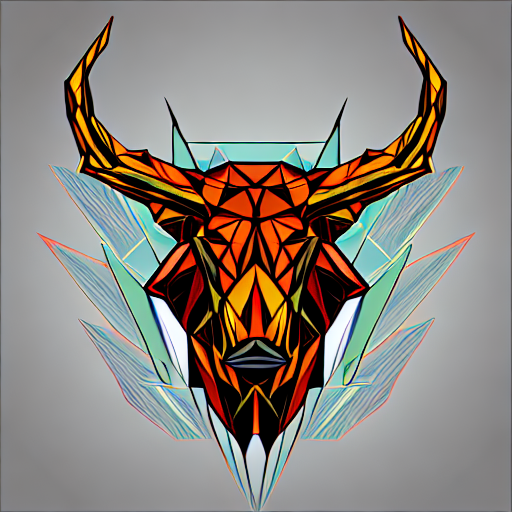} \\ \\
 \parbox{.3\linewidth}{green brown grasshopper on a leaf, watercolor}&\parbox{.3\linewidth}{set of fresh vegetables icons}&\parbox{.3\linewidth}{ink droplet splash in water}\\ \\
 \includegraphics[width=.3\linewidth]{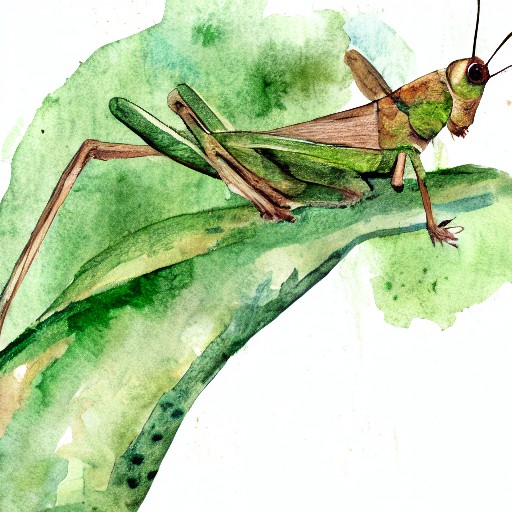}&
 \includegraphics[width=.3\linewidth]{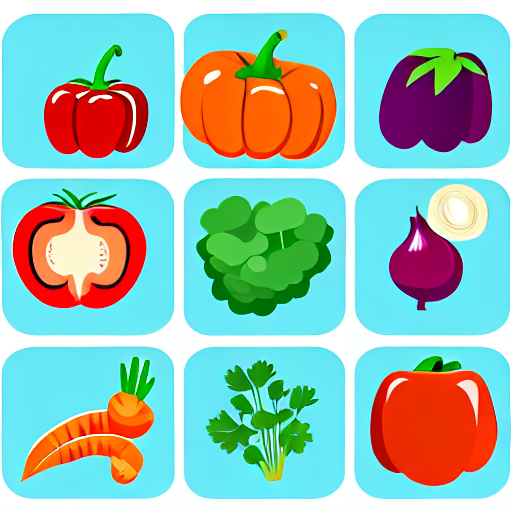}&
 \includegraphics[width=.3\linewidth]{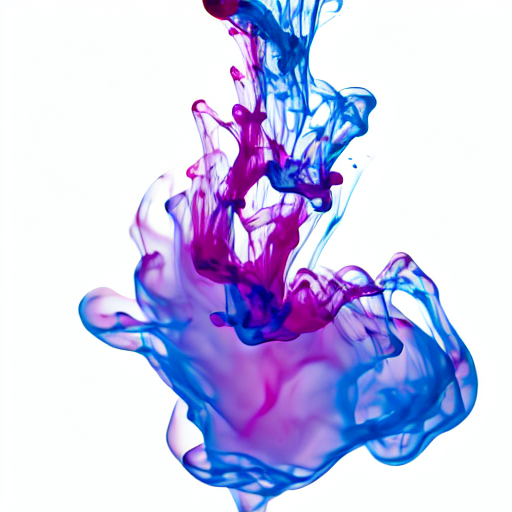} \\ \\
 \parbox{.3\linewidth}{coffee cup with latte art with coffee beans}&\parbox{.3\linewidth}{chicken tikka masala with rice and fresh coriander}&\parbox{.3\linewidth}{beautiful sunflower on a bright sunny day}\\ \\
 \includegraphics[width=.3\linewidth]{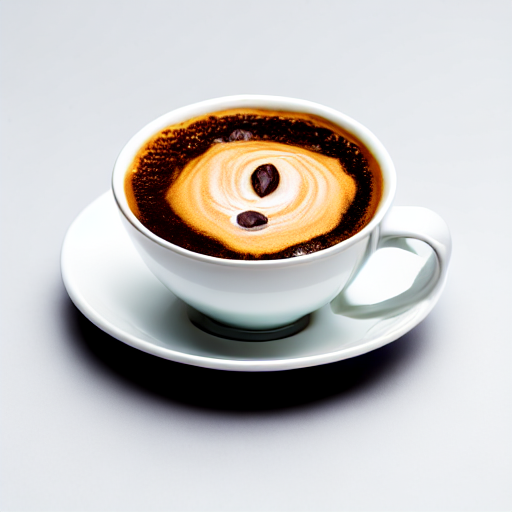}&
 \includegraphics[width=.3\linewidth]{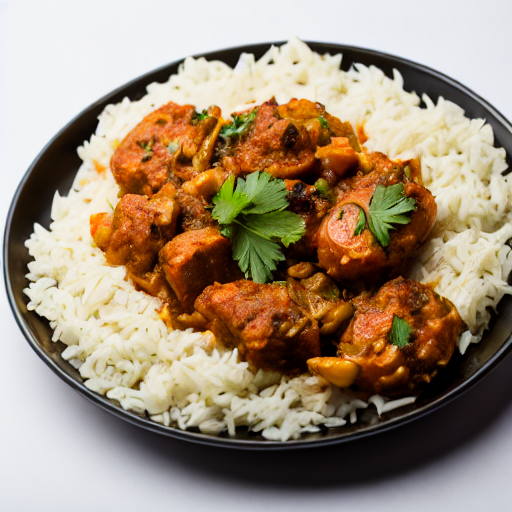}&
 \includegraphics[width=.3\linewidth]{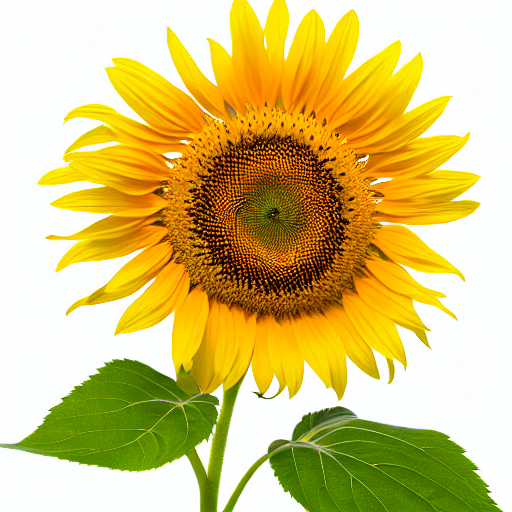}
 \end{tabular}
 \caption{Example images generated using the isolated objects prior with our LDM specific to isolated objects domain. We can see that all objects and concepts are isolated on a plain colored background.}
 \label{fig:appendix-isolated}
\end{figure*}

\begin{figure*}[t]
 \centering
 \begin{tabular}{c@{\hspace{0.3cm}}c@{\hspace{0.3cm}}c@{\hspace{0.3cm}}}
 \centering
 \parbox{.3\linewidth}{grater icon design}&\parbox{.3\linewidth}{Metal glossy shiny geometric shapes with 3d effect composition. Techno futuristic abstract background For Wallpaper, Banner, Background, Card, Book Illustration, landing page}&\parbox{.3\linewidth}{Silver line Bottle of olive oil icon isolated on dark red background. Jug with olive oil icon}\\ \\
 \includegraphics[width=.3\linewidth]{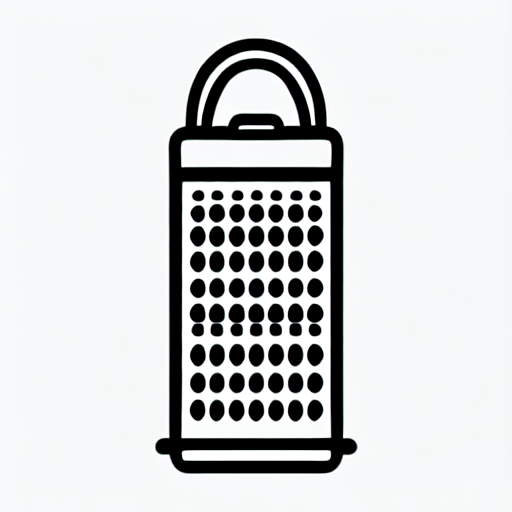}&
 \includegraphics[width=.3\linewidth]{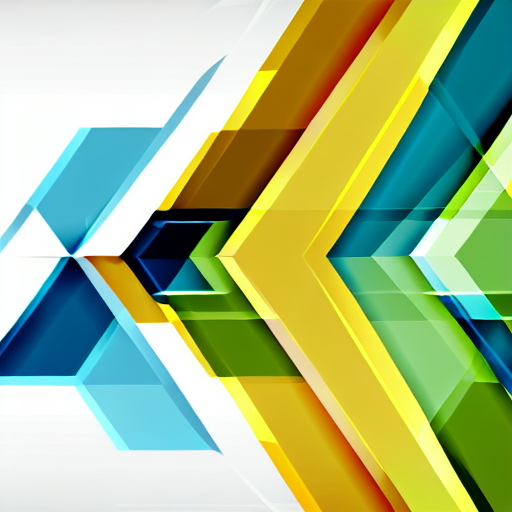}&
 \includegraphics[width=.3\linewidth]{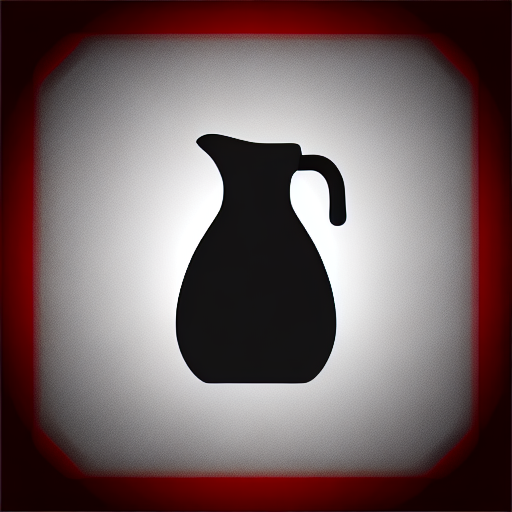} \\ \\
 \parbox{.3\linewidth}{Hummingbird}&\parbox{.3\linewidth}{retro cartoon caterpillar}&\parbox{.3\linewidth}{wardrobe icon image}\\ \\
 \includegraphics[width=.3\linewidth]{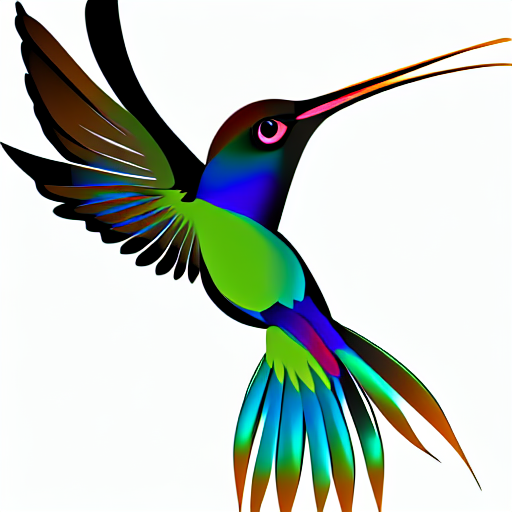}&
 \includegraphics[width=.3\linewidth]{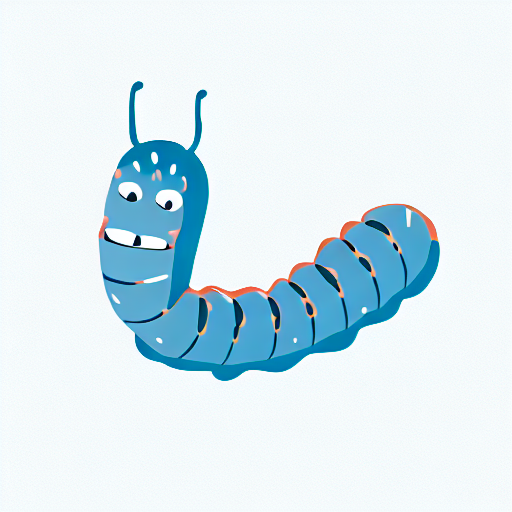}&
 \includegraphics[width=.3\linewidth]{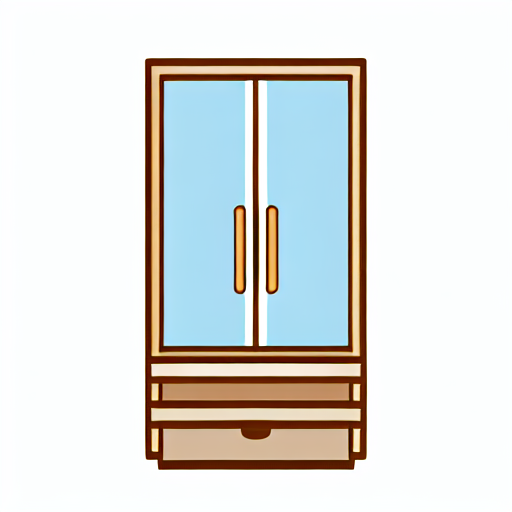} \\ \\
 \parbox{.3\linewidth}{Side view of cute turtle with small shell}&\parbox{.3\linewidth}{Isolated videogame portable console control line style icon}&\parbox{.3\linewidth}{Advertising billboards}\\ \\
 \includegraphics[width=.3\linewidth]{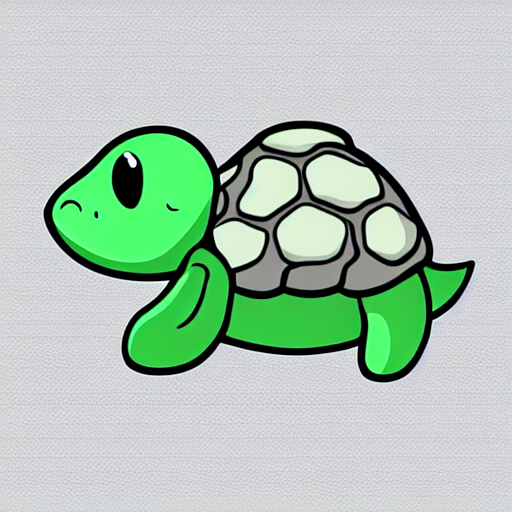}&
 \includegraphics[width=.3\linewidth]{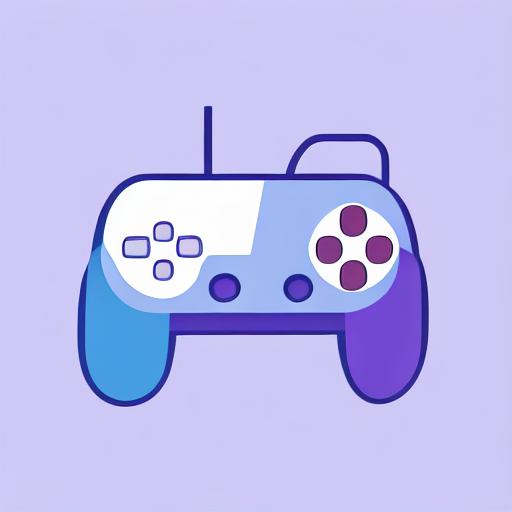}&
 \includegraphics[width=.3\linewidth]{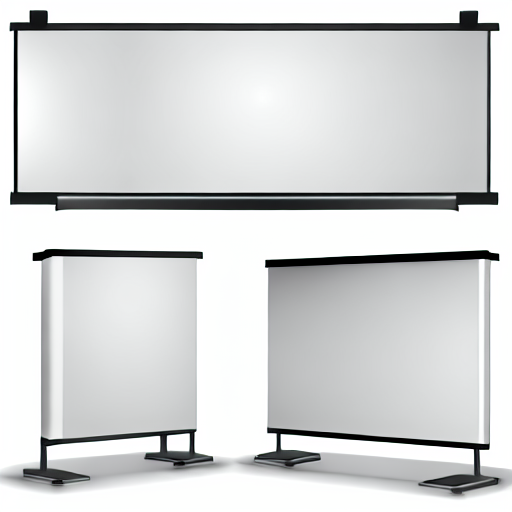}
 \end{tabular}
 \caption{Example images generated using the vector prior with our LDM. We can see that all objects and concepts have abstract vector shapes.}
 \label{fig:appendix-vector}
\end{figure*}

\begin{figure*}[t]
 \centering
 \begin{tabular}{c@{\hspace{0.3cm}}c@{\hspace{0.3cm}}c@{\hspace{0.3cm}}}
 \centering
 \parbox{.3\linewidth}{crochet pattern}&\parbox{.3\linewidth}{office carpet black white red pattern}&\parbox{.3\linewidth}{felt fabric}\\ \\
 \includegraphics[width=.3\linewidth]{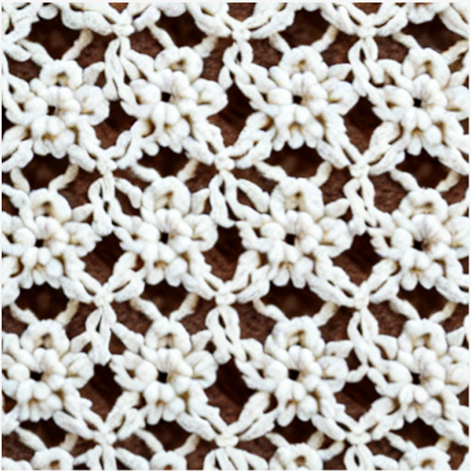}&
 \includegraphics[width=.3\linewidth]{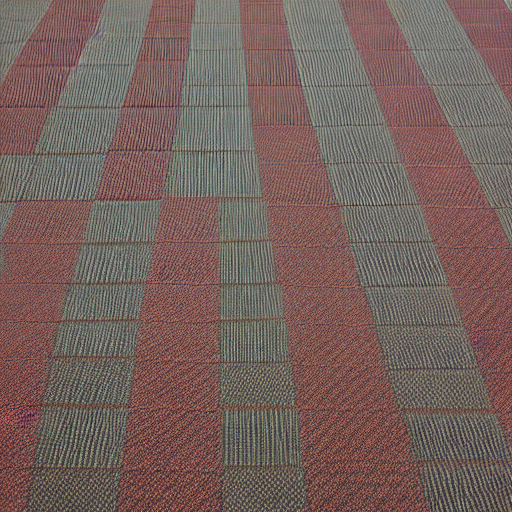}&
 \includegraphics[width=.3\linewidth]{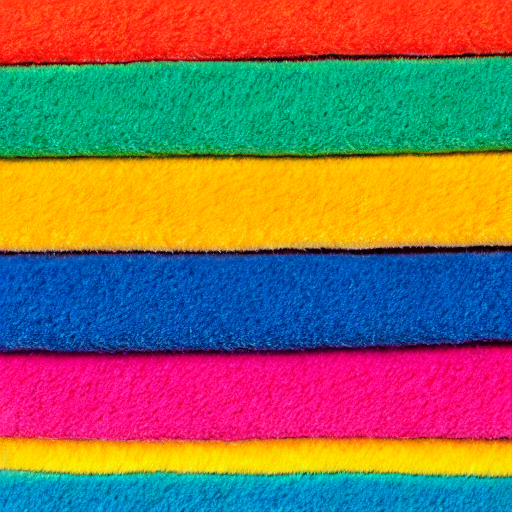} \\ \\
 \parbox{.3\linewidth}{majestic crocodile with mouth open}&\parbox{.3\linewidth}{old worn out hardwood floor}&\parbox{.3\linewidth}{polished concrete}\\ \\
 \includegraphics[width=.3\linewidth]{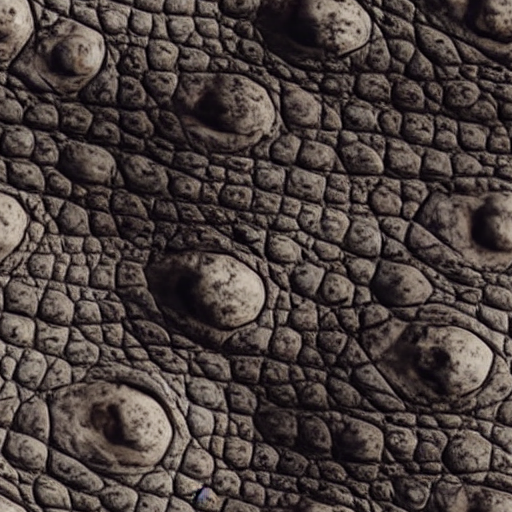}&
 \includegraphics[width=.3\linewidth]{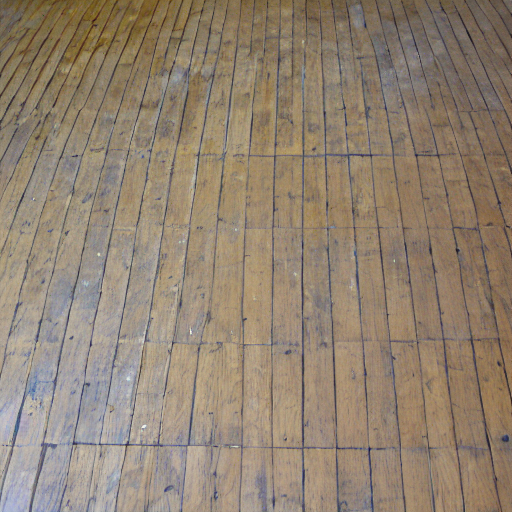}&
 \includegraphics[width=.3\linewidth]{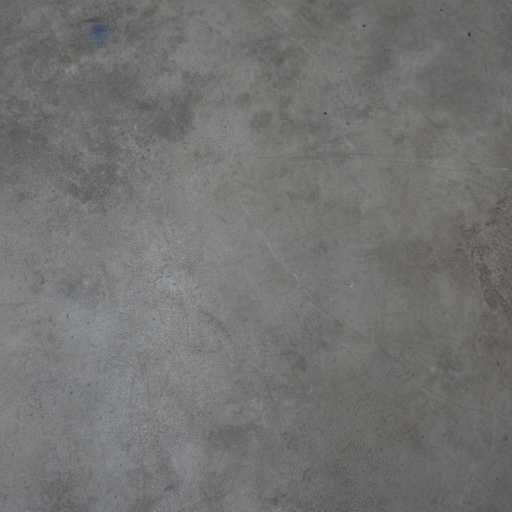} \\ \\
 \parbox{.3\linewidth}{strawberry milkshake with foam}&\parbox{.3\linewidth}{peeling painted interior wall, rusted}&\parbox{.3\linewidth}{fresh erupted lava from volcano}\\ \\
 \includegraphics[width=.3\linewidth]{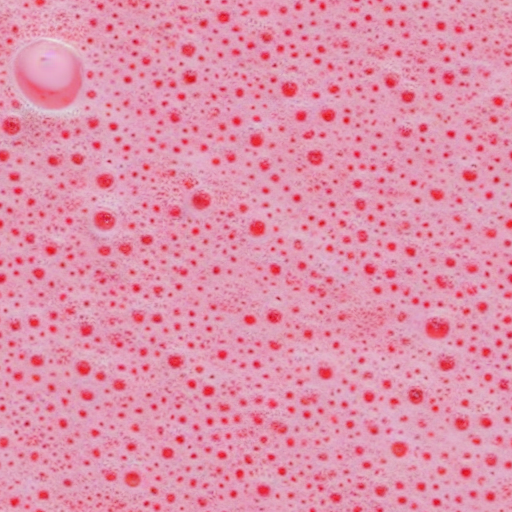}&
 \includegraphics[width=.3\linewidth]{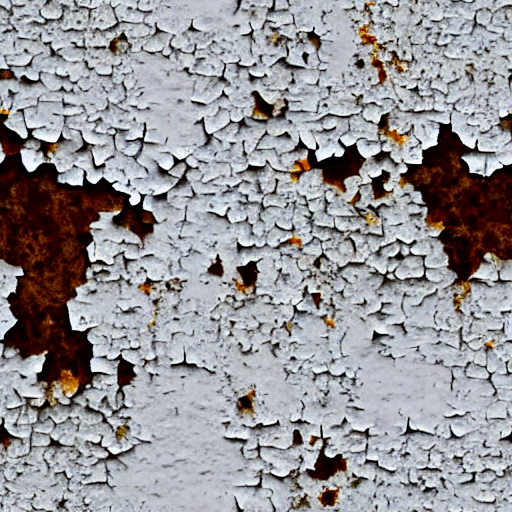}&
 \includegraphics[width=.3\linewidth]{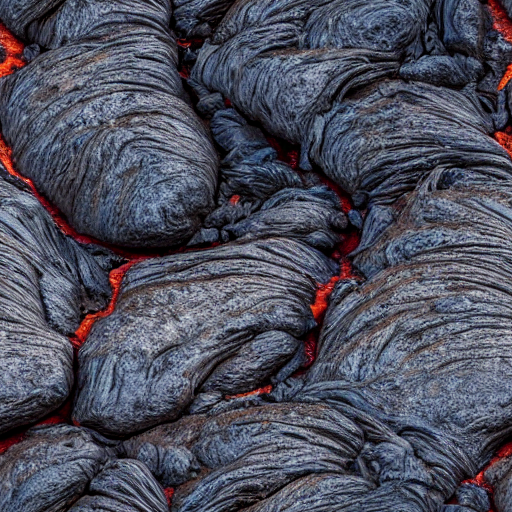}
 \end{tabular}
 \caption{Example images generated using the texture prior with our LDM. We can see that all objects and concepts have a global texture free of regular object specific images.}
 \label{fig:appendix-texture}
\end{figure*}

Further examples for domain specific generation using the isolated objects prior with our LDM is shown in Fig.\ref{fig:appendix-isolated}. Examples for text-to-vector domain image generation using the vector prior and the LDM is shown in Fig.\ref{fig:appendix-vector} while that of text-to-texture domain images is shown in Fig.\ref{fig:appendix-texture}.

\begin{figure*}[t]
 \centering
 \begin{tabular}{c@{\hspace{0.3cm}}c@{\hspace{0.3cm}}c@{\hspace{0.3cm}}}
 \centering
 \parbox{.3\linewidth}{Acrylic colors and ink in water. Ink blot}&\parbox{.3\linewidth}{bulb breaking with splash of color, creative idea}&\parbox{.3\linewidth}{Cute cat. Portrait of a black tabby cat sleeping}\\ \\
 \includegraphics[width=.3\linewidth]{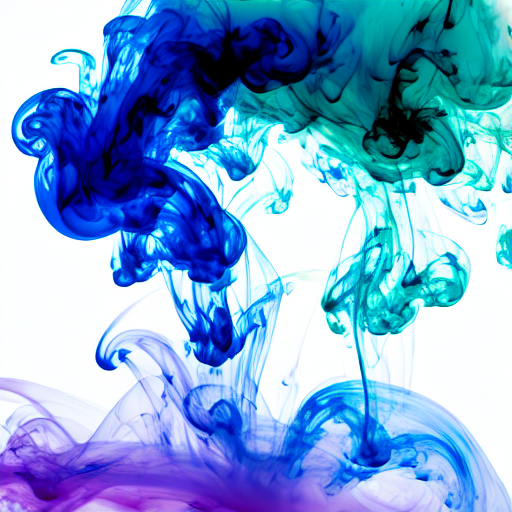}&
 \includegraphics[width=.3\linewidth]{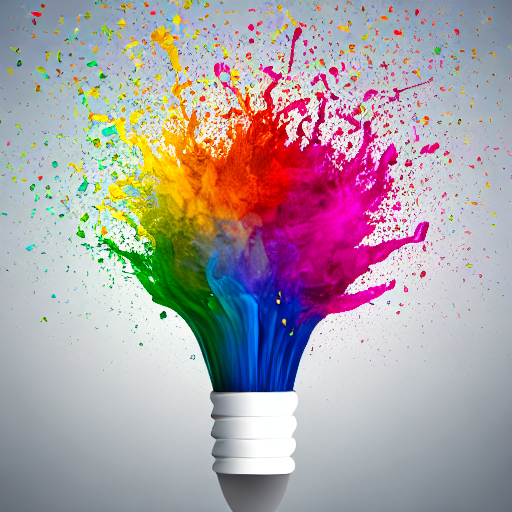}&
 \includegraphics[width=.3\linewidth]{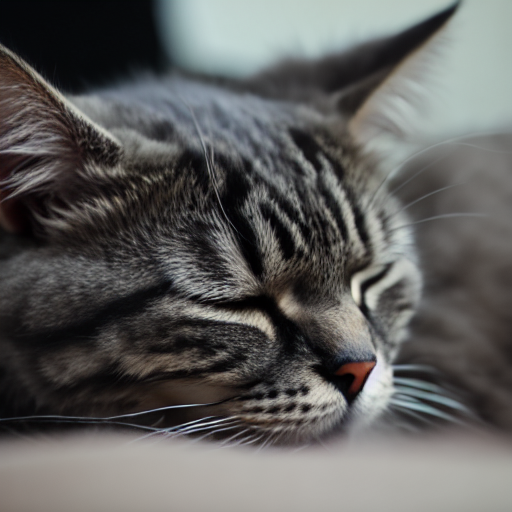} \\ \\
 \parbox{.3\linewidth}{geometric polygonal of earth globe}&\parbox{.3\linewidth}{Fantasy hero in armor. sketch art for artist creativity and inspiration}&\parbox{.3\linewidth}{The interior design of a lavish side outside garden, with a teak hardwood deck and pergola}\\ \\
 \includegraphics[width=.3\linewidth]{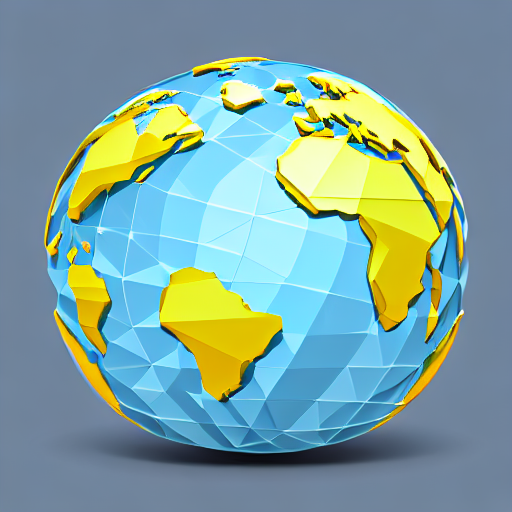}&
 \includegraphics[width=.3\linewidth]{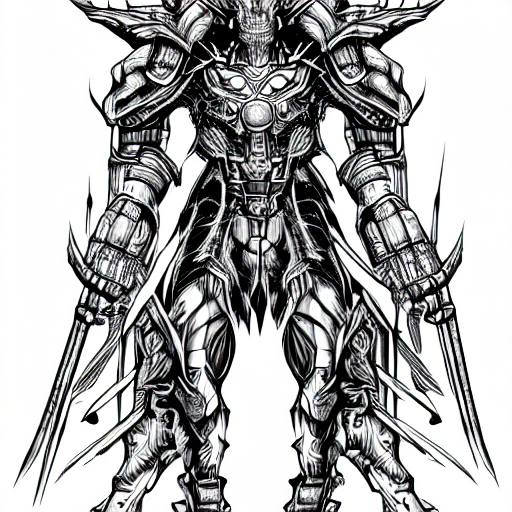}&
 \includegraphics[width=.3\linewidth]{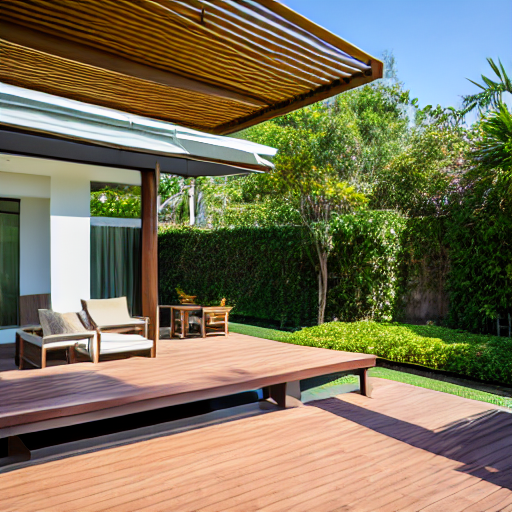} \\ \\
 \parbox{.3\linewidth}{dramatic sea dark sky thunderstorm ghost ship}&\parbox{.3\linewidth}{backpack covered with leaves}&\parbox{.3\linewidth}{new york city on a foggy afternoon, watercolor}\\ \\
 \includegraphics[width=.3\linewidth]{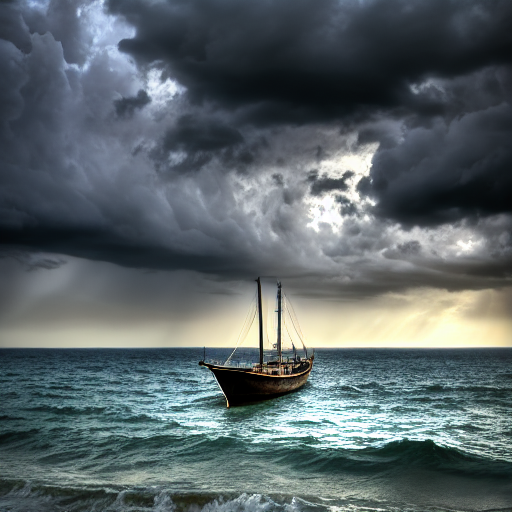}&
 \includegraphics[width=.3\linewidth]{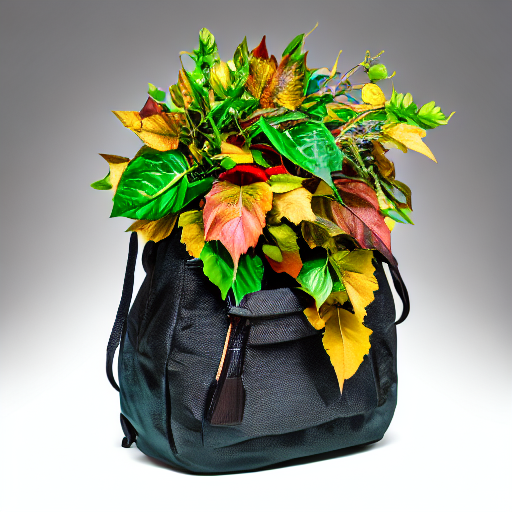}&
 \includegraphics[width=.3\linewidth]{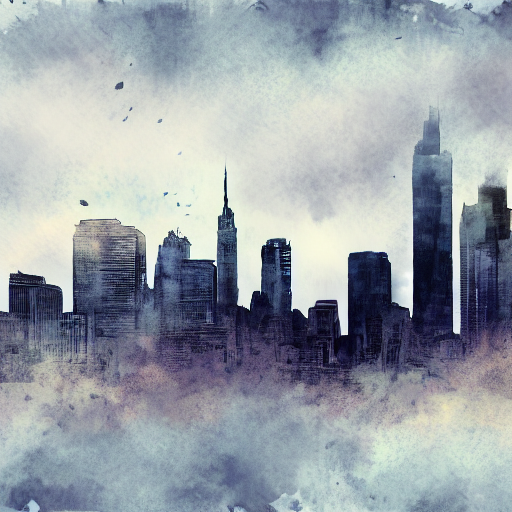}
 \end{tabular}
 \caption{Example images generated using the color conditional prior with our LDM, without using any color conditioning. The generations are similar in quality to the LAION prior + LDM showing no degradation by using color conditional prior instead of the regular prior. }
 \label{fig:appendix-without-color}
\end{figure*}

\begin{figure*}[t]
 \centering
 \begin{tabular}{c@{\hspace{0.3cm}}c@{\hspace{0.3cm}}c@{\hspace{0.3cm}}}
 \centering
 \parbox{.3\linewidth}{Acrylic colors and ink in water. Ink blot}&\parbox{.3\linewidth}{bulb breaking with splash of color, creative idea}&\parbox{.3\linewidth}{Cute cat. Portrait of a black tabby cat sleeping}\\ \\
 \includegraphics[width=.3\linewidth]{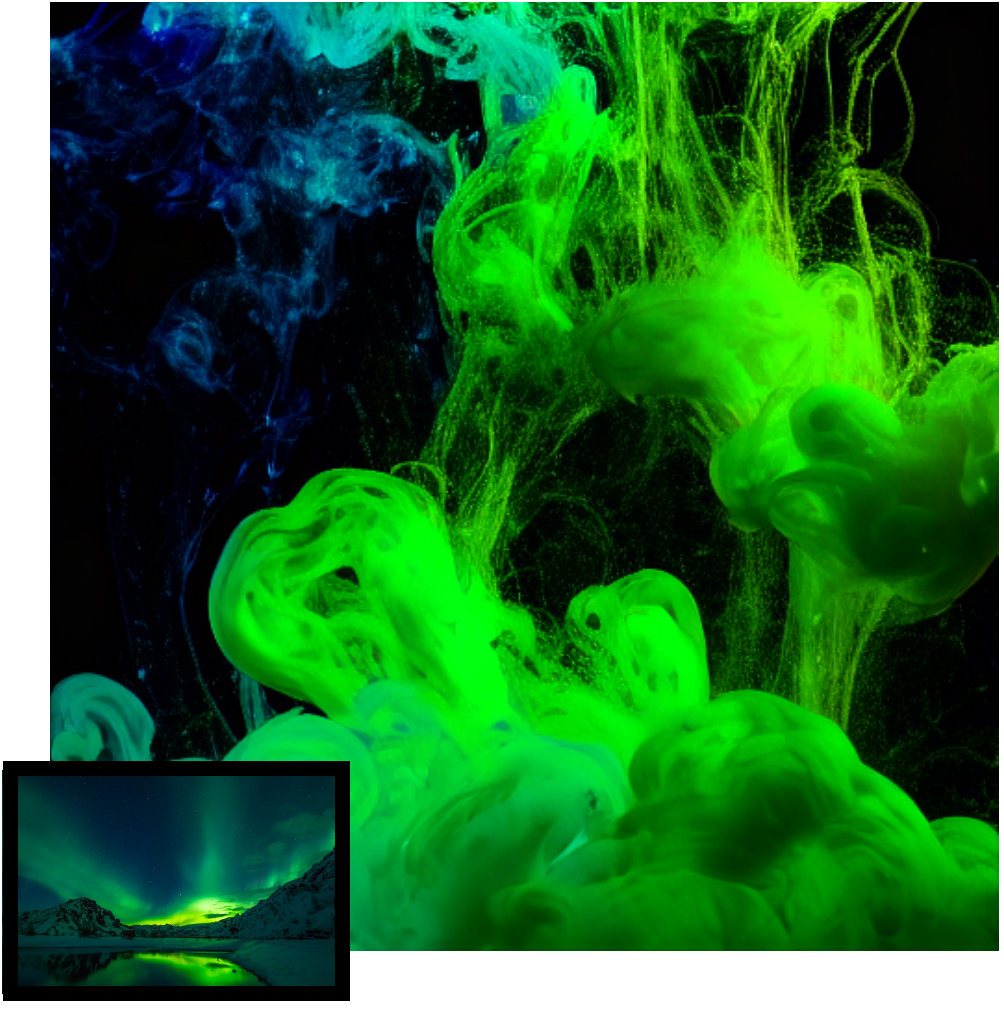}&
 \includegraphics[width=.3\linewidth]{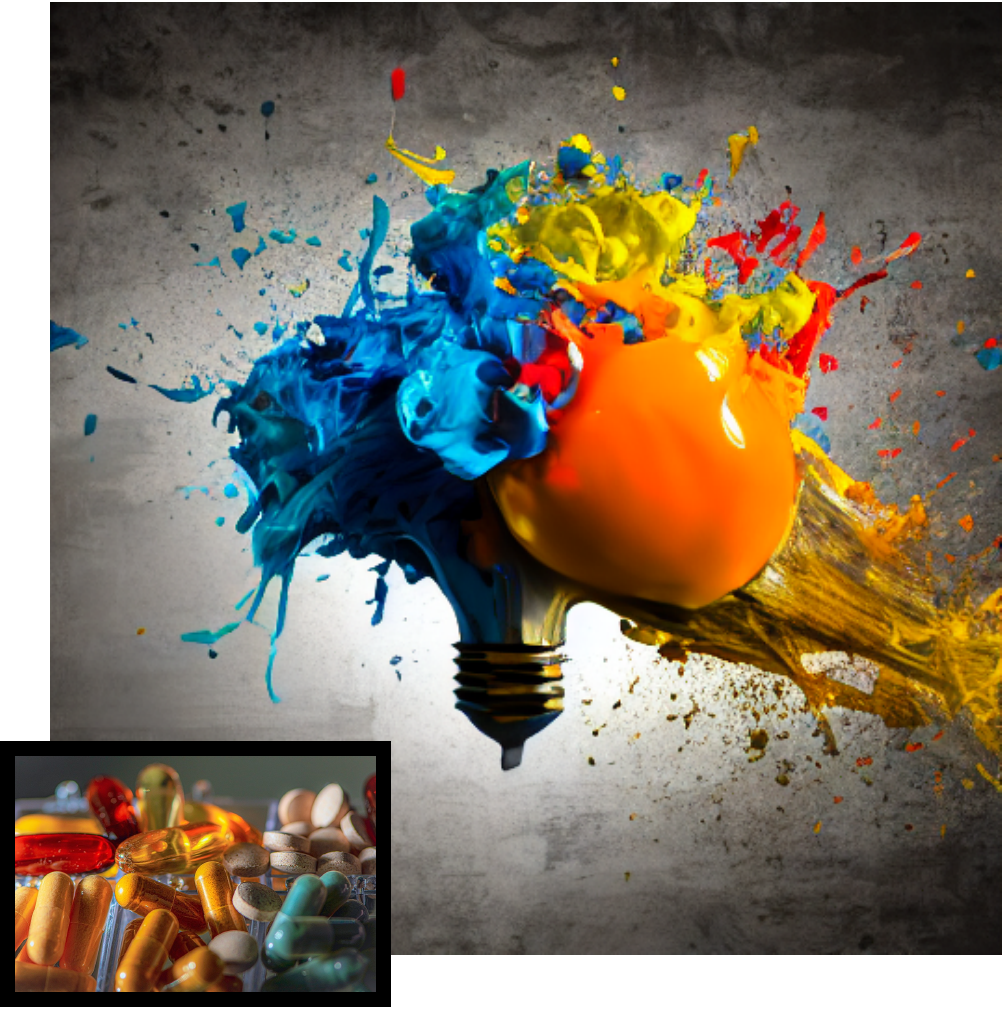}&
 \includegraphics[width=.3\linewidth]{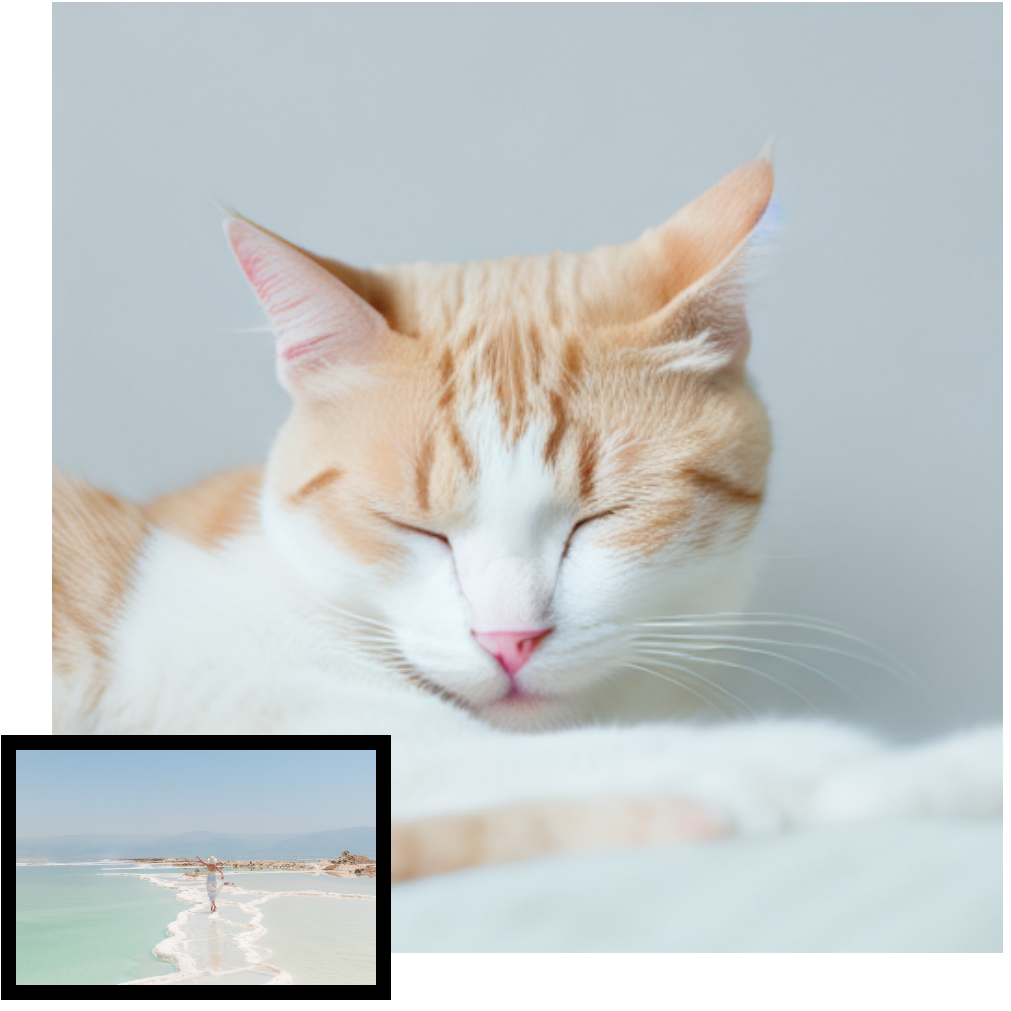} \\ \\
 \parbox{.3\linewidth}{geometric polygonal of earth globe}&\parbox{.3\linewidth}{Fantasy hero in armor. sketch art for artist creativity and inspiration}&\parbox{.3\linewidth}{The interior design of a lavish side outside garden, with a teak hardwood deck and pergola}\\ \\
 \includegraphics[width=.3\linewidth]{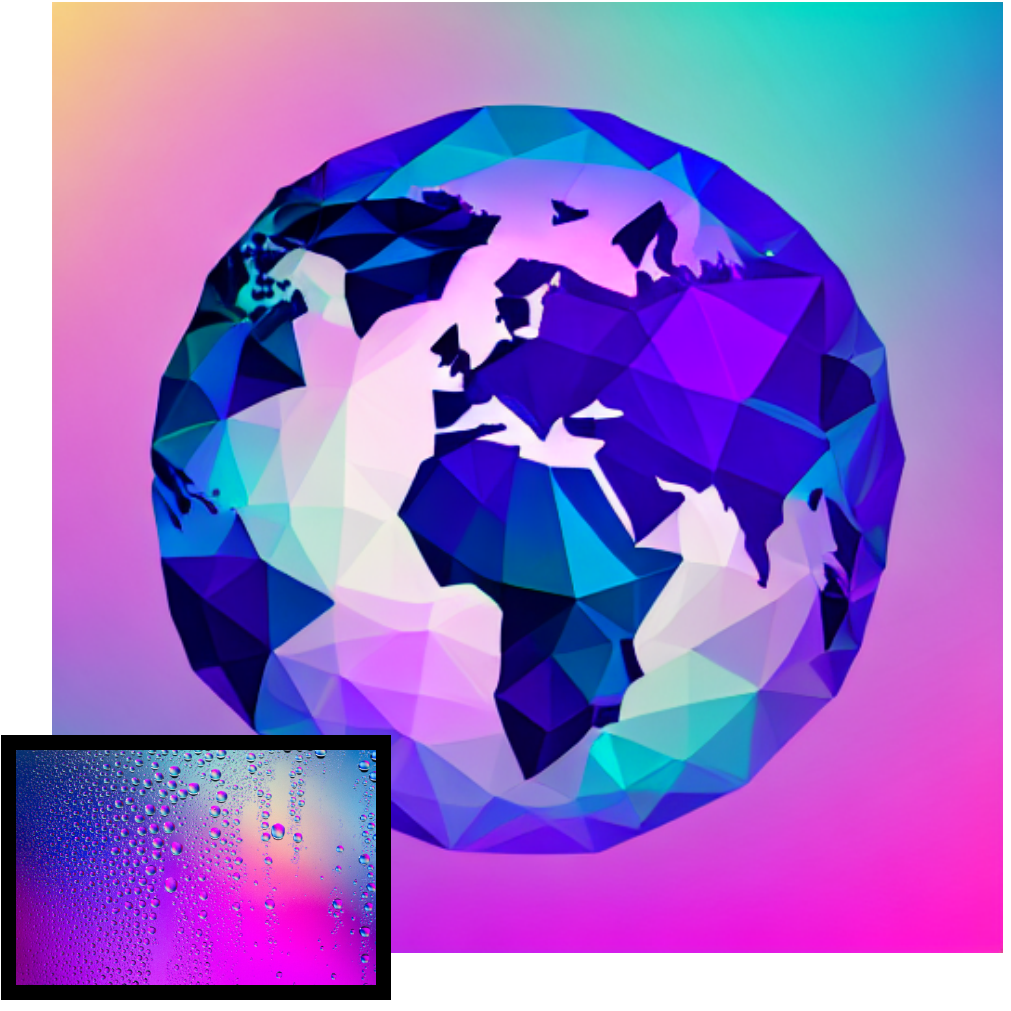}&
 \includegraphics[width=.3\linewidth]{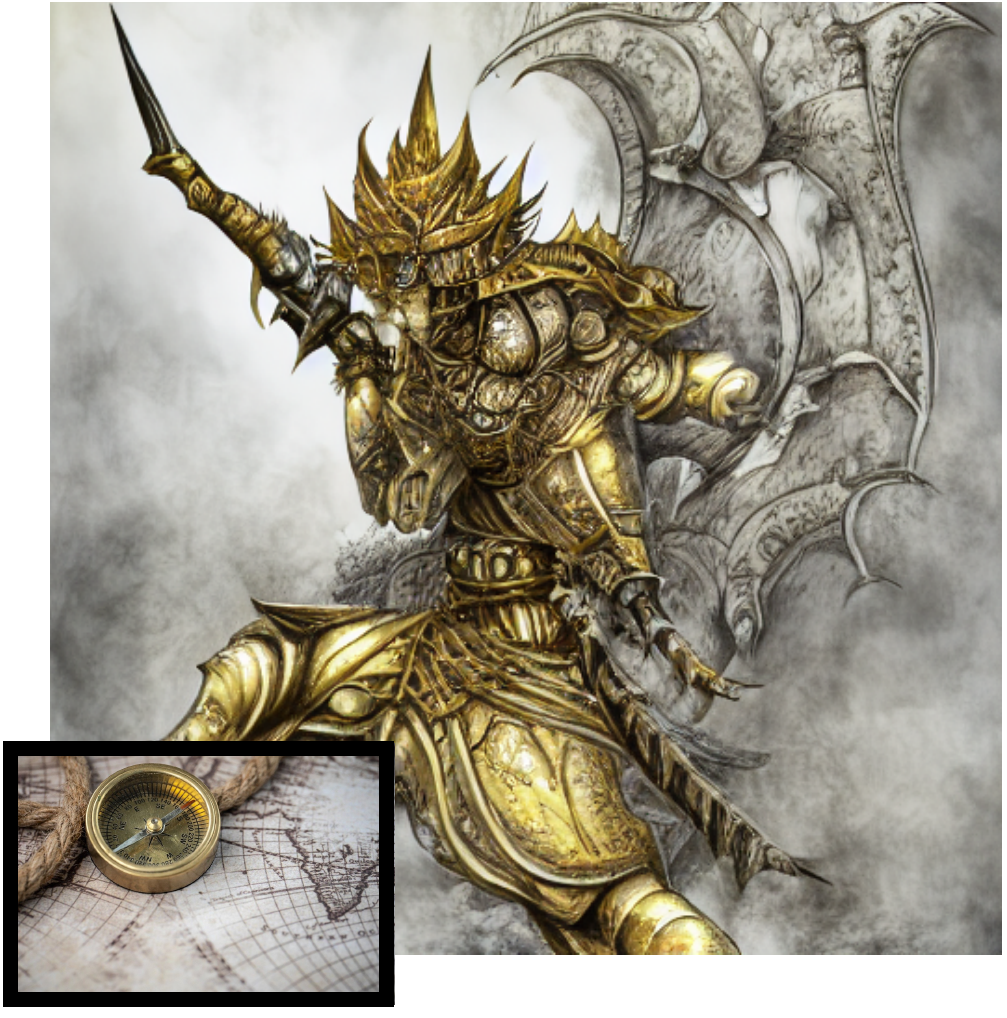}&
 \includegraphics[width=.3\linewidth]{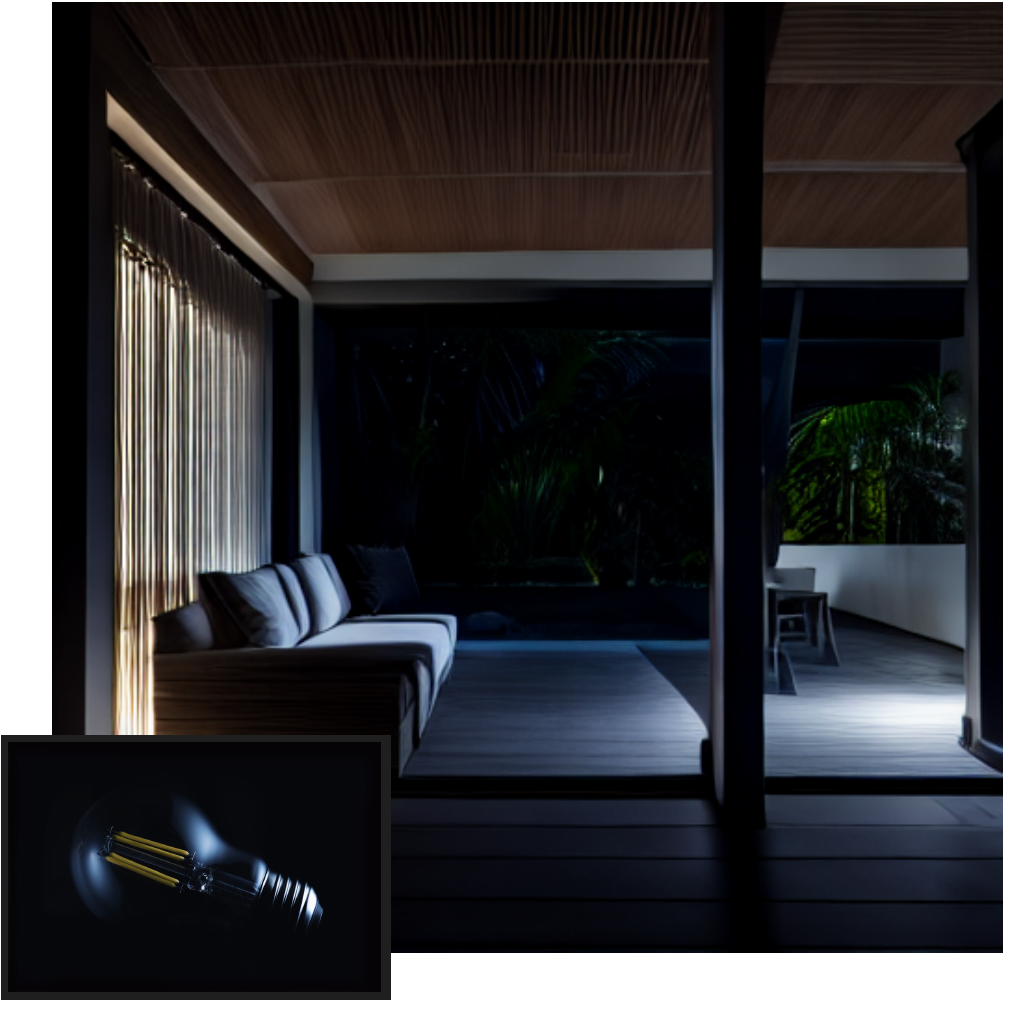} \\ \\
 \parbox{.3\linewidth}{dramatic sea dark sky thunderstorm ghost ship}&\parbox{.3\linewidth}{backpack covered with leaves}&\parbox{.3\linewidth}{new york city on a foggy afternoon, watercolor}\\ \\
 \includegraphics[width=.3\linewidth]{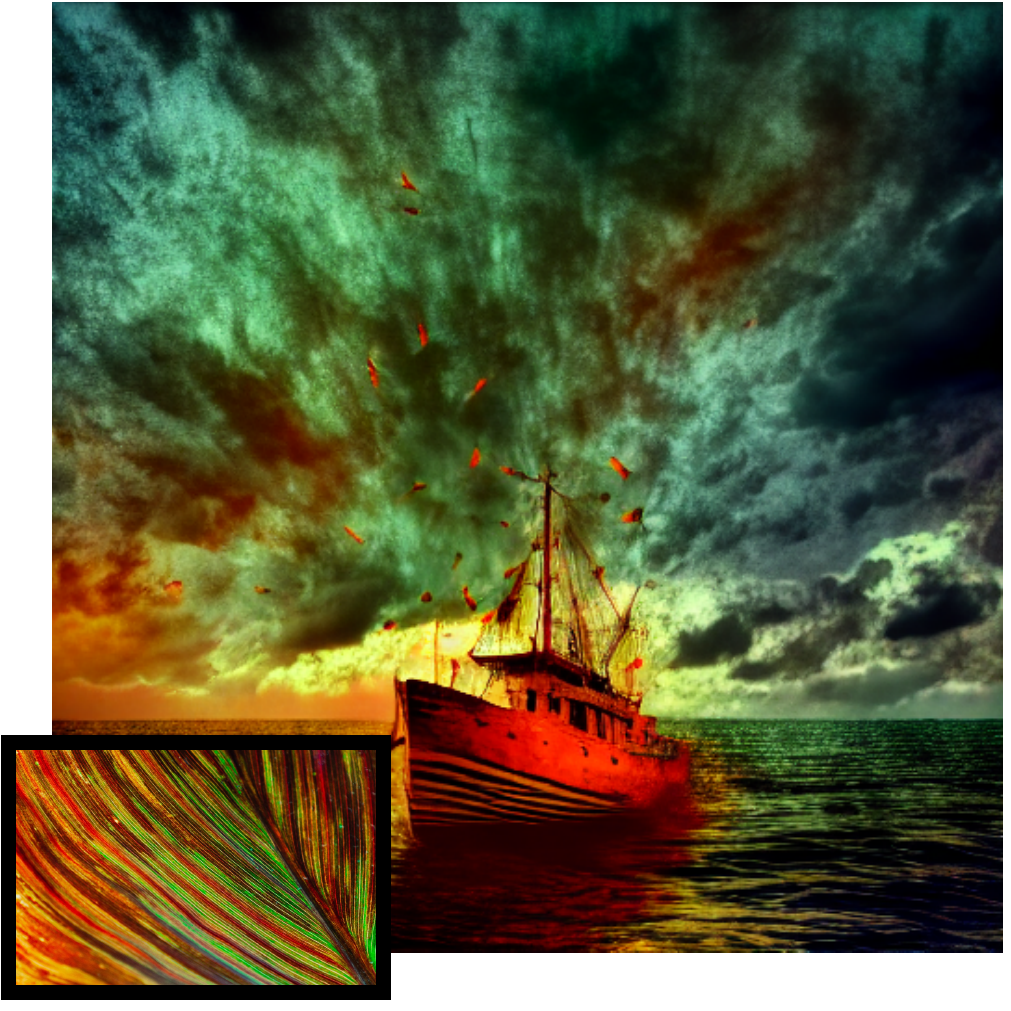}&
 \includegraphics[width=.3\linewidth]{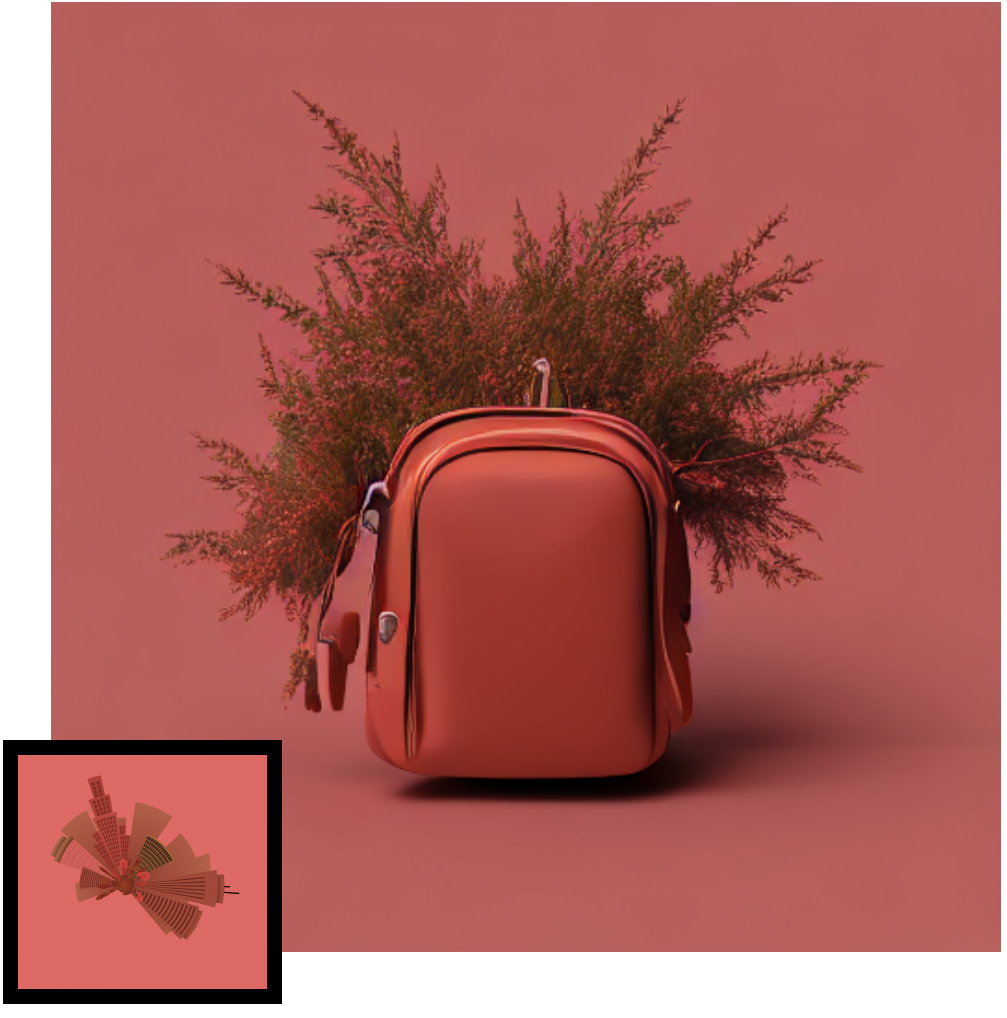}&
 \includegraphics[width=.3\linewidth]{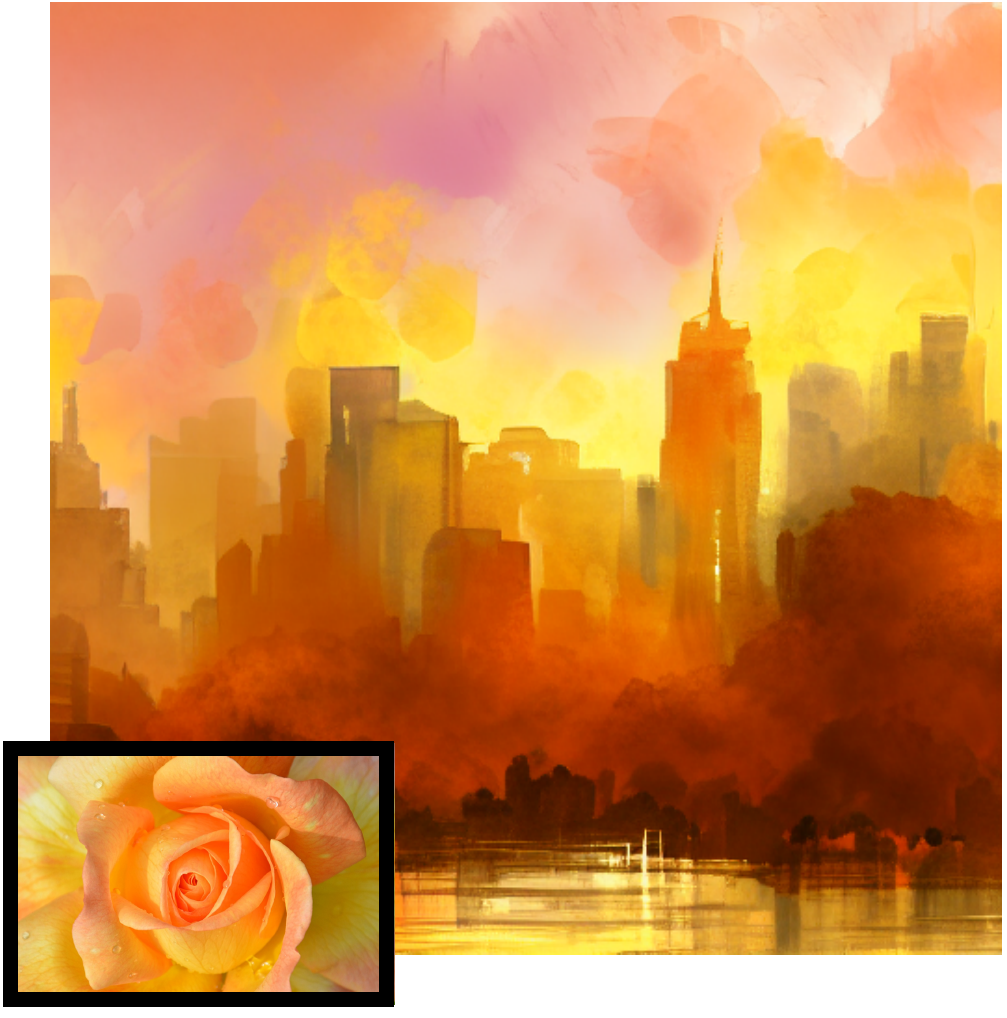}
 \end{tabular}
 \caption{Example images generated using the color prior with our LDM. We can observe that the color from exemplar image and the concept from input text are equally well represented in the generated images without loss of quality or relevance. Compare these generations with that of Fig.\ref{fig:appendix-without-color} to see the effect of color conditioned generation for the same prompts.}
 \label{fig:appendix-color}
\end{figure*}

We also provide additional examples of color conditional generation in Fig.\ref{fig:appendix-without-color} and Fig.\ref{fig:appendix-color}.


\end{document}